%% file: main_arxiv.tex
\documentclass[runningheads]{llncs}
\usepackage{graphicx}

\usepackage{tikz}
\usepackage{comment}
\usepackage{amsmath,amssymb} 
\usepackage{color}
\usepackage{booktabs}

\usepackage[width=122mm,left=12mm,paperwidth=146mm,height=193mm,top=12mm,paperheight=217mm]{geometry}
\usepackage{algorithm}
\usepackage{algorithmic}
\usepackage{caption}
\usepackage{subcaption}
\usepackage{multirow}
\usepackage{wrapfig}

\begin{document}
\pagestyle{headings}
\mainmatter
\def\ECCVSubNumber{7326}  

\title{Diverse Imagenet Models Transfer Better} 

\titlerunning{Diverse Imagenet Models Transfer Better}
%
\author{Niv Nayman\thanks{Equal contribution}\inst{1,2} \and
Avram Golbert$^*$\inst{1} \and \\
Asaf Noy\inst{1} \and
Tan Ping\inst{1} \and
Lihi Zelnik-Manor\inst{2}}
\authorrunning{N. Nayman et al.}
%
\institute{Alibaba Group, Tel Aviv, Israel \\ 
\email{\{niv.nayman,a.golbert,asaf.noy,xingye.tp\}@alibaba-inc.com}
\and
Technion - Israel Institute of Technology, Haifa, Israel \\
\email{lihi@technion.ac.il}}
\maketitle

\input{abstract}
\input{intro}
\input{related_work}
\input{method/feature_diversity}
\input{method/controlled_label_injection}
\input{experiments/exp}
\input{discussion}
\input{conclusions}

\clearpage
%
%

\bibliographystyle{splncs04}
\bibliography{references}

\input{appendix/appendix}
\end{document}

%% file: abstract.tex
\begin{abstract}
A commonly accepted hypothesis is that models with higher accuracy on Imagenet perform better on other downstream tasks, leading to much research dedicated to optimizing Imagenet accuracy. Recently this hypothesis has been challenged by evidence showing that self-supervised models transfer better than their supervised counterparts, despite their inferior Imagenet accuracy. This calls for identifying the additional factors, on top of Imagenet accuracy, that make models transferable.
In this work we show that high diversity of the features learnt by the model promotes transferability jointly with Imagenet accuracy. Encouraged by the recent transferability results of self-supervised models, we propose a method that combines self-supervised and supervised pretraining to generate models with both high diversity and high accuracy, and as a result high transferability. 
We demonstrate our results on several architectures and multiple downstream tasks, including both single-label and multi-label classification.
\keywords{Transfer Learning, Self-Supervised Learning}
\end{abstract}

%% file: intro.tex
\section{Introduction}
The success of Deep Neural Networks (DNNS) in a variety of computer vision tasks is largely related to their ability to transfer feature representations learned on a pre-trained task to leverage others.
A common practice is to pre-train a model on a large-scale supervised dataset such as ImageNet~\cite{russakovsky2015imagenet} and fine-tune it on the downstream (target) dataset that is typically of a smaller scale.
This practice has systematically advanced the state-of-the-art in tasks such as image classification~\cite{liu2021query2label,ridnik2021ml}, object detection~\cite{liu2021swin,ren2015faster} and semantic segmentation~\cite{he2017mask,liu2021swin}.
The pursuit after better pretrained models coincided with pushing the state-of-the-art performance on ImageNet, as it was shown that supervised pre-trained models that perform better on ImageNet also tend to perform better when fine-tuned on other vision tasks \cite{kornblith2019better}.

Recent works demonstrate that self-supervised pre-training (SSL) without any label information can also learn effective representations from upstream data (e.g., ImageNet) and even surpass supervised methods when transferring to downstream tasks.
This success in transfer learning, despite their relatively poor performance on ImageNet \cite{chen2020simple,he2020momentum,grill2020bootstrap,chen2021exploring,zbontar2021barlow,ericsson2021well} calls for identifying the additional factors, on top of Imagenet accuracy, that make models transferable.

While supervised training focuses on class-level discrimination, SSL focuses on instance discrimination, and models are trained to keep variants of the same instance close together in the representation space, and sometimes also, separated from different instances.
On the other hand, supervised models learn meaningfull high-level semantic features that are shared between instances of the same class, while SSL might capture irrelevant low level visual features (e.g., related to instance background). Thus high ImageNet performance guarantees that the features learnt are semantically meaningful and SSL learns diverse features.
This observation is supported by recent work that combines both supervised and self-supervised losses to improve transferability \cite{khosla2020supervised,islam2021broad}, yet those require intervention in the self-supervision stage and the transferability is attributed to other less important factors than feature diversity such as the abstraction of the representations learned (measured by Centered Kernel Alignment (CKA)~\cite{kornblith2019similarity} between layers) and intra-class variations.

\begin{figure}[htb]
\vspace{-5mm}
\centering
\begin{subfigure}{.49\textwidth}
  \centering
    \includegraphics[width=0.98\textwidth, height=!]{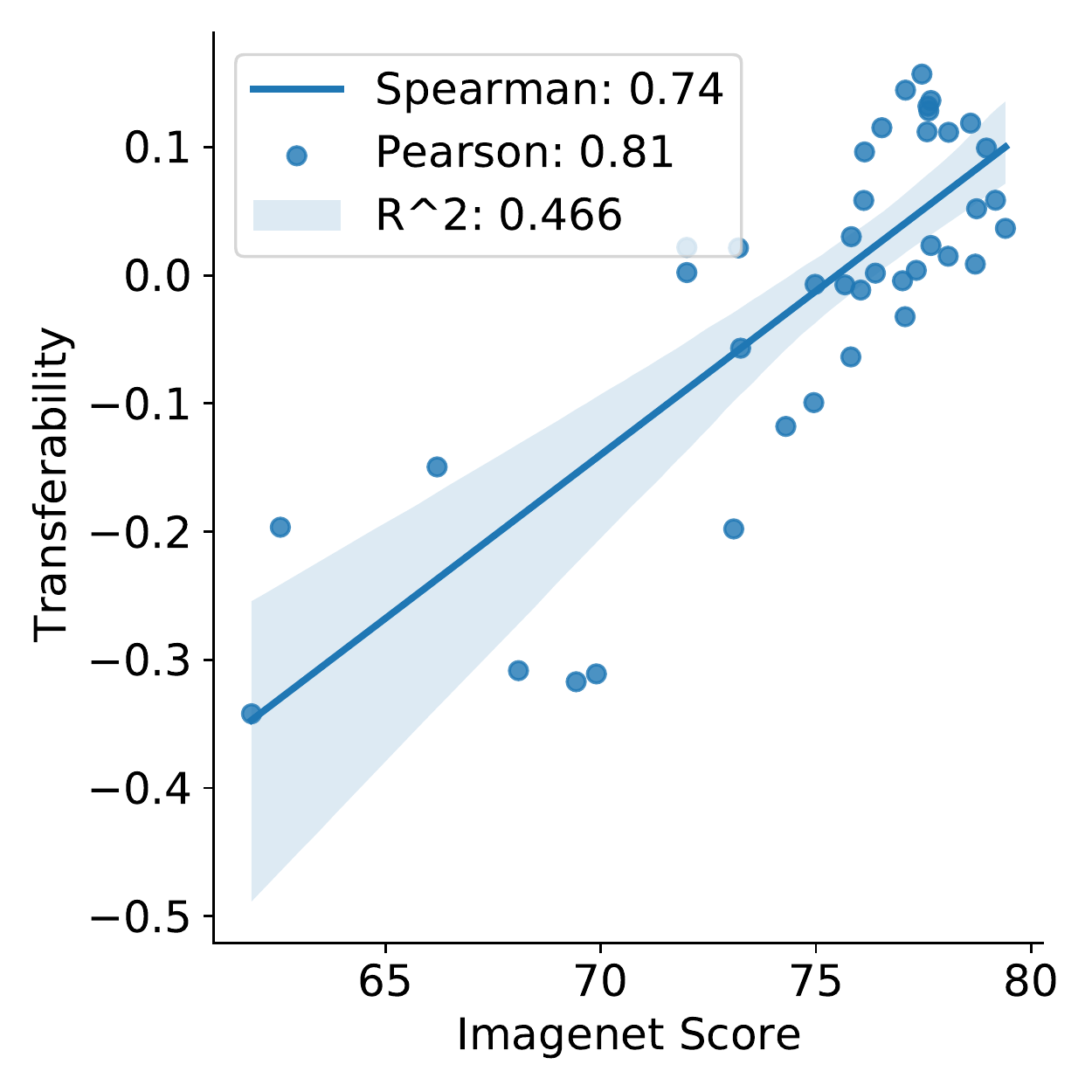}
  \label{fig:label_injection_cka}
\end{subfigure}
\begin{subfigure}{.49\textwidth}
  \centering
    \includegraphics[width=0.98\textwidth, height=!]{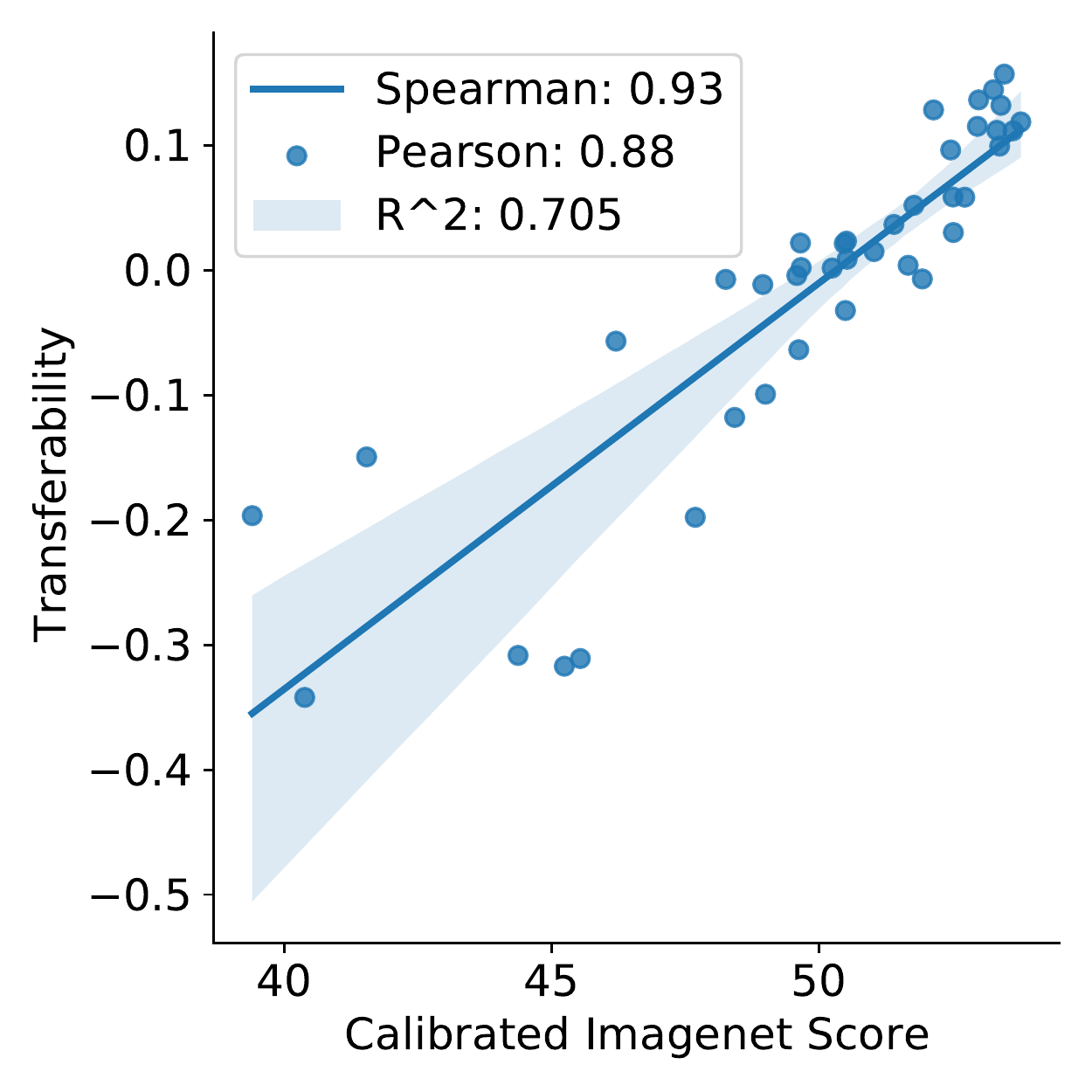}
  \label{fig:label_injection_calibrated}
\end{subfigure}

\caption{Transferability vs (Left) Imagent Score, and (Right) the \emph{Calibrated} Imagenet Score (CIS) for 40 models that were pre-trained with supervised learning, self-supervised learning or their combination. CIS correlates with transferability significantly better than imagenet score.}
\label{fig:tl_correlation}
\vspace{-5mm}
\end{figure}

Our contribution is two-fold: (1) We propose \textit{Feature Diversity} as a calibration for the Imagenet accuracy for assessing the transferability of 
models.
\begin{align}
\textit{Calibrated Imagenet Score} = \text{Imagenet Accuracy} \times \text{Feature Diversity}
\end{align}
As we empirically validate that \textit{Calibrated Imagenet Score (CIS)} better correlates with trasferability, as shown in Figure~\ref{fig:tl_correlation}.

(2) We introduce a simple training method \textit{Controlled Label Injection (CLI)}, that enables the injection of label information into any self-supervised pre-trained model in a controlled manner, for generating models of different feature diversity and Imagenet accuracy. The resulted models increase ImageNet performance while either improving or maintaining feature diversity of the self-supervised model.
This both allows us to make observations about the connection between the CIS and transferability, while at the same time this leads to models with higher transferability.

We validate our approach over both CNNs (ResNets~\cite{he2016deep}) and vision transformers (ViT~\cite{dosovitskiy2021an}), several self-supervised pre-training methods (e.g., MoCo-v2~\cite{chen2020improved},
SimCLR~\cite{chen2020simple},
SwAV~\cite{caron2020unsupervised}, DINO~\cite{caron2020unsupervised} and MAE~\cite{MaskedAutoencoders2021}), two formulations of Feature Diversity, several downstream vision tasks, including multi-label classification on the MS-COCO~\cite{lin2014microsoft} dataset and a variety of 14 single-label classification datasets.

In section~\ref{sec:feature_diversity} we introduce the notion of feature diversity.  In section~\ref{sec:method}, we describe the \textit{Controlled Label Injection} scheme, for generating models of different feature diversity and Imagenet accuracy.
In section~\ref{sec:exp}, we describe the experimental settings and show that the label injected models have higher transferability than alternative ways for combining SSL with superived models. In section~\ref{sec:fesature_importance}, we show that Feature Diversity is more important for transferability than other previously suggested factors~\cite{kornblith2021better,islam2021broad}, such as the level of abstraction learnt by the model and class separation.

%% file: related_work.tex
\section{Related Work}
\subsubsection{Transfer learning}
was shown to be highly effective in transferring knowledge from upstream (source) datasets to typically much smaller datasets given that their domains are similar~\cite{mensink2021factors}. 
Huh et al.~\cite{huh2016makes} searched for the properties that make a dataset a good choice for transfer learning.
Kornblith et al~\cite{kornblith2019better} 
showed that when it comes to supervised models, ImageNet accuracy score is highly correlated with performance over downstream tasks, 
confirming the common practice of selecting pre-trained models for transfer learning based on their Imagenet accuracy.
We show that when self-supervised models are included,
the correlation significantly drops, calling for improved measures for selection. The architecture and depth of CNNs were also shown to impact transfer performance~\cite{azizpour2015factors}.
The effects of the pre-training loss function were studied by~\cite{kornblith2021better,khosla2020supervised,islam2021broad}. Improved Imagenet score might actually lead to worse transfer learning results when used as fixed feature extractor, while the choice of the loss has little effect when networks are fully fine-tuned on the new tasks as shown by~\cite{kornblith2021better}. A combination of contrastive and supervised learning was shown to improve transfer leaning performance~\cite{khosla2020supervised,islam2021broad}, but the factors driving the performance are still not completely
understood.
Centered Kernel Alignment (CKA)~\cite{kornblith2019similarity} was utilized by \cite{kornblith2021better} 
to show that differences among loss functions are apparent only in the last few layers of the network, and \cite{islam2021broad} further showed that contrastive models contain more low level and mid-level features in those layers.
Both \cite{kornblith2021better} connect this to intra-class variations, concluding that representations with higher class separation obtain higher accuracy on the upstream task, but their features are less useful for downstream tasks.
In this work, we identify feature diversity as a more important factor that implies on the transferability of the model, even in the more practical use-case of fine-tuning the pretrained models on the downstream tasks.
\subsubsection{Self-Supervised Learning (SSL)}
is a subset of unsupervised learning, where neural networks are explicitly trained with automatically generated labels. In earlier works, labels were generated by diverse pre-text tasks such as prediction of rotation~\cite{komodakis2018unsupervised}, colorization~\cite{zhang2016colorful}, pathces positions~\cite{doersch2015unsupervised} and others~\cite{jing2020self}. More recent methods can be roughly divided to contrastive methods~\cite{chen2020improved,he2020momentum,chen2020simple,grill2020bootstrap} and clustering methods~\cite{caron2020unsupervised,li2020prototypical,asano2019self}.
Notably, MoCo-v2~\cite{chen2020improved}, SimCLR~\cite{chen2020simple}, SwAV~\cite{caron2020unsupervised} have shown a dramatic improvement in representation quality  learned from unlabeled Imagenet images, surpassing the performance of modern supervised methods over various downstream tasks~\cite{ericsson2021well}. 
They also showed that while self-supervised features seems to discard color information, their attentive focus is higher compared to their supervised counterparts.
This motivated the proposals of hybrid methods. \cite{khosla2020supervised} proposed a new contrastive loss to leverage the label information and \cite{islam2021broad} combined it with both constrastive and cross-entropy losses. However, it is yet unclear 
what self-supervised features should be maintained and how in order to improve resulting models' transferability.
In this work, we propose a measure that captures the diversity of the information encoded in different networks, and a simple method to inject supervised label information to a pre-trained SSL model, in a way that maintains this diversity in order to improve transfer learning performance.

\subsubsection{Feature Diversity.}
It has been shown that a significant portion of features extracted by DNNs are redundant \cite{NIPS2013_7fec306d,rodriguez2016regularizing,bengio2009slow,changpinyo2017power,ayinde2017nonredundant}.
By simply training on a low-rank decomposition of the weight matrices, \cite{denil2013predicting} demonstrated that a fraction of the parameters is sufficient to reconstruct the entire network. 
\cite{ayinde2019correlation} estimated the number of redundant features in each layer, by hierarchically clustering those according to their relative cosine distances in feature space. 
\cite{Rodriguez0CGR17,ayinde2019regularizing} proposed regularizing correlated features based on their relative cosine distances to yield a network with diverse features, with less overfitting, and better generalization. 
 \cite{MarietS15a,acharyya2021diversity} use determinantal point processes to select a subset of diverse neurons or connections and subsequently fuse the redundant ones into the selected ones for the purpose of pruning. Differently from the aforementioned, which deal mainly with reducing overfitting and pruning, in this work we focus on the importance of learning diverse features for the purpose of transfer learning.

%% file: method/feature_diversity.tex
\section{Measures of Feature Diversity}
\label{sec:feature_diversity}
Previous work connected the transferability of a model to data-dependent measures, such as the abstraction of representations learnt for the upstream data and the variations in its embedding space~\cite{kornblith2021better,islam2021broad}, the number of non-zero elements in the activations~\cite{kornblith2021better} and robustness to corrupted data~\cite{islam2021broad}. Considering that transfer learning deals with different, sometimes unknown in advance, downstream tasks, we instead search for a connection to a data-independent intrinsic property of the model. 
Intuitively, the more diverse the information learnt by the pre-trained model is, the more likely this information can be utilized in transfer learning to a larger variety of downstream tasks.
With this intuition, since the information learnt by the pre-trained model is encoded in its weights, 
we need a way to quantify the diversity of features learnt by the pre-trained model.

\begin{figure}[htb]
\vspace{-6mm}
    \centering
    \includegraphics[width=0.98\textwidth, height=!]{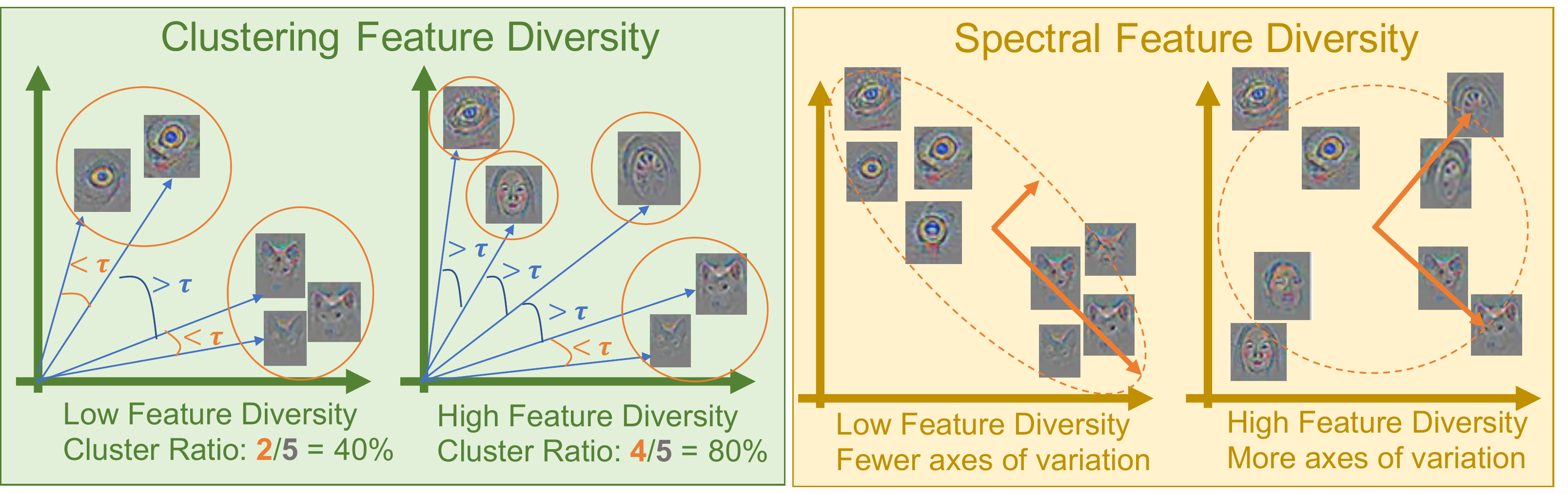}
  \caption{Illustration of the two proposed measures for Feature Diversity. (Left) Low and high diversity result in low and high clustering ratio respectively. (Right) For low and high diversity the variance of features is explained by fewer or more directions respectively.}
  \vspace{-10mm}
  \label{fig:div_illustration}
\end{figure}

We propose two measures, illustrated in Figure~\ref{fig:div_illustration}, both of which view the weights of the various neural layers as vectors in a metric space. 
The measures quantify the scatter of those vectors in the feature space. 
We describe in more detail the first measure, which is based on clusterability properties of the features. 
The second measure, based on spectral analysis of the features distribution, is presented in more details in Appendix~\ref{apdx:spectral_diversity} for brevity.
Empirical evaluation with both measures leads to similar conclusions and validates the importance of feature diversity for transferability (Figure~\ref{fig:xgboost}).

\subsubsection{Clustering Feature Diversity}
\label{sec:cluster_diversity}
The first measure we propose to evaluate feature diversity is based on assessing the organization of the features into clusters, and is inspired by~\cite{ayinde2019correlation}. Features that are grouped together into tight clusters imply low diversity, while features that are sparsely spread imply high diversity. We next propose a method to measure the overall clusterability of a deep neural network's features across all of its layers.

Let $W=[w_i,\dots,w_n]\in\mathbb{R}^{d\times n}$ be a weight matrix whose columns $\{w_i\}_{i=1}^n$ are its features. 
We apply the agglomerative clustering approach of~\cite{ding2002cluster,walter2008fast}, while adjusting it to fit our purpose. The clustering continues agglomeratively, merging two clusters $C_a$ and $C_b$ as long as their average mutual cosine similarity $\mathcal{S}_C(C_a,C_b)$~\cite{leibe2004combined,manickam2000intelligent} crosses some threshold $\tau$:
\begin{align}
    \label{eqn:cluster}
    \mathcal{S}_C(C_a,C_b)=\frac{1}{|C_a|\cdot|C_b|}\sum_{w_a\in C_a, w_b\in C_b}
    cosine(w_a,w_b)
    > \tau
\end{align}
The \emph{cluster ratio} between the number of clusters and the number of features for a given threshold $\tau$ quantifies the resulted clusterability, as illustrated in Figure~\ref{fig:div_illustration} (Left). 
Due to different neural layers of the same model learning different levels of abstractions, a single threshold $\tau$ value does not fit all. Hence, differently from \cite{ayinde2019correlation}, we evaluate the clusterability of the entire model by averaging the cluster ratio of all layers across a spectrum of threshold values. For the full technical details and illustrations see Appendix~\ref{apdx:cluster_diversity}.
\begin{figure}[htb]
    \centering
    \includegraphics[width=0.8\textwidth, height=!]{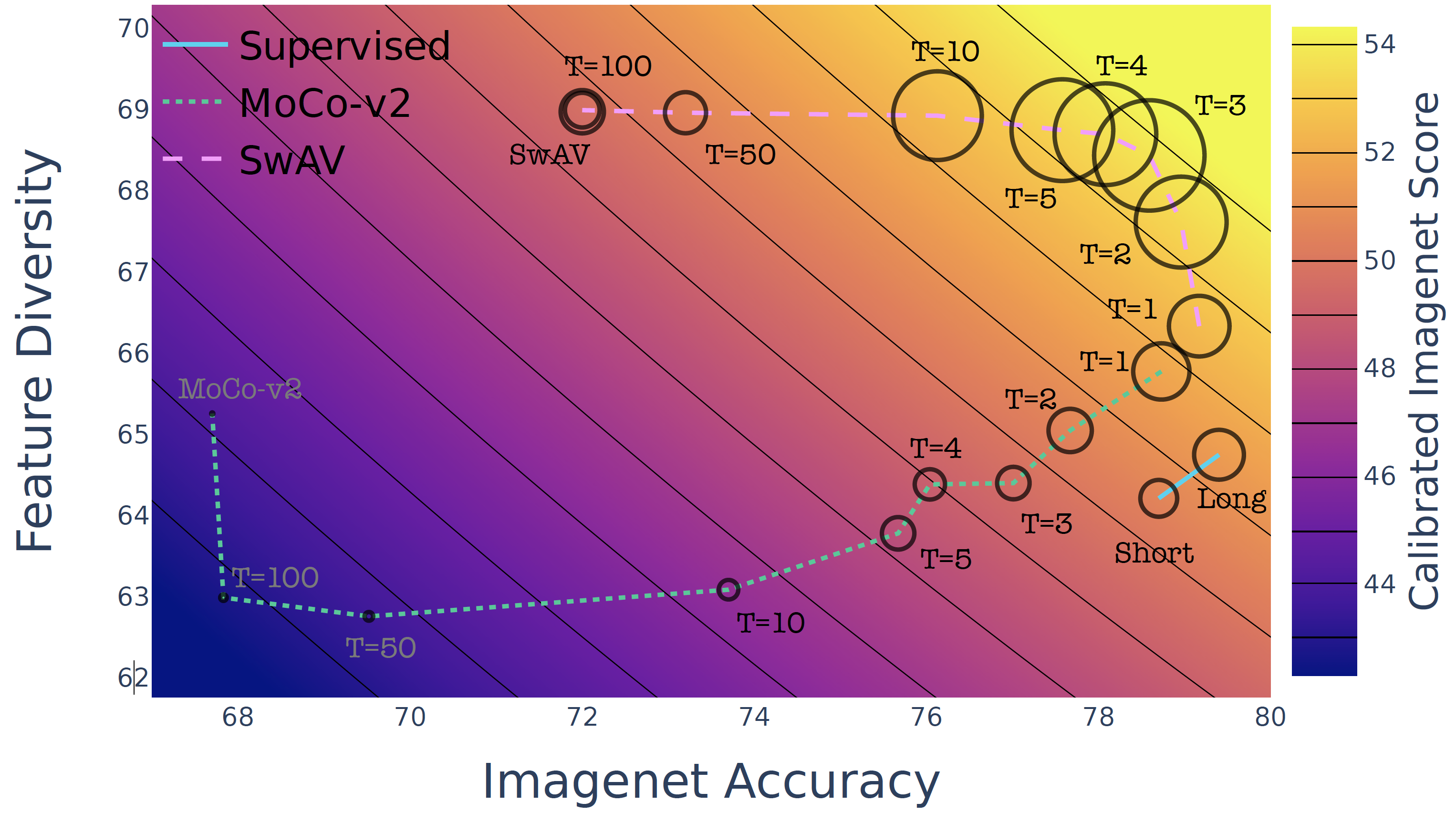}
  \caption{Circle size corresponds to the transferability averaged over 14 downstream tasks, as a function of Imagenet top-1 accuracy (x axis) and feature diversity (y axis). Results shown for 3 different training methods (see legend). The background colors and curves show the Calibrated Imagenet Score. Evidently, models with both high Imagenet accuracy and high Feature Diversity, that together result in high Calibrated Imagenet Score (in yellow), transfer better (larger circles).}
  \label{fig:imgnt_div_cor_cnn}
\end{figure}

\subsubsection{Spectral Feature Diversity}
\label{sec:spectral_diversity}
Another way to evaluate the distribution of features is suggested next, based on spectral analysis of the feature vectors. Principal component analysis (PCA)~\cite{karl1901pca} is an effective approach for evaluating variance along principal directions in the feature space.
Low diversity implies that the feature distribution can be captured by a small number of principal directions, while high diversity implies requiring many principal directions, as illustrated in Figure~\ref{fig:div_illustration} (Right) while the technical details and exact calculations are provided in Appendix~\ref{apdx:spectral_diversity} for brevity. Table~\ref{tab:PearsonTable} shows that both measures of Feature Diversity improve the correlation of the Calibrated Imagenet Score with the transferability in both cases of linear probin and finetuning, while for the former the spectral version is favourable, and for the latter the clustering based one is. For the corresponding Figure~\ref{fig:tl_correlation} in the case of linear probing see Appendix~\ref{apdx:linprob_conc}.

\begin{table}[H]
    \centering
    \begin{tabular}{l|cccc|cccc}
        \toprule
        {} & \multicolumn{4}{c|}{Linear Probing} & \multicolumn{4}{c}{Finetune} \\
        {} & $\rho$ & $r$ &    $R^2$ & $\tau$ & $\rho$ & $r$ &    $R^2$ & $\tau$ \\
        \midrule
        ImageNet Score &  0.55 &    0.73 &  0.13 &        0.38 &   0.74 &    0.81 &  0.47 &        0.53 \\
        \hline
        Spectral CIS   &    \textbf{0.91} &    \textbf{0.89} &  \textbf{0.74} &        \textbf{0.75} &  0.89 &    0.83 &  0.56 &        0.72 \\
        Cluster CIS    &  0.89 &    0.88 &  0.71 &        0.72 &   \textbf{0.93} &    \textbf{0.88} &  \textbf{0.70} &  \textbf{0.77} \\
        \bottomrule
    \end{tabular}
    \caption{The Spearman ($\rho$), Pearson ($r$), $R^2$ and Kendall-tau ($\tau$) correlation coefficients between transferability and standard or Calibrated Imagenet Score (CIS) computed with the proposed feature diversity measures. Both diversity based CIS measures show significantly higher correlation than the plain Imagenet accuracy, with an advantage to Spectral Feature Diversity for linear probing and to Clustering Feature Diversity for finetune.}
    \label{tab:PearsonTable}
\end{table}


%% file: method/controlled_label_injection.tex
\section{Training Scheme for Calibrated Imagenet Score }
\label{sec:method}

\begin{figure*}
\vspace{-12mm}
\noindent\begin{minipage}[b]{.65\textwidth}
\input{method/algo}
\end{minipage}%
\begin{minipage}[b]{.34\textwidth}
    \centering
    \includegraphics[width=0.98\textwidth, height=!]{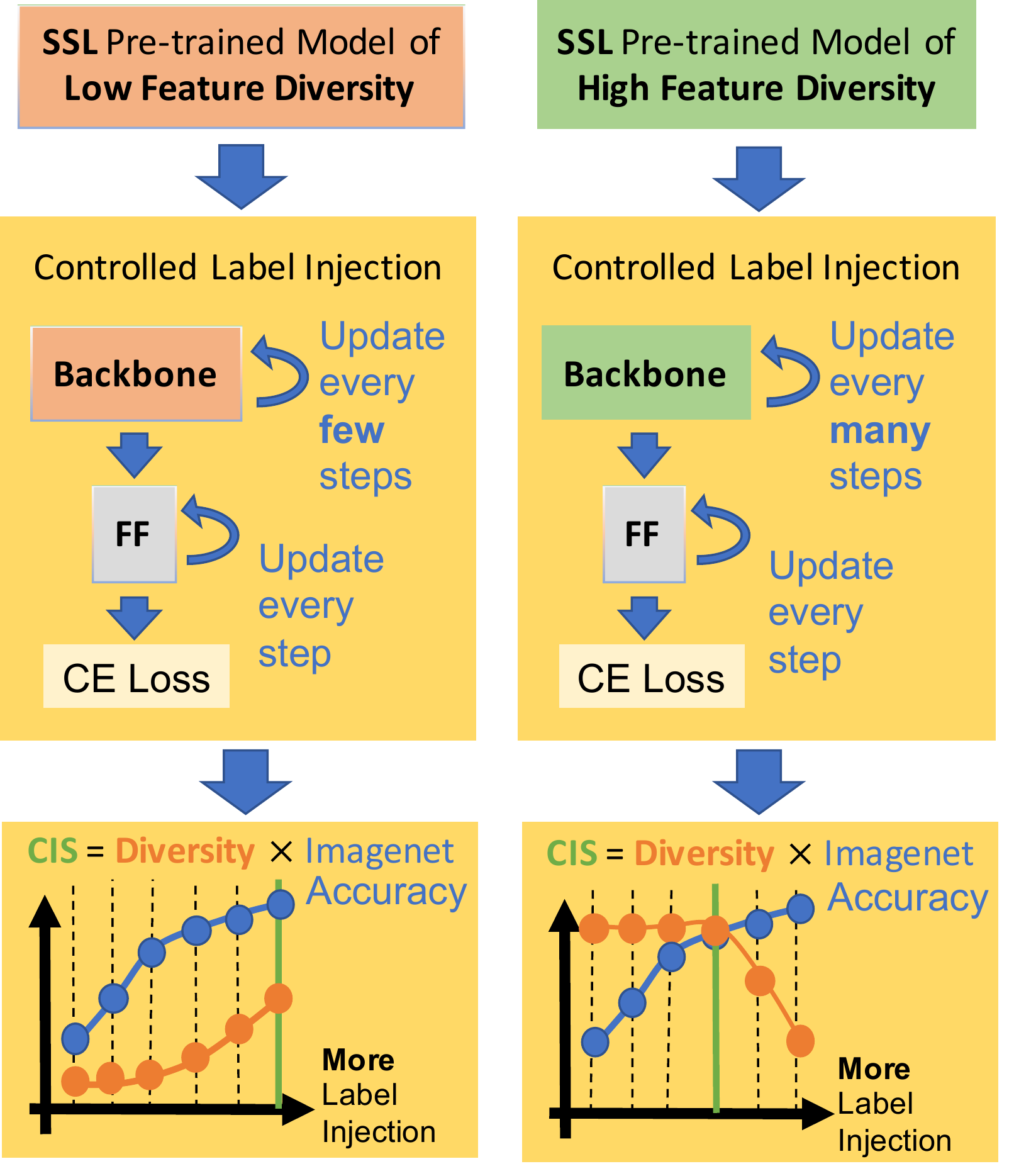}
  \captionof{figure}{CLI Scheme}
  \label{fig:scheme}
\end{minipage}
\end{figure*}

We present a simple scheme for training that produces models with both high feature diversity as well as high ImageNet accuracy. The training scheme is generic in the sense that it can be applied to any type of model and training method. 
The proposed scheme, illustrated in Figure~\ref{fig:scheme}, is composed of two stages. First, we train a model using Self-Supervised Learning (SSL). 
Then, we gradually introduce supervision by injecting label information in a controlled manner, while fine-tuning the model. 

\subsection{The Controlled Label Injection Algorithm}
\begin{wrapfigure}{r}{0.5\textwidth}
\vspace{-15mm}
  \begin{center}
    \includegraphics[width=0.48\textwidth]{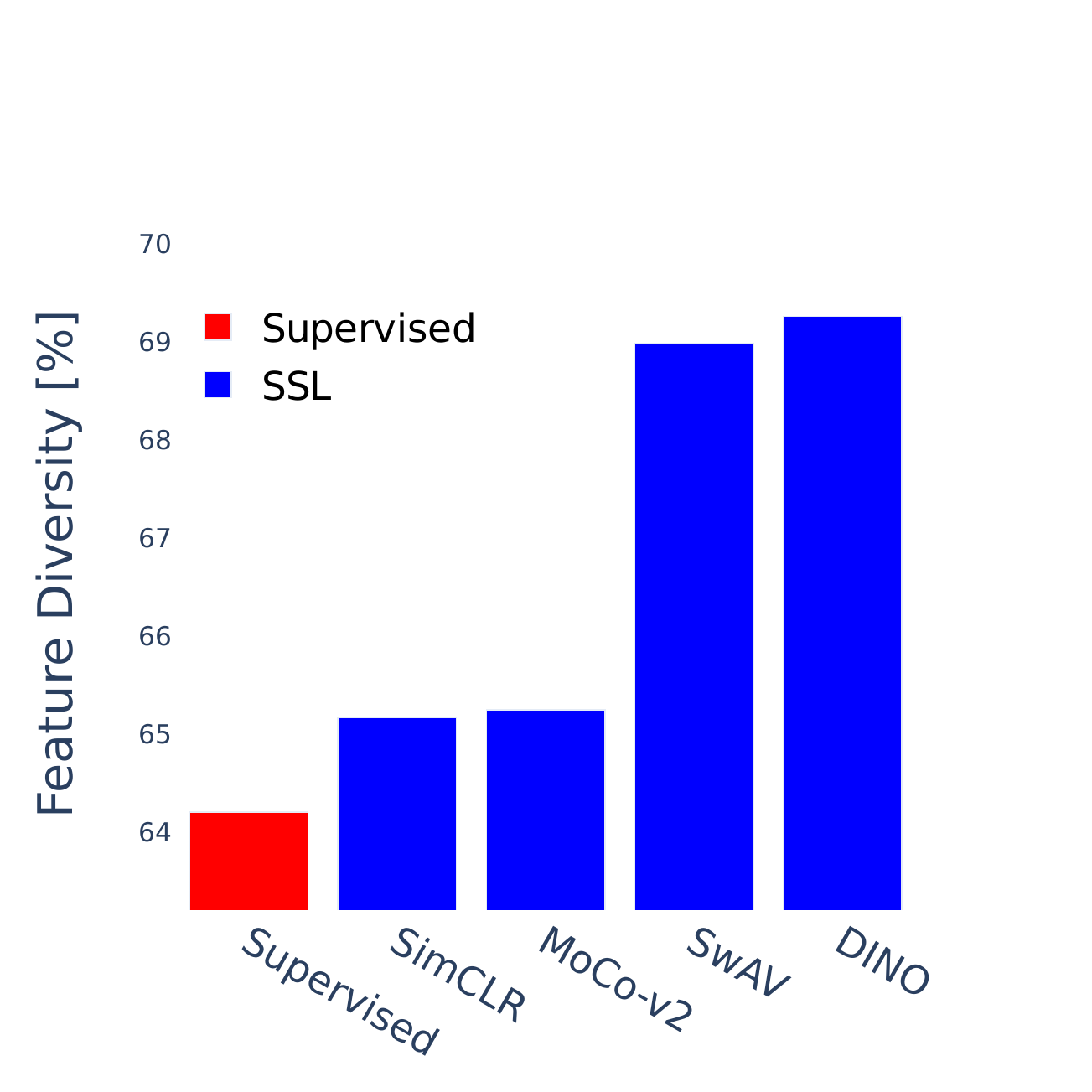}
  \end{center}
  \vspace{-5mm}
  \caption{Self-supervised methods learn more diverse features.}
  \vspace{-8mm}
  \label{fig:ssl_high_diversity}
\end{wrapfigure}
Our scheme has two stages. It starts with Self-supervised learning (SSL) that typically yields pre-trained models with higher diversity, because the underlying contrastive learning views each sample as a unique class. We observe that features learnt by SSL methods are richer and more diverse compared to supervised models that effectively capture much fewer classes, as shown in Figure~\ref{fig:ssl_high_diversity}.
The second stage injects label information into any self-supervised pre-trained model in a controlled manner, and through that increases ImageNet accuracy while at the same time either improving or maintaining feature diversity of the self-supervised model. 
We show empirically that this scheme leads to models with higher transferability.

Denote by $f_{W_B}$ and $g_{W_{FF}}$ the backbone model and the classification model on top of it, with weights $W_B$ and $W_{FF}$ respectively, such that $\hat{y}=g_{W_{FF}}(f_{W_B}(x))$ holds for an input sample $x$ and its predicted label $\hat{y}$. The backbone weights $W_B$ were trained by any self-supervised method and $W_{FF}$ is randomly initialized.

We fine-tune the weights by training with controlled supervision, according to Algorithm~\ref{alg:controlled_label_injection}. The algorithm updates the backbone $W_B$ once every $\mathcal{T}$ updates of the classifier $W_{FF}$, thus the classifier is encouraged to undertake most of the classification burden, and only some of it is passed through to the backbone, whose weights change more slowly. Effectively $\mathcal{T}=1$ is a standard fine-tuning and $\mathcal{T}\rightarrow\infty$ is linear probing. Hence, the diversity is maintained for a large enough control cycle $\mathcal{T}$, when starting from models of high feature diversity.

\vspace{-10mm}
\subsection{Empirical Analysis of the Controlled Label Injection.}
\begin{figure}
\vspace{-5mm}
\centering
\begin{subfigure}{.49\textwidth}
  \centering
    \includegraphics[width=0.98\textwidth, height=!]{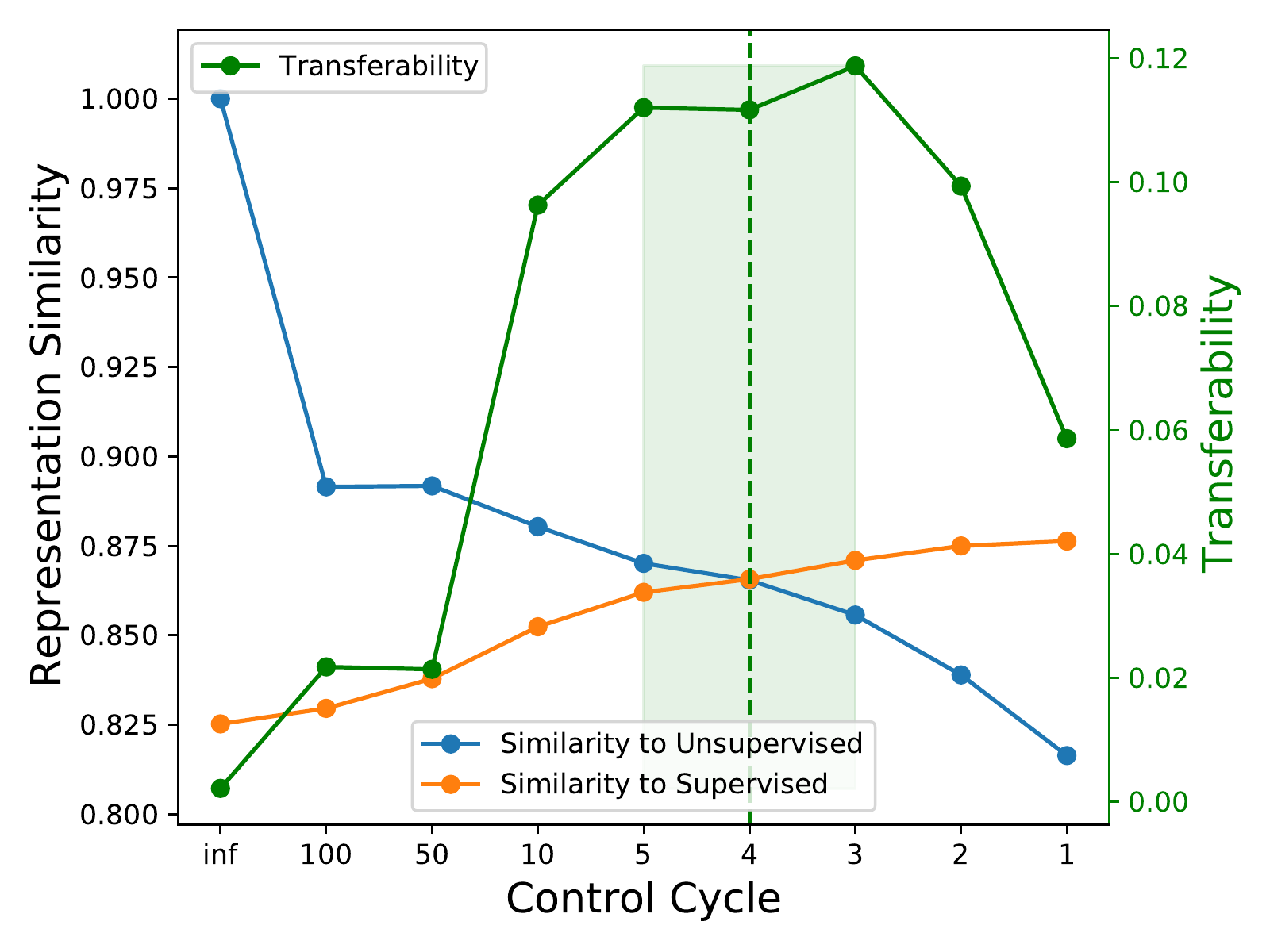}
\end{subfigure}
\begin{subfigure}{.49\textwidth}
  \centering
    \includegraphics[width=0.98\textwidth, height=!]{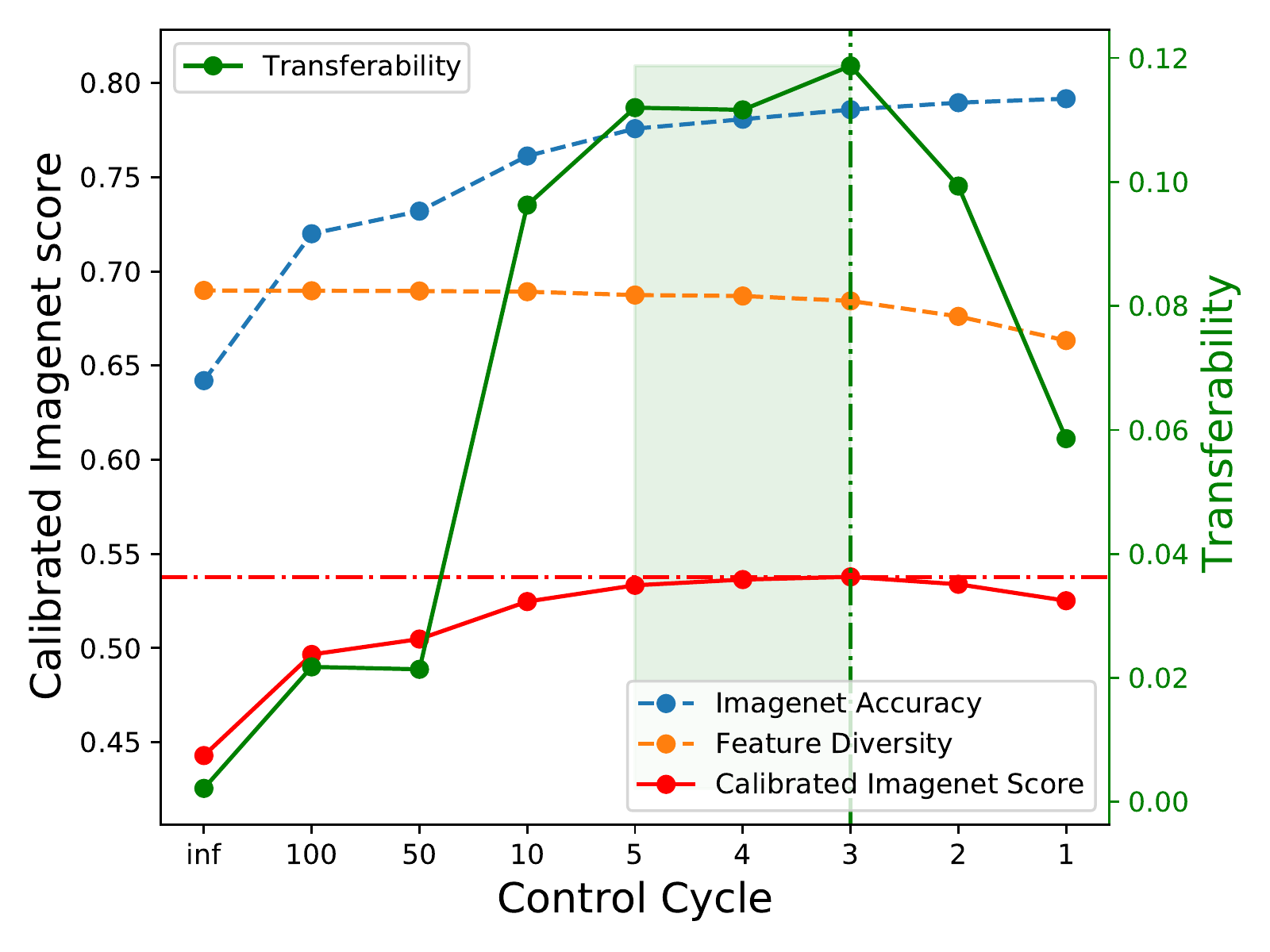}
\end{subfigure}
\vspace{-3mm}
\caption{The impact of controlling label injection. (Left) Representation similarity measured by CKA between models with label injected of different control cycle values and fully supervised and SSL (SwAV) models. The best transferability is obtained when the similarity to supervised and unsupervised models is balanced in shaded area. (Right) Feature Diversity, Imagenet accuracy and the resulted Calibrated Imagenet Score are plotted for the same models. Label injection improves Imagenet accuracy, but when too aggressive it can ruin Feature Diversity. The point of the best transferability is obtained when the Calibrated Imagenet Score is the highst in the same shaded area.}
\label{fig:label_injection}
\vspace{-5mm}
\end{figure}

Figure~\ref{fig:label_injection} shows the impact of the control cycle. Starting from a SSL pretrained model of high feature diversity (SwAV), we apply Algorithm~\ref{alg:controlled_label_injection} with different control cycle values. 
At the left side we show the similarity between the learnt representations of the final model and: (i) the original SSL model, and (ii) a fully supervised model. The similarity is measured by the average Centered Kernel Alignment (CKA)~\cite{kornblith2019similarity} between all pairs of stages of two Resnet50 models. When the label injection is high (low control cycle $\mathcal{T}$) the similarity to the initial SSL model is low and the similarity to a fully supervised model is high. Our experiments show that the best transferability is obtained when those similarities are similar. 
At the right side of the figure, we empirically validate that the proposed label injection scheme improves Imagenet accuracy while maintaining most of the feature diversity of the input SSL model (SwAV). This is true up to a certain tipping point ($\mathcal{T}=3$ in this case), where the tranferability is the highest and right after the aggressive label injection ruins the initial feature diversity and thus the Calibrated Imagenet Score drops together with the transferability.

\begin{figure}[htb]
\centering
\begin{subfigure}{.32\textwidth}
  \centering
    \includegraphics[width=0.98\textwidth, height=!]{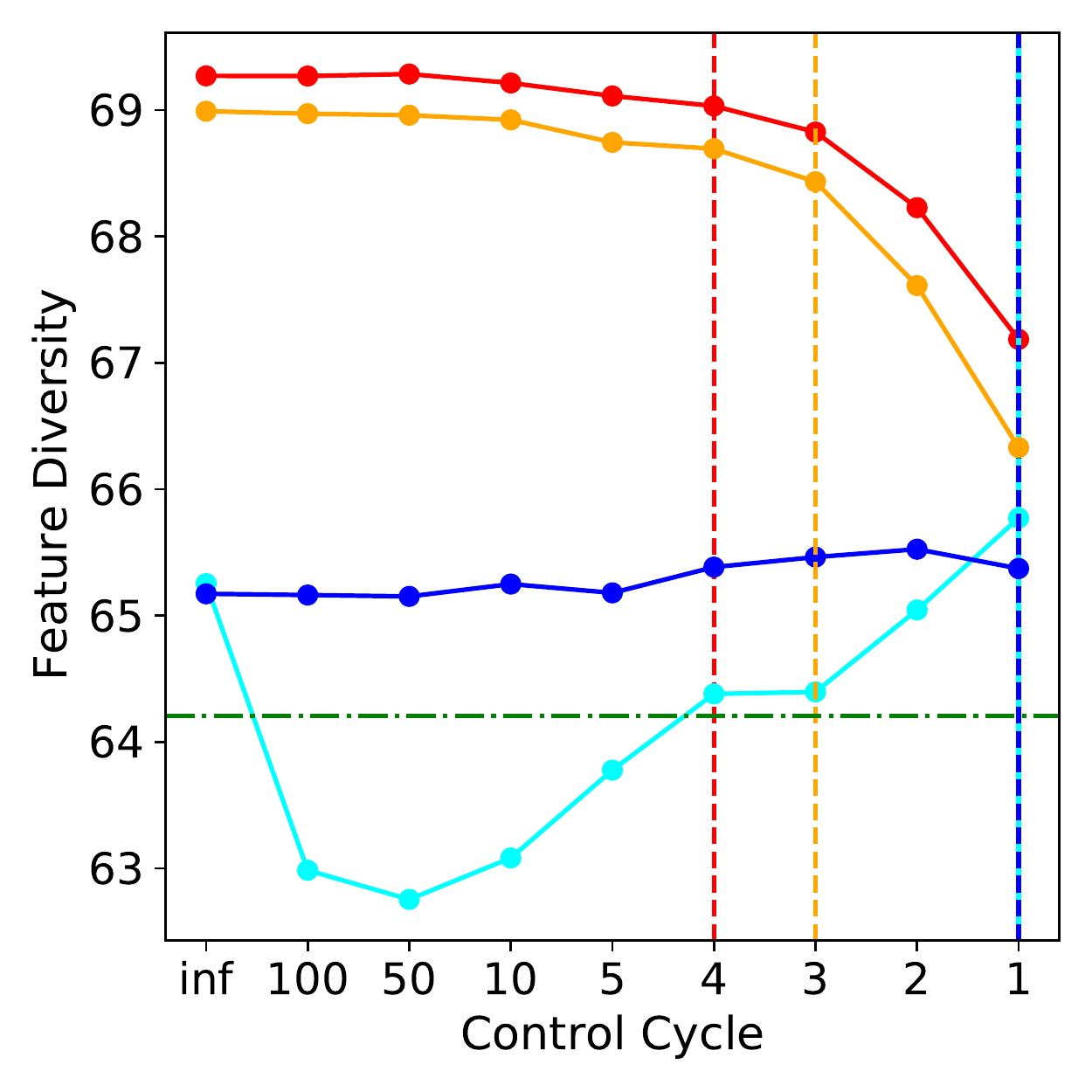}
\end{subfigure}
\begin{subfigure}{.32\textwidth}
  \centering
    \includegraphics[width=0.98\textwidth, height=!]{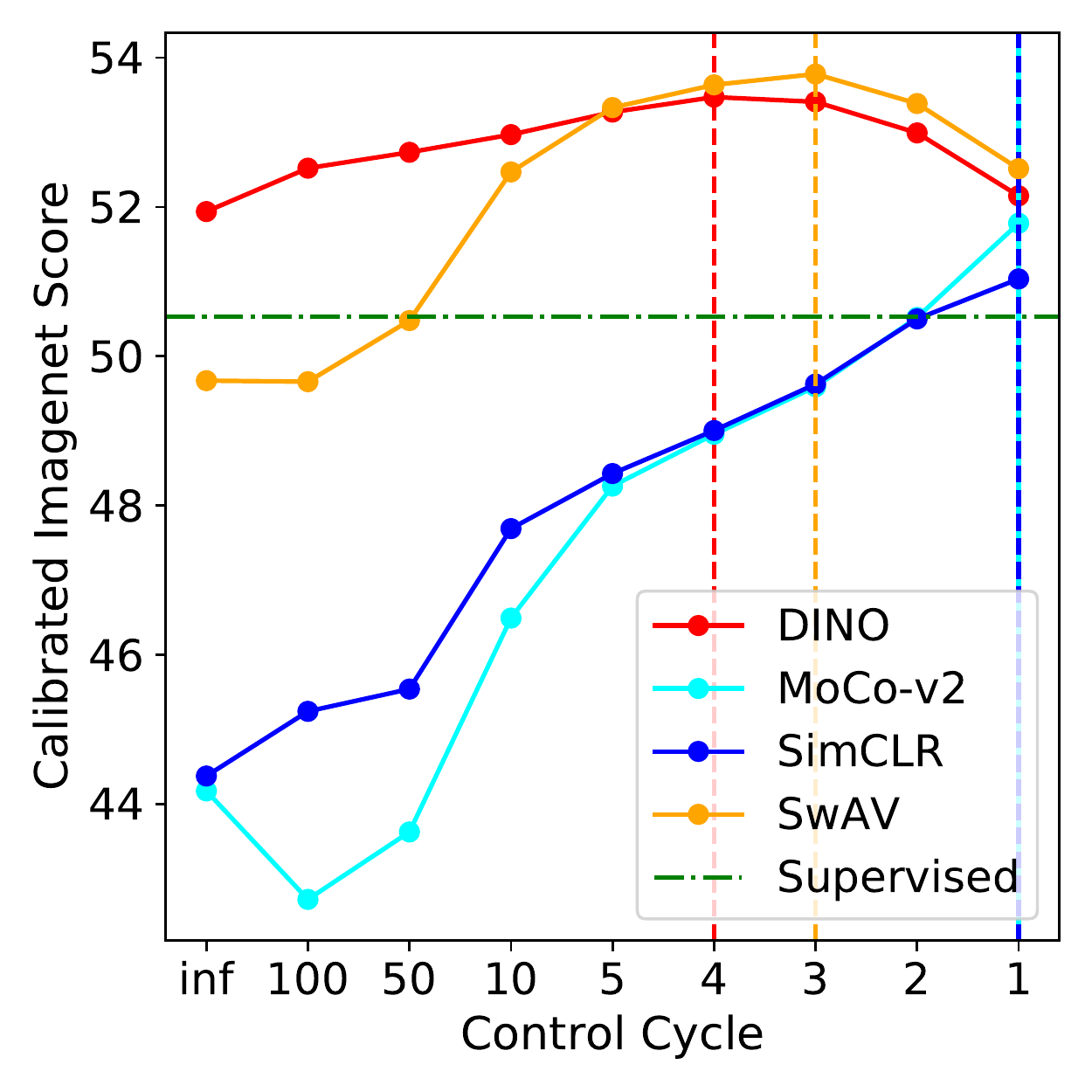}
\end{subfigure}
\begin{subfigure}{.32\textwidth}
  \centering
    \includegraphics[width=0.98\textwidth, height=!]{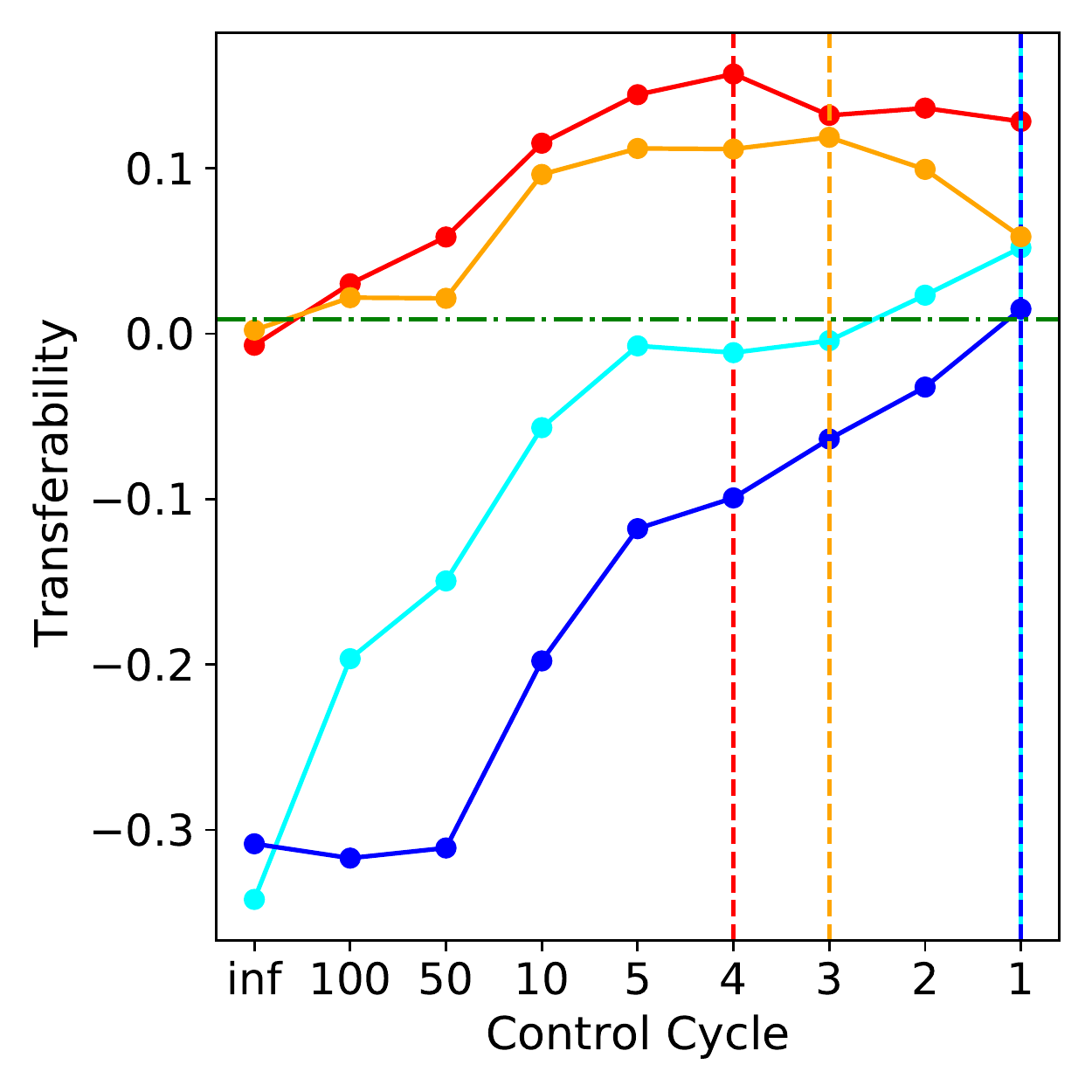}
\end{subfigure}
\caption{Demonstrating the effect of the Controlled Label Injection (CLI) on the Feature Diversity (Left) Calibrated Imagenet Score (Middle) and Transferability (Right) on different SSL pre-trained models. SSL models of low feature diversity benefit from the maximal label injection, while the feature diversity of highly diverse models is to be maintained. The points of highest transferability are aligned with those of the highest CIS (vertical lines).}
\label{fig:tau_to_transferability}
\vspace{-5mm}
\end{figure}


Figure~\ref{fig:imgnt_div_cor_cnn} shows how the control label injection can start off from different SSL pre-trained models of both low (MoCo-v2) and high (SwAV) feature diversity and generate models of different levels of Imagenet accuracy and feature diversity for different control cycle values. Those generated models allow us to make observations about the connection between Imagenet Score and Feature Diversity to the transferability through the Calibrated Imagenet Score, as shown in Figure~\ref{fig:tl_correlation}. The trajectory for every origin SSL model traverses the Calibrated Imagenet Score contour lines towards more transferable regions, as expressed by the size of the circles. 

 This is further shown for more SSL methods applied on CNN in Figure~\ref{fig:tau_to_transferability}, where the connection between the control cycle, Feature Diversity, CIS and transferability is shown. Noteably, SSL models of low feature diversity benefit from the maximal label injection, while the feature diversity of highly diverse models is to be maintained by strenghtning label injection for increasing the Imagenet accuracy right up to the point where the diversity drops. Those control cycle values are aligned with the points of highest CIS and ultimately highest transferability. Besides, SSL method undergoing CLI, two supervised models are presented, the first is trained shortly for 200 epochs and the second is trained longer for 600 epochs. Longer supervised training increases both Imagenet accuracy and Feature Diversity.

%% file: method/algo.tex
\begin{algorithm}[H]
   \captionof{algorithm}{Controlled Label Injection (CLI)}
   \label{alg:controlled_label_injection}
\begin{algorithmic}[1]
\INPUT Self-supervided pretrained weights $W_B$, \\ 
Upstream train set $\mathcal{D}_{train}$,  Control cycle $\mathcal{T}$ \\
Fine-tuning steps $T$, Learning rate $\eta$
\STATE $W_{FF}\leftarrow$RandomInit()
\FOR{$t=1,\dots,T$}
\STATE Sample an i.i.d train batch $(x_t, y_t)\sim\mathcal{D}_{train}$
\STATE $W_{FF}
\leftarrow W_{FF}-\eta\nabla_{W_{FF}}\mathcal{L}_{CE}\left(g_{W_{FF}}(f_{W_B}(x_t)), y_t\right)$
\IF{mod($t$, $\mathcal{T}$)==0}
\STATE $W_{B}\leftarrow W_{B}-\eta\nabla_{W_B}\mathcal{L}_{CE}\left(g_{W_{FF}}(f_{W_B}(x_t)), y_t\right)$
\ENDIF
\ENDFOR
\OUTPUT $W_B$
\end{algorithmic}
\end{algorithm}

%% file: experiments/exp.tex
\section{Experimental Settings}
\label{sec:exp}
\input{experiments/datasets}
\input{experiments/tl_score}
\vspace{-3mm}
\subsection{Training and Fine-tuning Settings}
Our pipeline consists of three stages we shall refer to as Pretrain, Label Injection and Transfer Learning. The Pretrain may be any SSL model, while we experimant with MoCo-v2, SwAV, SimCLR, DINO and MAE. This stage may be completely skipped for standard ImageNet Supervised Training. During Label Injection we train the initialized model in an augmented supervised manner on ImageNet while preserving label diversity (see Section \ref{sec:method}). Finally we evaluate performance of the proposed networks on downstream tasks by training the entire network in a standard supervised manner with identical hyper-parameters for all datasets.

\vspace{-3mm}
\subsubsection{Label Injection}
We train Resnet50 following \cite{rw2019timm} on the full ImageNet~\cite{krizhevsky2012imagenet} training set  using a NVIDIA 8$\times$V100 GPU cluster, RandAugment~\cite{cubuk2020randaugment}, batch size $200$, Cosine learning rate schedule~\cite{loshchilov2016sgdr} with one cycle, $3$ warmup epochs with lr=$1e^{-4}$, initial learning rate $1e^{-1}$, using SGD optimizer~\cite{SGD} with weight decay of $1e^{-4}$ and model EMA for 100 epochs unless otherwise stated (e.g Vanilla from scratch). Fully connected layer is initialized using Linear Probe or from class centroids. Label injection control cycle $\mathcal{T} \in [1,2,3,4,5,10,50,100]$ where $\mathcal{T}=1$ is effectively a standard fine-tuning. ImageNet top-1 accuracy is reported.

ViT models follow the ViT-B architecture and finetuning scripts from \cite{MaskedAutoencoders2021} with batch size $32$, AdamW optimizer~\cite{loshchilov2017decoupled} with base learning rate $5e^{-4}$, layer decay $0.65$, weight decay $0.05$, drop path $0.1$, input resolution  $224$, mixup~\cite{zhang2018mixup} $0.8$, cutmix~\cite{yun2019cutmix} $1.0$ and cutout~\cite{devries2017cutout} $0.25$ for 100 epochs. We do not use gradient accumulation effectively reducing out batch size.

\subsubsection{Fine-tuning}
The same fine-tuning settings is used for all downstream datasets, without a designated hyperparameter search for each one. The single hyperparameter set used for all fine-tuning used Adam with Momentum 0.9, Weight decay with a $1e^{-4}$ coefficient, single cycle cosine learning rate schedule, AutoAugment~\cite{cubuk2018autoaugment}, batch size of $128$, cutout with length $0.5$, model exponent moving average (EMA) and label smoothing~\cite{label_smoothing} with a $0.1$ coefficient for 60 epochs on a single NVIDIA V100 GPU. The initialization of the fully connected is detailed in Appendix~\ref{apdx:fc_init}.

For ViT fine-tuning we use the same scripts and parameters used for label injection except increase in gradient accumulation to $4$ which increases the effective batch size to $128$.

\subsection{Comparison to Other Supervised and SSL Combining Methods}\label{sec:comparison}
\subsubsection{Improved Transferability for CNN. }
Table~\ref{tab:cnn_wide_linprob} and  Table~\ref{tab:cnn_wide} compare the transfer learning performance of Resnet50 pretrained models across 15 downstream tasks, specified in section~\ref{tab:datasets}, and averaged according to section~\ref{sec:tl_score} for linear probing and finetuning respectively. The compared models include pure supervised and unsupervised (MoCO-v2, SwAV, SimCLR, DINO) learning, supervised constrastive learning (SupCon~\cite{khosla2020supervised}), a pretraining combining supervised and self-supervised losses (CE+SelfSupCon~\cite{islam2021broad}) and label injected models following Algorithm~\ref{alg:controlled_label_injection}.
The behaviour for linear probing and finetuning is similar. Specifically, certain label injected models transfer best.

\input{experiments/single_label_res_linprob}
\input{experiments/single_label_res}

As SwAV and DINO benefit from high feature diversity (see Figure~\ref{fig:ssl_high_diversity}), once label injected to the point of high Calibrated Imagenet score (see Figure~\ref{fig:tau_to_transferability}) it transfers better than all the rest both in terms of overall transferability score and for the most downstream datasets individually. Since MoCo-v2 and SimCLR have relatively low feature diversity (see Figure~\ref{fig:ssl_high_diversity}), those benefit from more label injection that increases both their Imagenet accuracy and feature diversity together, as shown in Figure~\ref{fig:tau_to_transferability}. Indeed, those attain the best transferability at $\mathcal{T}=1$. Those observations call for further future research about the underlying mechanisms that make different SSL methods resulting in different levels of feature diversity, as discussed in section~\ref{sec:discussion}.

\subsubsection{Improved Transferability for ViT. }
\input{experiments/single_label_vit}
Similar results are shown for vision transformers (ViT) in Table~\ref{tab:vit_wide} and Appendix~\ref{apdx:vit_2d}. Specifically, label injected ViT models obtain better transfer learning performance than their pure SSL counterparts. Notably, for all the SSL methods examined for ViT, the maximal label injection strength results in the best transferability. Interstingly, this is also true for DINO applied to ViT, when this is not true when applied to CNNs. This observation invites future research on the reasons why ViT tend to learn less diverse features than CNNs when pre-trained with SSL method, see section~\ref{sec:discussion} for a further discussion.
\input{experiments/feature_importance}

%% file: experiments/datasets.tex
\vspace{-2mm}
\subsection{Downstream Datasets}
We evaluated models for multi-label image classification on the popular MS-COCO~\cite{lin2014microsoft} dataset and another $14$ single-label image classification datasets ranging in training set size from 2,040 to 75,750 images (20 to 5,000 images per class; Table~\ref{tab:datasets}). These datasets covered a wide range of image classification tasks, including superordinate-level object classification (CIFAR-10~\cite{krizhevsky2009learning}, CIFAR-100~\cite{krizhevsky2009learning}, Caltech-256~\cite{griffin2007caltech256}); fine-grained object classification of different kinds (Food-101~\cite{bossard2014food}, NABirds~\cite{van2015nabirds}, Stanford Cars~\cite{krause2013collecting}, FGVC Aircraft~\cite{maji13fine-grained}, OxfordIIIT Pets~\cite{parkhi2012cats}, Oxford Flowers-102~\cite{Nilsback08}, Stanford Dogs~\cite{khosla2011novel}, CUB-200~\cite{WahCUB_200_2011}); texture classification (DTD~\cite{cimpoi2014describing}); and scene classification (MIT indoor 67~\cite{quattoni2009recognizing}, SUN397~\cite{xiao2010sun}).

\input{experiments/dataset_table}

%% file: experiments/dataset_table.tex
\begin{table}[htb]
    \centering
    \begin{tabular}{|c|l|c|c|c|c|}
    \hline
    Category & Name & Symbol & Classes & Train Size & Test Size \\
    \toprule
    Upstream & Imagenet~\cite{krizhevsky2012imagenet} & ImNet & 1000 & 1,281,167 & 100,000 \\
    \toprule
    Superordinate-
    & CIFAR-10~\cite{krizhevsky2009learning} & CIFAR10 & 10 & 50,000 & 10,000 \\
    level object
    & CIFAR-100~\cite{krizhevsky2009learning} & CIFAR100 & 100 & 50,000 & 10,000 \\
    classification
    & Caltech-256~\cite{griffin2007caltech256} & Caltech & 256 & 24,581 & 6,026 \\
    \hline
    & Food-101~\cite{bossard2014food} & Food & 101 & 80,800 & 20,200 \\
    & NABirds~\cite{van2015nabirds} & Birds & 555 & 24,615 & 23,912 \\
    & Stanford Cars~\cite{krause2013collecting} & Cars & 196 & 8,041 & 8,144 \\
    Fine-grained
    & FGVC Aircraft~\cite{maji13fine-grained} & Aircraft & 100 & 3,334 & 3,333 \\
    object
    &  OxfordIIIT Pets~\cite{parkhi2012cats} & Pets & 37 & 3,680 & 3,669 \\
    classification
    & Oxford Flowers-102~\cite{Nilsback08} & Flowers & 102 & 1,020 & 6,149 \\
    & Standord Dogs~\cite{khosla2011novel} & Dogs & 120 & 12,000 & 8,580 \\
    & CUB-200~\cite{WahCUB_200_2011} & CUB & 200 & 5,994 & 5,794   \\
    \hline
    Texture
    & DTD~\cite{cimpoi2014describing} & DTD & 47 & 1,880 & 1,880 \\
    \hline
    Scene
    & MIT indoor 67~\cite{quattoni2009recognizing} & Indoor & 67 & 5360 & 1,340 \\
    classification
    & SUN397~\cite{xiao2010sun} & SUN & 397 & 19,850 & 19,850 \\
    \hline
    Multi-label & MS-COCO~\cite{lin2014microsoft} & COCO & 80 & 82,081 & 40,137 \\ 
    \hline
    \end{tabular}
    \caption{Datasets examined in transfer learning}
    \label{tab:datasets}
    \vspace{-10mm}
\end{table}

%% file: experiments/tl_score.tex
\vspace{-3mm}
\subsection{Measuring Transferability}
\label{sec:tl_score}

Much of the analysis in this work requires comparing accuracies across datasets of differing difficulty. Directly comparing the top-1 accuracy across datasets is problematic (e.g., as done in \cite{khosla2020supervised,islam2021broad}).
The meaning of a $1\%$ additive increase in accuracy is different if it is relative to a base accuracy of $50\%$ vs. $99\%$.
Instead, we follow the transferability evaluation protocol proposed by \cite{kornblith2019better} and consider the log odds, i.e., the accuracy after the $logit$ transformation $logit(p)=log(p/(1-p))=sigmoid^{-1}(p)$. We repeat the details here for completeness: 
The logit transformation is the most commonly used transformation for analysis of proportion data, and an additive change $\Delta$ in logit-transformed accuracy has a simple interpretation as a multiplicative change $e^\Delta$ in the odds of correct classification.
Given logit-transformed accuracies $y_m^d$ of model $m\in\mathcal{M}$ on dataset $d \in \mathcal{D}$, we compute adjusted accuracies $acc(m, d) = y_m^d - \sum_{m'\in\mathcal{M}}y_{m'}^d/|\mathcal{M}|$.
For each model, we take the mean and standard error of the adjusted accuracy across datasets, and multiply the latter by a correction factor $|\mathcal{M}|/(|\mathcal{M}|-1)$.

%% file: experiments/single_label_res_linprob.tex
\begin{table}[htb]
\centering
\tiny
\begin{tabular}{l|c|cccccccccccccc|c}
\toprule
Pretrain &  
\rotatebox[origin=c]{-70}{ImNet} &  
\rotatebox[origin=c]{-70}{Aircraft}&  
\rotatebox[origin=c]{-70}{Birds}   &
\rotatebox[origin=c]{-70}{CIFAR10}    & 
\rotatebox[origin=c]{-70}{CIFAR100}&  
\rotatebox[origin=c]{-70}{CUB}   &
\rotatebox[origin=c]{-70}{Caltech}     &
\rotatebox[origin=c]{-70}{Cars}   &
\rotatebox[origin=c]{-70}{DTD}   &
\rotatebox[origin=c]{-70}{Dogs}   &
\rotatebox[origin=c]{-70}{Flowers} &
\rotatebox[origin=c]{-70}{Food} &
\rotatebox[origin=c]{-70}{Indoor}    &
\rotatebox[origin=c]{-70}{Pets}   &
\rotatebox[origin=c]{-70}{SUN}   &
\rotatebox[origin=c]{-70}{Transfer}\\
\midrule
Supervised           &          \underline{78.7} &      46.4 &   \underline{60.9} &     93.0 &      77.1 & \underline{71.5} &     \underline{89.1} &  67.4 & 69.5 &  90.4 &     86.5 &  \underline{70.0} &    78.7 &  \underline{93.0} & 63.1 &       7.3 \\
SupCon~\cite{khosla2020supervised} & 77.3 &  \underline{50.9} &   56.6 &     \underline{94.9} &     \underline{79.2} & 69.0 &     88.5 &  \underline{72.6} & \underline{70.2} &  \underline{90.5} &     \underline{89.1} &  68.6 &    \underline{79.3} &  92.5 & \underline{63.6} &   \underline{12.6}
\\ \hline
CE + SelfSupCon~\cite{islam2021broad} & 77.3 &      40.2 &   52.8 &     93.3 &      \underline{76.5} & 63.0 &     \underline{87.3} &  58.1 & 67.6 &  \underline{\textbf{94.0}} &     \underline{85.6} &  67.4 &    76.6 &  92.6 & \underline{61.3} &  \underline{-3.9} \\
MoCo-v2                        &           61.9 &      \underline{43.9} &   38.9 &     \underline{93.4} &      76.4 & 53.8 &     83.5 &  \underline{59.3} & \underline{69.7} &  68.0 &     85.3 &  \underline{68.5} &    76.0 &  84.6 & 60.6 &     -31.1  \\ 
MoCo-v2 ($\mathcal{T} = 1$)    &           \underline{78.7} &      35.8 &   55.9 &     92.4 &      74.6 & 65.6 &     \underline{87.3} &  52.7 & 67.8 &  91.6 &     80.6 &  65.4 &    76.3 &  \underline{93.1} & 60.5 &     -12.4 \\
MoCo-v2 ($\mathcal{T} = 4$)    &           76.0 &      41.3 &   \underline{57.3} &     92.2 &      74.4 & \underline{66.2} &     87.0 &  57.0 & 66.9 &  87.5 &     83.5 &  67.2 &    \underline{76.9} &  91.9 & 60.8 &     -11.8 \\ \hline
SwAV                         &           72.0 &      52.0 &   53.3 &     93.2 &      77.8 & 66.7 &     86.5 &  \underline{71.0} & 71.4 &  76.4 &     \underline{90.6} &  \underline{73.2} &    \underline{81.6} &  88.9 & 65.1 &       0.3  \\ 
SwAV ($\mathcal{T} = 1$)     &           \underline{\textbf{79.2}} &      44.0 &   62.8 &     \underline{93.3} &      77.4 & 71.2 &     89.2 &  64.2 & 70.7 &  \underline{91.1} &     86.6 &  70.2 &    80.1 &  \underline{93.3} & 63.5 &       8.9 \\
SwAV ($\mathcal{T} = 4$)     &           78.1 &      \underline{53.4} &   \underline{65.7} &     93.1 &      \underline{78.8} & \underline{73.2} &     \underline{\textbf{89.8}} &  69.8 & \underline{72.2} &  87.0 &     90.4 &  \underline{73.2} &    \underline{81.6} &  93.1 & \underline{65.8} &      \underline{18.1} \\ \hline
DINO                         &           75.0 &      \underline{\textbf{54.8}} &   54.8 &     93.7 &      78.6 & 68.9 &     87.1 &  \underline{74.5} & 72.7 &  75.9 &     \underline{\textbf{92.5}} &  74.7 &    81.7 &  89.3 & \underline{\textbf{66.1}} &       8.1 \\ 
DINO ($\mathcal{T} = 1$)     &           \underline{77.6} &      46.0 &   63.4 &     93.5 &      78.3 & 73.2 &     89.2 &  66.2 & 71.0 &  \underline{90.9} &     87.0 &  71.2 &    80.9 &  \underline{93.8} & 64.2 &      13.1 \\
DINO ($\mathcal{T} = 4$)     &           77.5 &      53.8 &   \underline{\textbf{66.0}} &     \underline{\textbf{93.8}} &      \underline{\textbf{79.5}} & \underline{\textbf{74.3}} &     \underline{89.7} &  70.5 & \underline{\textbf{73.0}} &  86.9 &     91.8 &  \underline{\textbf{74.8}} &    \underline{\textbf{82.2}} &  93.3 & 66.0 &      \underline{\textbf{22.6}} \\ \hline
SimCLR                       &           68.1 &      43.4 &   35.3 &     89.1 &      69.0 & 50.5 &     82.4 &  56.2 & 65.4 &  65.4 &     85.2 &  62.2 &    72.4 &  83.8 & 58.2 &     -48.0  \\ 
SimCLR ($\mathcal{T} = 1$)   &           \underline{78.1} &      47.8 &   \underline{61.1} &     \underline{\textbf{93.8}} &      \underline{77.7} & \underline{71.3} &     \underline{88.7} & 65.8 & \underline{70.9} &  \underline{88.9} &     87.5 &  \underline{70.5} &    \underline{80.2} &  \underline{92.9} & \underline{64.0} &       \underline{8.9} \\
SimCLR ($\mathcal{T} = 4$)   &           75.0 &      \underline{50.4} &   55.3 &     93.7 &      77.6 & 67.5 &     87.5 &  \underline{66.8} & 69.8 &  82.4 &     \underline{88.0} &  69.2 &    77.0 &  91.3 & 63.0 &      -1.5 \\
\bottomrule
\end{tabular}
\caption{Linear probing performance of different \textbf{CNN} models, including different levels of label injected models) fit on the downstream datasets in terms of top-1 accuracy (\%) and the overall transferability score. The models are grouped by the underlying base SSL method. The best performance of each column appears in \textbf{bold} and the best in each group is \underline{underlined}. Label injected models transfer best.} 
\label{tab:cnn_wide_linprob}
\end{table}

%% file: experiments/single_label_res.tex
\begin{table}[htb]
\centering
\tiny

\begin{tabular}{l|c|cccccccccccccc|c||c}
\toprule
Pretrain & \rotatebox[origin=c]{-70}{ImNet} 
& \rotatebox[origin=c]{-70}{Caltech} 
& \rotatebox[origin=c]{-70}{CIFAR10} 
& \rotatebox[origin=c]{-70}{CIFAR100} 
& \rotatebox[origin=c]{-70}{CUB} 
& \rotatebox[origin=c]{-70}{DTD} 
& \rotatebox[origin=c]{-70}{Aircraft} 
& \rotatebox[origin=c]{-70}{Food} 
& \rotatebox[origin=c]{-70}{Indoor} 
& \rotatebox[origin=c]{-70}{Birds} 
& \rotatebox[origin=c]{-70}{Flowers} 
& \rotatebox[origin=c]{-70}{Pets} 
& \rotatebox[origin=c]{-70}{Cars} 
& \rotatebox[origin=c]{-70}{Dogs} 
& \rotatebox[origin=c]{-70}{SUN} 
& \rotatebox[origin=c]{-70}{Transfer} 
& \rotatebox[origin=c]{-70}{COCO} \\
\toprule
Supervised & \underline{78.7} & \underline{86.9} & \underline{97.6} & \underline{86.0} & 85.9 & 69.3 & 82.0 & 85.3 & 81.3 & \underline{74.9} & \underline{99.1} & \underline{92.4} & 94.4 & 82.9 & \underline{65.3} & \underline{0.012} & \underline{81.9} \\   
SupCon~\cite{khosla2020supervised} & 77.3 & 86.3 & 97.6 & 85.7 & 85.0 & 69.8 & \underline{\textbf{84.1}} & \underline{86.1} & \underline{81.4} & 73.1 & 99.0 & 91.9 & \underline{94.7} & \underline{83} & 65.2 & \underline{0.007} & 81.2 \\   
\hline
                                    
CE+SelfSupCon & 76.4 & 86.3 & 97.6 & 85.8 & 85.9 & 69.5 & \underline{83.3} & \underline{86.3} & 80.9 & \underline{74.1} & 98.7 & 92.4 & \underline{\textbf{94.8}} & \underline{\textbf{85.3}} & 64.4 &  0.003 & \underline{82.0} \\
MoCo-v2 & 61.9 & 78.1 & 96.7 & 82.0 & 79.4 & 66.3 & 80.1 & 85.4 & 75.6 & 61.1 & 98.5 & 86.8 & 93.5 & 73.4 & 56.3 & -0.352 & 78.7 \\   
MoCo-v2 ($\mathcal{T}=1$) & \underline{78.7} & \underline{87.3} & \underline{97.6} & \underline{86.3} & \underline{85.9} & \underline{70} & 83.0 & 86.1 & 82.2 & 73.9 & \underline{99.2} & \underline{\textbf{92.8}} & 94.4 & 84.4 & \underline{65.1} & \underline{0.054} & 81.9 \\   
MoCo-v2 ($\mathcal{T}=4$) & 76.0 & 86.4 & 97.6 & 86.1 & 85.7 & 69.7 & 82.8 & 86.1 & \underline{82.3} & 73.3 & 98.9 & 92.0 & 94.3 & 81.5 & 64.8 & -0.010 & 81.7 \\   
\hline
                                    
SwAV & 64.2 & 87.0 & 97.8 & 86.6 & 84.3 & 72.1 & 82.3 & 87.4 & 83.1 & 75.1 & 98.9 & 90.3 & 93.6 & 80.6 & \underline{\textbf{67.8}} & 0.005 & 82.8 \\   
SwAV ($\mathcal{T}=1$) & \underline{\textbf{79.2}} & 87.5 & 97.7 & 86.5 & 86.5 & 71.3 & \underline{83.4} & 86.8 & 82.3 & 75.8 & 99.0 & \underline{92.5} & 94.6 & \underline{83.6} & 66.4 & 0.062 & 82.3 \\ 
SwAV ($\mathcal{T}=4$) & 78.1 & \underline{\textbf{88.4}} & \underline{97.8} & \underline{87.1} & \underline{\textbf{86.6}} & \underline{\textbf{72.9}} & 82.8 & \underline{87.6} & \underline{\textbf{84.4}} & \underline{76.1} & \underline{99.3} & 91.9 & \underline{94.6} & 82.1 & 67.8 & \underline{0.118} & \underline{\textbf{83.2}} \\   
\hline
                                    
DINO & 75.0 & 87.2 & \underline{97.8} & 86.9 & 83.7 & 72.1 & 80.6 & 87.5 & 83.2 & 74.5 & 98.7 & 89.6 & 93.8 & 80.1 & 67.6 & -0.024 & 82.7 \\   
DINO ($\mathcal{T}=1$) & \underline{77.6} & 87.4 & 97.7 & 86.7 & 86.3 & 71.1 & \underline{83} & 87.1 & 82.7 & 76.2 & 99.4 & \underline{92.5} & 94.7 & \underline{82.9} & 66.2 & 0.095 & 82.3 \\   
DINO ($\mathcal{T}=4$) & 77.5 & \underline{88.2} & 97.8 & \underline{\textbf{87.5}} & \underline{86.5} & \underline{72.1} & 82.8 & \underline{\textbf{87.6}} & \underline{84.1} & \underline{\textbf{76.5}} & \underline{\textbf{99.4}} & 92.1 & \underline{94.7} & 82.1 & \underline{67.6} & \underline{\textbf{0.126}} & \underline{83.1} \\   
\hline
                                    
SimCLR & 68.1 & 85.4 & \underline{\textbf{97.9}} & 86.3 & 78.3 & 65.9 & 77.2 & 84.0 & 75.4 & 62.6 & 97.6 & 86.6 & 91.6 & 73.8 & 62.9 & -0.318 & 80.5 \\   
SimCLR ($\mathcal{T}=1$) & \underline{78.1} & 86.3 & 97.6 & 86.3 & \underline{85.6} & \underline{70} & \underline{82.8} & \underline{85.9} & \underline{80.7} & \underline{72.8} & \underline{99.1} & \underline{91.4} & \underline{94.5} & \underline{80.8} & \underline{65.5} & \underline{-0.013} & 81.2 \\   
SimCLR ($\mathcal{T}=4$) & 75.0 & \underline{86.6} & 97.7 & \underline{86.7} & 82.7 & 69.5 & 81.1 & 84.9 & 79.3 & 68.0 & 98.7 & 89.5 & 93.5 & 77.9 & 64.9 & -0.120 & \underline{81.5} \\   
\bottomrule
\end{tabular}
\caption{Performance of different \textbf{CNN} models fine-tuned on the downstream datasets in terms of top-1 accuracy (\%) (averaged over 3 runs) and the overall transferability score. The models are grouped by the underlying base SSL method. The best performance of each column appears in \textbf{bold} and the best in each group is \underline{underlined}. Label injected models transfer best.} 
\label{tab:cnn_wide}
\vspace{-6mm}
\end{table}

\small

%% file: experiments/single_label_vit.tex
\begin{table}[htb]
    \centering
    \tiny
\begin{tabular}{l|r|rrrrrrrrrrrrrr|r}
\toprule
Pretrain & \rotatebox[origin=c]{-70}{ImNet} 
& \rotatebox[origin=c]{-70}{Caltech} 
& \rotatebox[origin=c]{-70}{CIFAR10} 
& \rotatebox[origin=c]{-70}{CIFAR100} 
& \rotatebox[origin=c]{-70}{CUB} 
& \rotatebox[origin=c]{-70}{DTD} 
& \rotatebox[origin=c]{-70}{Aircraft} 
& \rotatebox[origin=c]{-70}{Food} 
& \rotatebox[origin=c]{-70}{Indoor} 
& \rotatebox[origin=c]{-70}{Birds} 
& \rotatebox[origin=c]{-70}{Flowers} 
& \rotatebox[origin=c]{-70}{Pets} 
& \rotatebox[origin=c]{-70}{Cars} 
& \rotatebox[origin=c]{-70}{Dogs} 
& \rotatebox[origin=c]{-70}{SUN} 
& \rotatebox[origin=c]{-70}{Transfer}  \\
\midrule
Supervised & 81.0 & 90.9 & 98.6 & 89.6 & 82.3 & 70.8 & 60.8 & 87.9 & 83.2 & 79.6 & 91.3 & 93.8 & 85.4 & \textbf{91.5} & 68.1 & -0.101 \\
\hline
MAE & 68.0 & 89.2 & 98.1 & 86.9 & 75.4 & 68.5 & 53.8 & 88.8 & 82.2 & 81.2 & 70.6 & 91.2 & 79.6 & 83.2 & 67.6 & -0.425 \\
MAE ($\mathcal{T}=1$) & \underline{\textbf{83.4}} & \underline{93.0} & 98.8 & \underline{90.6} & 84.3 & \underline{73.7} & \underline{73.1} & \underline{\textbf{90.6}} & \underline{85.4} & \underline{\textbf{86.2}} & 94.4 & \underline{\textbf{94.8}} & \underline{89.8} & \underline{89.5} & \underline{71.1} & \underline{0.131} \\
MAE ($\mathcal{T}=4$) & 81.3 & 92.2 & \underline{98.8} & 90.1 & \underline{84.5} & 73.2 & 72.4 & 90.2 & 85.0 & 85.6 & \underline{94.6} & 94.6 & 89.2 & 88.2 & 71.0 & 0.095 \\
\hline
DINO & 78.2 & 87.2 & 97.8 & 86.9 & 83.7 & 72.1 & \underline{\textbf{80.6}} & 87.5 & 83.2 & 74.5 & \underline{\textbf{98.7}} & 89.6 & \underline{\textbf{93.8}} & 80.1 & 67.6 & -0.187 \\
DINO ($\mathcal{T}=1$)  & \underline{83.2} & \underline{\textbf{93.1}} & \underline{\textbf{99.0}} & \underline{\textbf{91.2}} & 84.7 & 74.6 & 72.1 & \underline{89.9} & \underline{\textbf{86.3}} & 84.9 & 94.7 & 94.3 & 89.4 & \underline{90.5} & \underline{\textbf{71.5}} & \underline{\textbf{0.148}} \\
DINO ($\mathcal{T}=4$) & 82.3 & 92.8 & 98.8 & 91.0 & \underline{\textbf{84.7}} & \underline{\textbf{75.2}} & 71.2 & 90.0 & 85.8 & \underline{85.2} & 95.4 & \underline{94.7} & 89.1 & 88.3 & 71.5 & 0.131 \\

\bottomrule
\end{tabular}
\small
\caption{Performance of different \textbf{ViT} models fine-tuned on the downstream datasets in terms of top-1 accuracy (\%) and the overall transferability score. The models are grouped by the underlying base SSL method. The best performance of each column appears in \textbf{bold} and the best in each group is \underline{underlined}. Label injected models transfer best.} 
\label{tab:vit_wide}
\vspace{-5mm}
\end{table}

%% file: experiments/feature_importance.tex
\section{Feature Importance for Transfer Learning} \label{sec:fesature_importance} 
\begin{figure}[htb]
    \vspace{-5mm}
    \centering
    \includegraphics[width=0.9\textwidth, height=!]{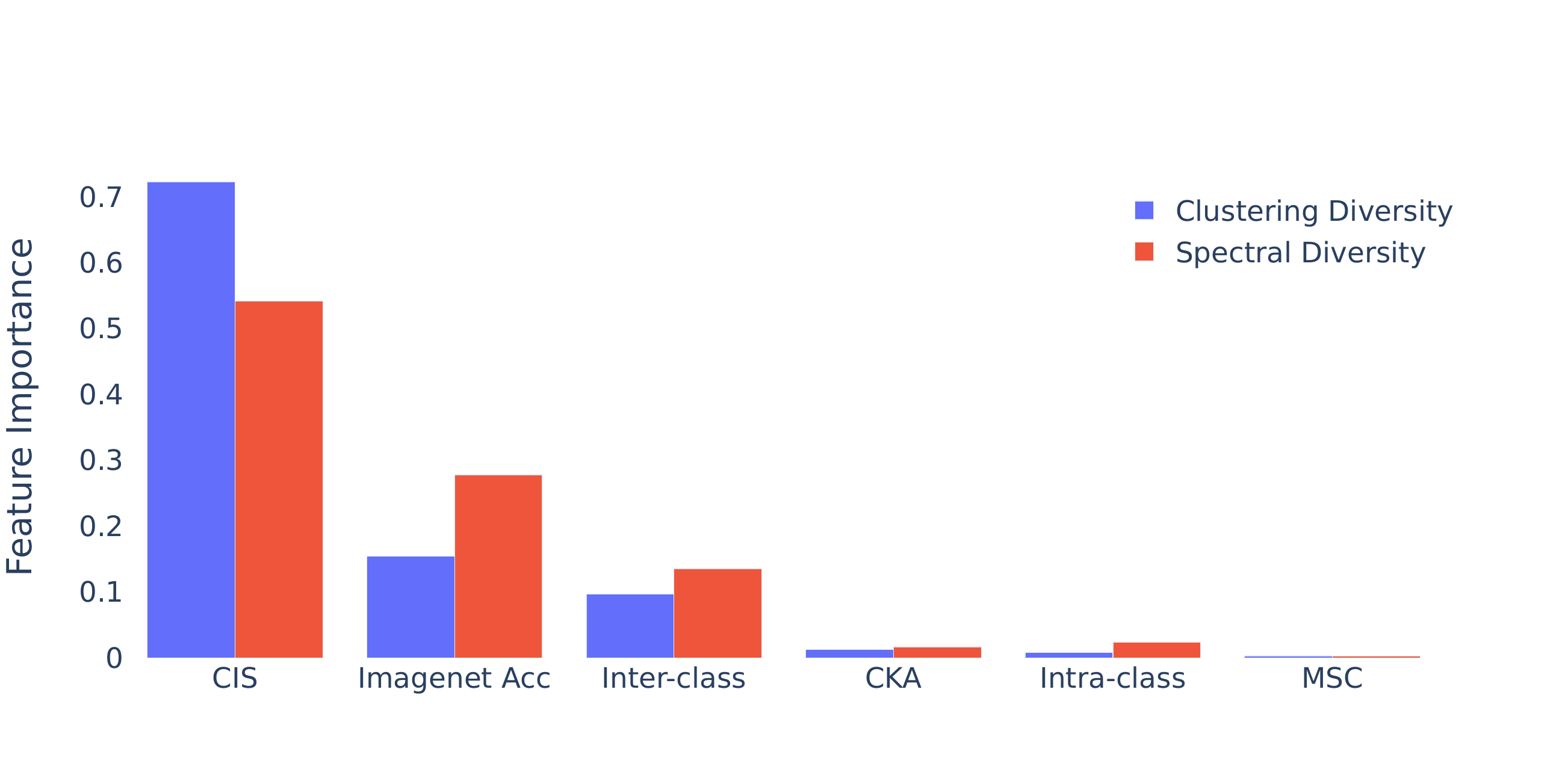}
  \caption{The relative importance of different factors are quantified by the popular feature importance derived from XGBoost. Most of the importance is attributed to the Calibrated Imagenet Score (CIS).}
    \vspace{-5mm}
\label{fig:xgboost}
\end{figure}

In this section we empirically analyze the importance of different factors to transferability. We consider the \textit{Calibrated Imagenet Score} (CIS) as well as CKA, intra-class variance and class separation, that are identified \cite{kornblith2021better,islam2021broad} to imply on transferability. The relative importance of those factors is quantified in Figure~\ref{fig:xgboost} by the ability to predict the transfer learning performance when finetuning by the popular XGBoost~\cite{chen2016xgboost}, that allows feature importance analysis~\cite{zheng2017short}, for more technical details see Appendix~\ref{apdx:xgboost}. It is evident that CIS is highly predictive of the transferability, and specifically significantly more important than the other factors inspected. Notably, both measures of Feature Diversity capture the importance of CIS compared to other factors. This shows that the very notion of feature diversity and the way it calibrates Imagenet accuracy are valid. Similar results are shown in Appendix~\ref{apdx:linprob_conc} for the case of linear probing.

%% file: discussion.tex
\section{Discussion and Future Work}
\label{sec:discussion}
While we showed that models with high diversity transfer better, a natural extension would be to better understand how and why some training methods produce higher diversity than others. 
Indeed \cite{ayinde2019regularizing} uses an explicit diversity regularization. We don't expect a diversity regularization during training to work well since diversity encourages high complexity models. In comparison, standard regularization restrict model complexity as a balance to the models overparamtrization.
 Indeed, there are many ways the network can increase the diversity metric with no real change to the model behavior.
 One such trivial way to increase the Spectral Feature Diversity is to scale each filter $\hat{W_i}=\frac{W_i}{\|W_i\|}$ and then insert $\|W_i\|$ into subsequent BatchNorm.
 Similarly, Cluster Diversity is based on cosine similarity, which is scale agnostic, thus filters that are redundant, or close to zero can be set to orthogonal vectors with epsilon scale.
 We hope this paper motivates future work in ways to increase real filter diversity, and transferability.

%% file: conclusions.tex
\section{Conclusions}
In this paper, we analyse the importance of different properties of pre-trained models to their transferability. We identify the notion of feature diversity as one of the key factors for transferability, together with the performance on the upstream task. A simple fine-tuning procedure is proposed for improving the transferability of given self-supervised pretrained models, by injecting controlled supervision to those, while maintaining their feature diversity and improving their performance on the upstream task. Our study holds for different popular architectures of CNNs and ViTs and self-supervised methods, two different formulations for capturing feature diversity and many downstream tasks of multi-label and single-label classification over more than $15$ different datasets.

%% file: appendix/appendix.tex
\newpage
\begin{center}
    \huge
    Appendix
\end{center}
\appendix

\section{Weight Features of Neural Layers}
\label{sec:layer_features}
In this section we specify the way we extract weight features out of the weight tensors of convolutional \cite{lecun1989backpropagation}, fully-connected and multi-head self-attention (MSA) layers. \cite{vaswani2017attention}, that compose modern Resnets~\cite{he2016deep} and vision transformers (ViT)~\cite{dosovitskiy2021an}.

\subsubsection{Convolutional Layers Features. }
Denote by $W^{(l)}\in\mathbb{R}^{k\times k \times n_l \times n_{l+1}}$ the weights of the $l$-th convolutional layer of kernel size $k$ and $n_l,n_{l+1}$ the number of input and output channels respectively. 
Thus, every weight matrix can be reshaped to $\hat{W}^{(l)}\in\mathbb{R}^{k^2\cdot n_l\times n_{l+1}}$, such that the $i$-th feature of the $l$-th layer, $\mathbf{w}^{(l)}_i\in\mathbb{R}^{k^2\cdot n_l}$ with $i\in\{1,\dots,n_{l+1}\}$, is the $i$-th column of the reshaped weight matrix  $\hat{W}^{(l)}=[\mathbf{w}^{(l)}_1,\dots,\mathbf{w}^{(l)}_{n_{l+1}}]$.

\subsubsection{Fully-connected Layers Features. }
The $l$-th fully-connected layer performs a vector-matrix multiplication with a weight matrix $W^{(l)}\in\mathbb{R}^{w_l\cdot h_l \cdot n_l \times n_{l+1}}$, where $w_l,h_l$ are the spatial width and height of the input tensor. Thus, its $i$-th feature, $\mathbf{w}^{(l)}_i\in\mathbb{R}^{w_l\cdot h_l \cdot n_l}$ with $i\in\{1,\dots,n_{l+1}\}$, is the $i$-th column of the weight matrix  $\hat{W}^{(l)}=[\mathbf{w}^{(l)}_1,\dots,\mathbf{w}^{(l)}_{n_{l+1}}]$.

\subsubsection{Multi-head Self-attention Layers Features. } A multi-head self-attention (MSA) layer \cite{vaswani2017attention} operates directly on its input $x^{(l)}\in\mathbb{R}^{w_l\cdot h_l \cdot n_l}$ by dividing it into $H^{(l)}$ groups of channels $x^{(l)}\in\mathbb{R}^{w_l\cdot h_l \cdot \frac{n_l}{H^{(l)}}}$ and applying direct three vector-matrix multiplications on each group with the corresponding matrices $Q^{(l)}_h,K^{(l)}_h\in\mathbb{R}^{w_l\cdot h_l \cdot \frac{n_l}{H^{(l)}}\times d_{QK}}$ and $V^{(l)}_h\in\mathbb{R}^{w_l\cdot h_l \cdot \frac{n_l}{H^{(l)}}\times d_{V}}$ for $h=1,\dots,H^{(l)}$ and some desing parameters $d_{QK},d_{V}\in\mathbb{N}_+$. 
For each head $h=1,\dots,H^{(l)}$:
\begin{itemize}
    \item The $i$-th feature, $q^{(l)}_{h,i}\in\mathbb{R}^{w_l\cdot h_l \cdot \frac{n_l}{H^{(l)}}}$ with $i\in\{1,\dots,d_{QK}\}$, is the $i$-th column of the weight matrix  $Q^{(l)}_h=[q^{(l)}_{h,1},\dots,q^{(l)}_{h,d_{QK}}]$.
    \item The $i$-th feature, $k^{(l)}_{h,i}\in\mathbb{R}^{w_l\cdot h_l \cdot \frac{n_l}{H^{(l)}}}$ with $i\in\{1,\dots,d_{QK}\}$, is the $i$-th column of the weight matrix  $K^{(l)}_h=[k^{(l)}_{h,1},\dots,k^{(l)}_{h,d_{QK}}]$.
    \item The $i$-th feature, $v^{(l)}_{h,i}\in\mathbb{R}^{w_l\cdot h_l \cdot \frac{n_l}{H^{(l)}}}$ with $i\in\{1,\dots,d_{V}\}$, is the $i$-th column of the weight matrix  $V^{(l)}_h=[q^{(l)}_{h,1},\dots,q^{(l)}_{h,d_{V}}]$.
\end{itemize}
Effectively, for the purpose of quantifying the feature diversity of a MSA layer, the feature diversity of each such direct operation on the input is quantified separately and eventually averaged together with all the others.

\section{Threshold-Agnostic Clustering Feature Diversity}
\label{apdx:cluster_diversity}
As explained in section~\ref{sec:cluster_diversity}, the agglomerative clustering stops when the similarity between all pairs of clusters is below the threshold $\tau$ (Equation~\eqref{eqn:cluster}) . 
We define the \emph{cluster ratio} $\mathcal{C}_\tau(W)$ for a given threshold $\tau$ as the ratio between the number of clusters and the number of features. Note, that the more features are clustered together, the lower will be the cluster ratio, indicating low overall feature diversity. 

\begin{figure}[htb]
\begin{subfigure}{0.5\textwidth}
    \centering
    \includegraphics[width=1\textwidth, height=!]{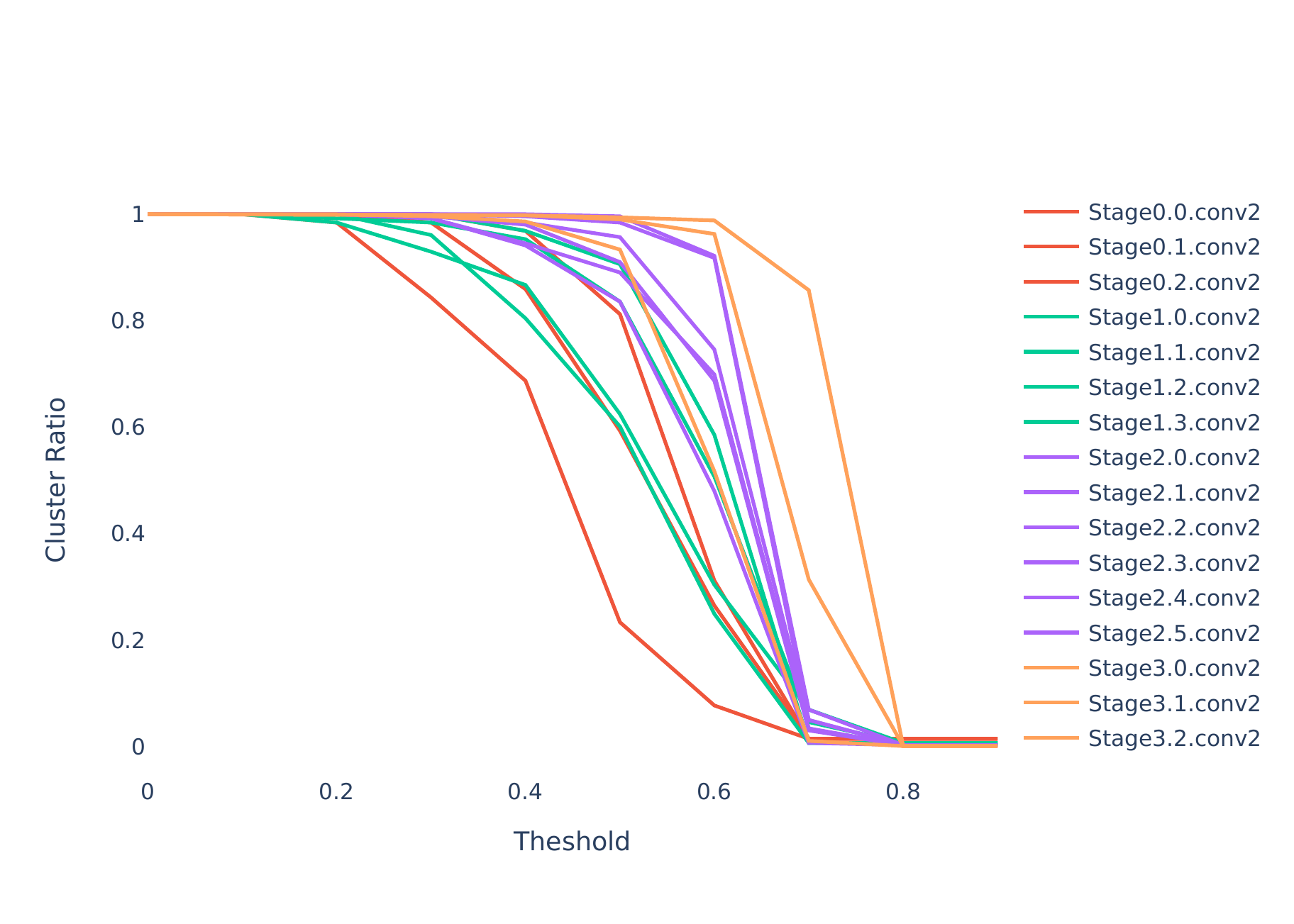}
\end{subfigure}
\begin{subfigure}{0.5\textwidth}
  \centering
    \includegraphics[width=1\textwidth, height=!]{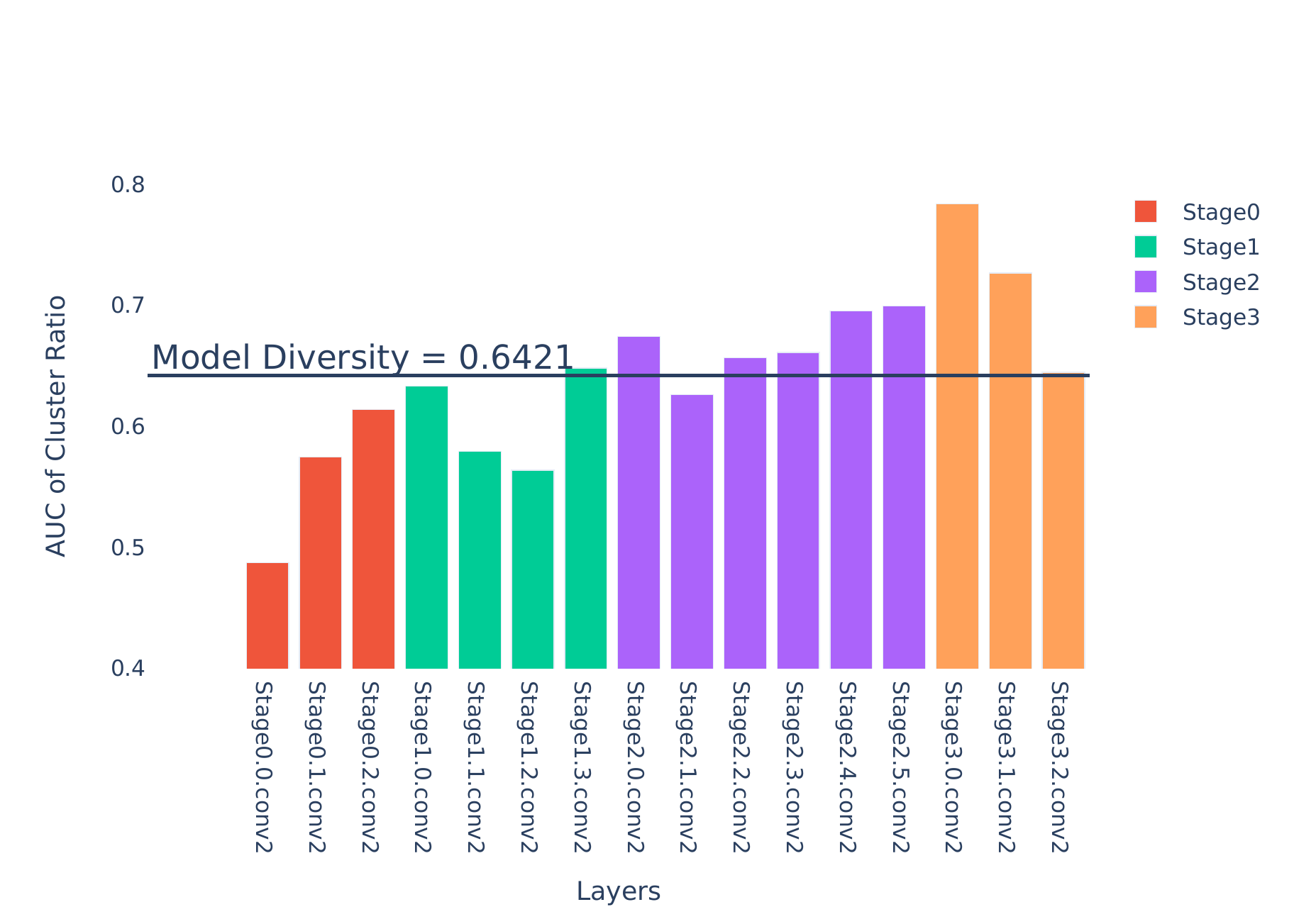}
\end{subfigure}
\begin{subfigure}{0.5\textwidth}
    \centering
    \includegraphics[width=1\textwidth, height=!]{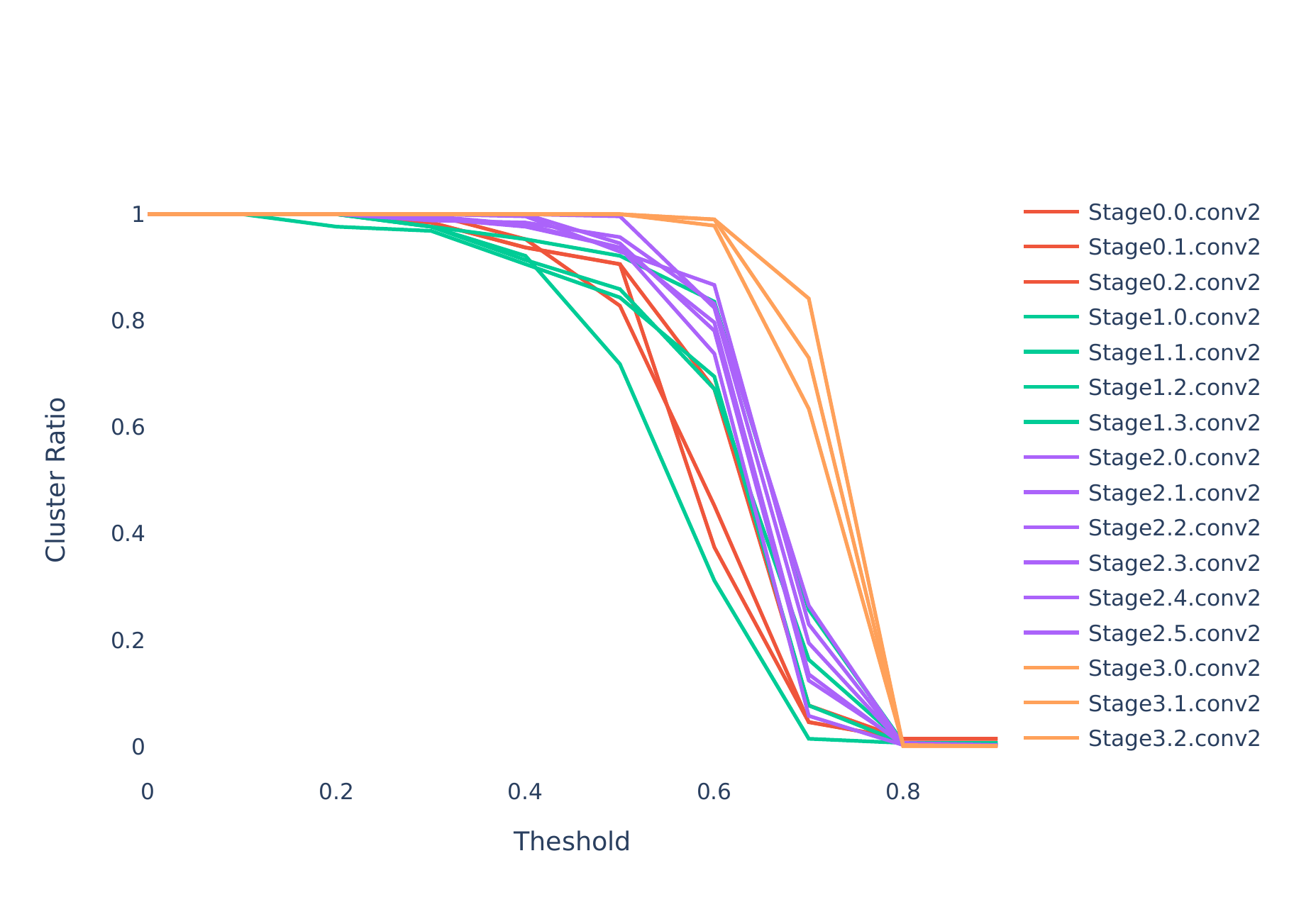}
\end{subfigure}
\begin{subfigure}{0.5\textwidth}
  \centering
    \includegraphics[width=1\textwidth, height=!]{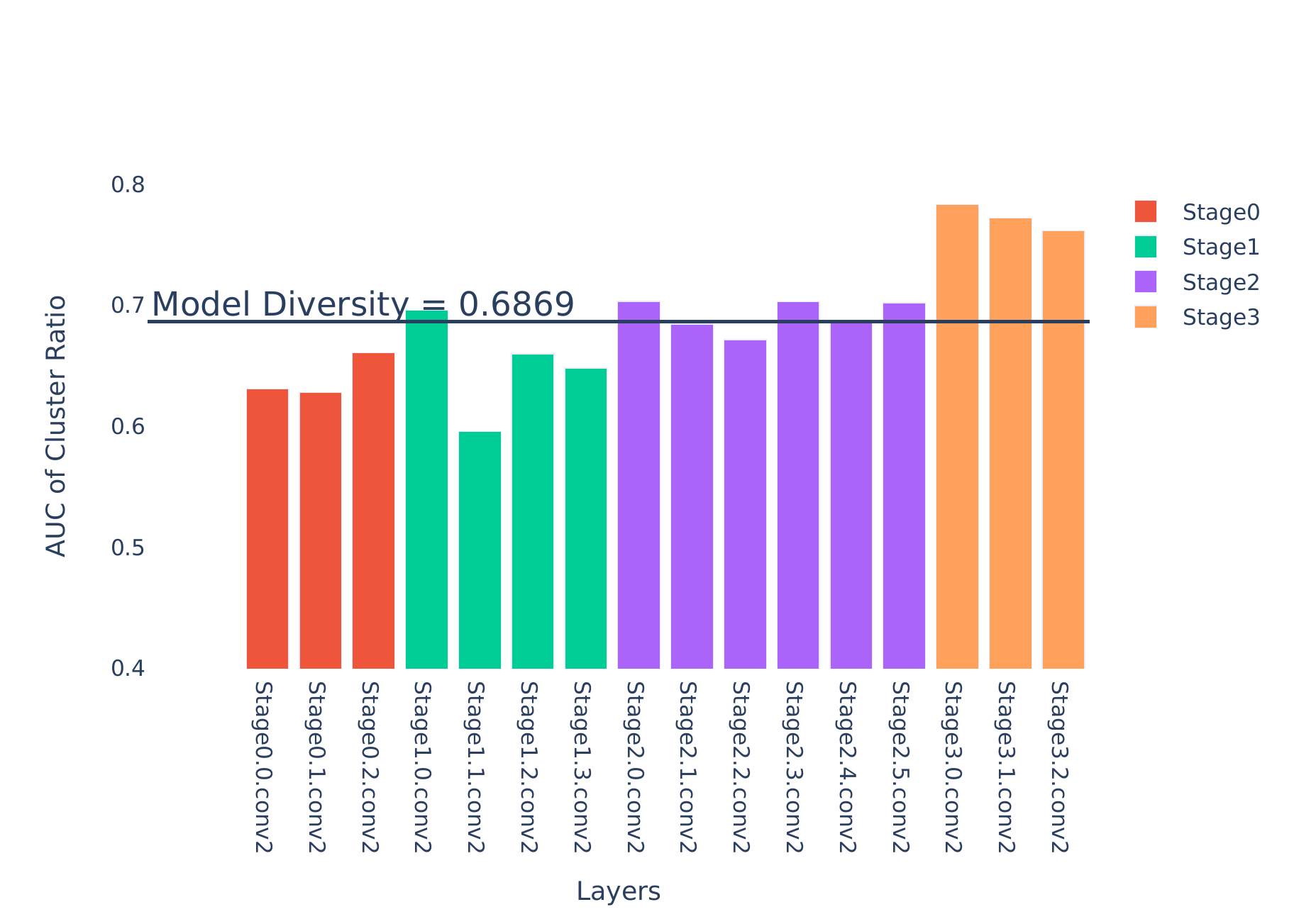}
\end{subfigure}
\begin{subfigure}{0.5\textwidth}
    \centering
    \includegraphics[width=1\textwidth, height=!]{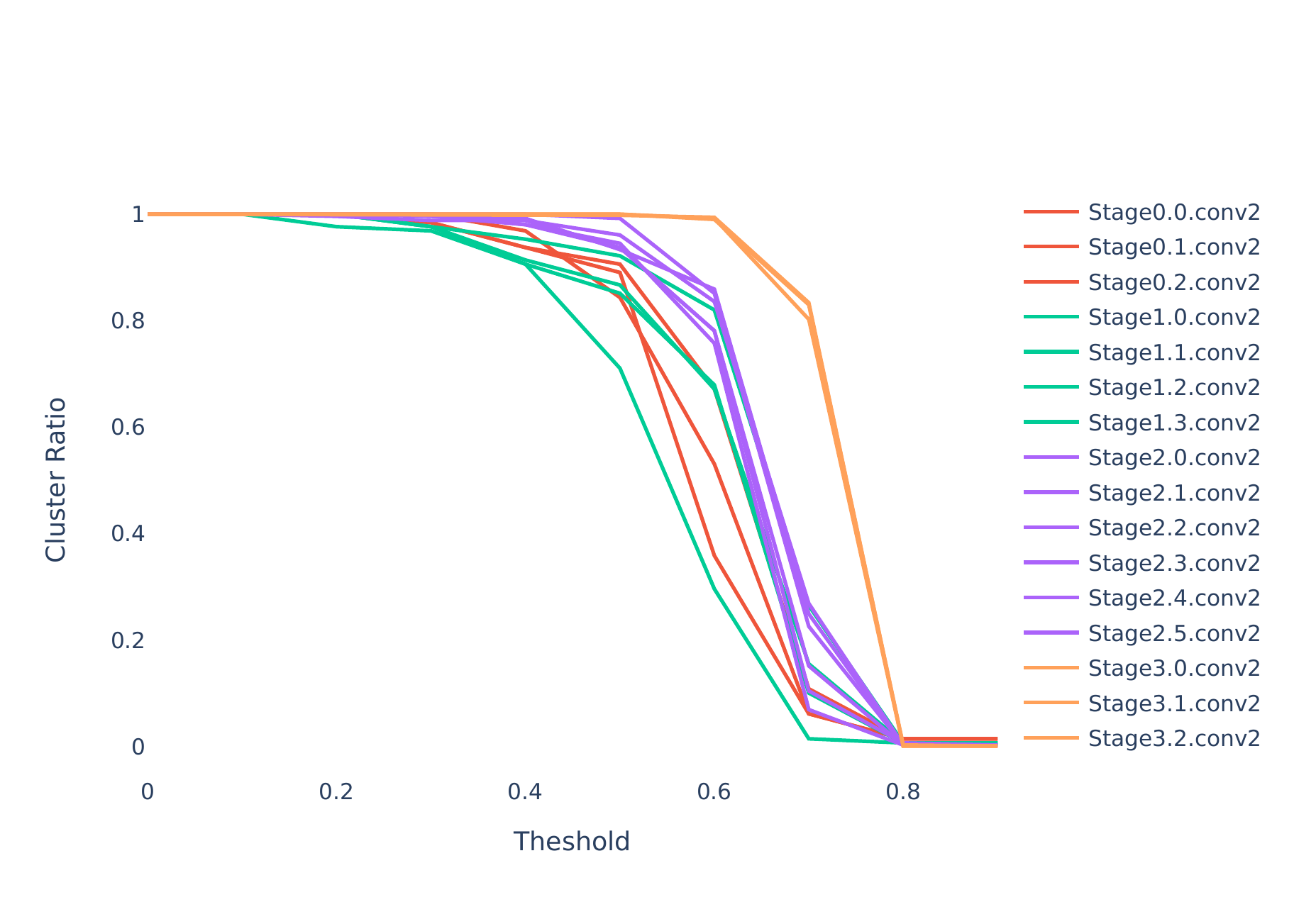}
\end{subfigure}
\begin{subfigure}{0.5\textwidth}
  \centering
    \includegraphics[width=1\textwidth, height=!]{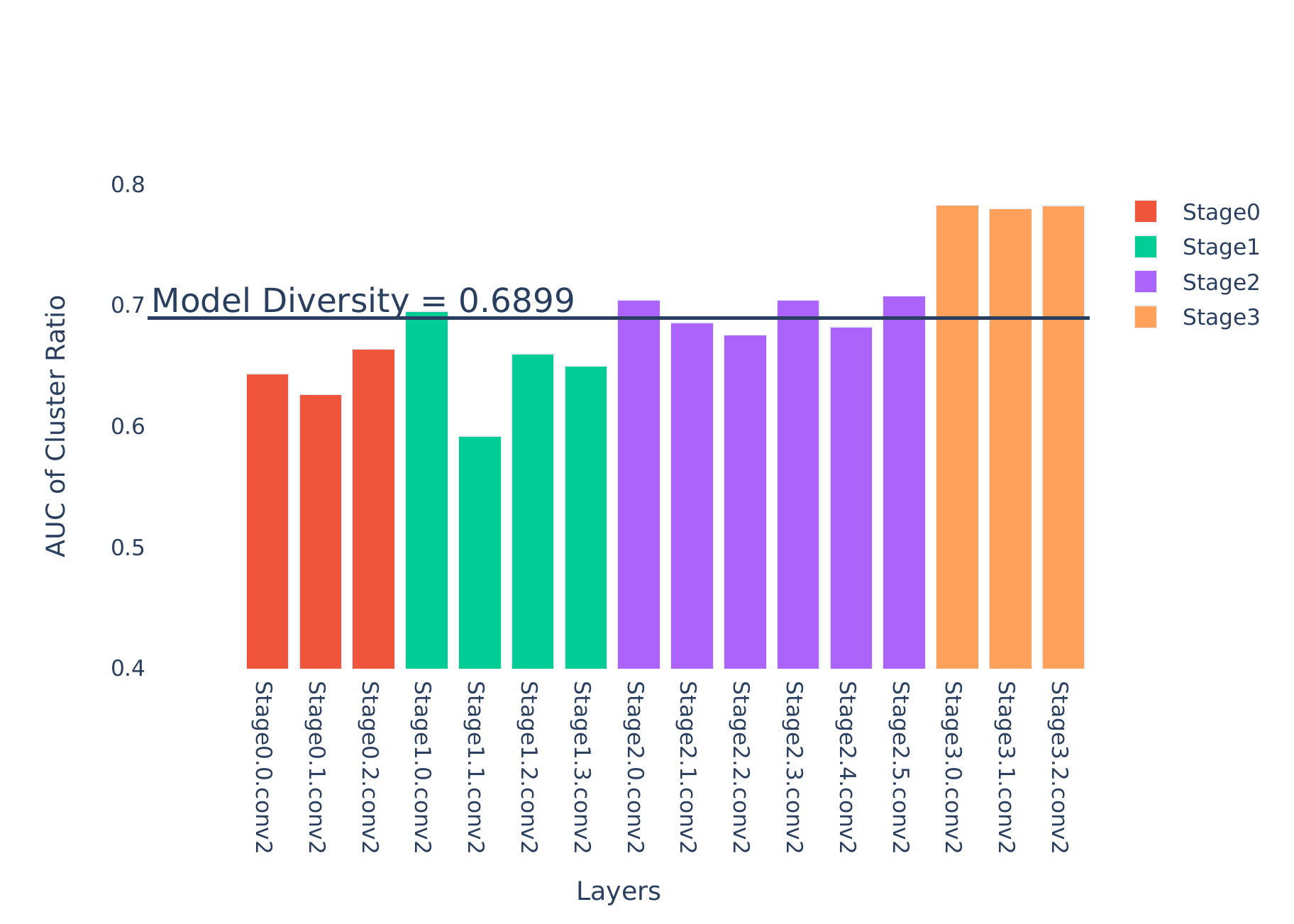}
\end{subfigure}
  \caption{Clustering Diversity of Convolutional Layers of Supervised (Top), SwAV($\mathcal{T}=4$) (Center) and SwAV (Bottom) pretrained models. (Left) The cluster ratio $\mathcal{C}_\tau(W)$ of various convolutional layers of Resnet50 as a function of the clustering threshold $\tau$. (Right) The corresponding area-under-curve (AUC) of the cluster ratio $\mathcal{D}_C(W^{(l)})$. The average AUC over all curves is taken as the overall feature diversity measure.}
  \label{fig:cluster_ratio}
\end{figure}

A well known main caveat of agglomerative clustering is its sensitivity to the threshold $\tau$. Moreover, due to different neural layers of the same model learning different levels of abstractions, a single threshold $\tau$ value does not fit all. 
We observe this sensitivity also in our experiments, illustrated in Figure~\ref{fig:cluster_ratio} (Left), which shows that the cluster ratio $\mathcal{C}_\tau(W)$ is highly sensitive to the choice of thresholds. At the same time, we note that layers that maintain high cluster ratio for high values of $\tau$ are more diverse. We wish to capture this property to obtain a threshold-agnostic measure. 
Hence, differently from \cite{ayinde2019correlation}, we evaluate the clusterability of the entire model by averaging the cluster ratio of layers across a spectrum of threshold values.

This is done by computing the area under the curve (AUC) $\mathcal{D}_C(W^{(l)})=\int_0^1\mathcal{C}_\tau(W^{(l)})d\tau$ for each layer $l=1,\dots,L$, and averaging over all layers, as shown in Figure~\ref{fig:cluster_ratio} (Right).
Thus, the final clustering-based feature diversity measure is computed as:\\ $\bar{\mathcal{D}}_C=\frac{1}{L}\sum_{l=1}^L \mathcal{D}_C(W^{(l)})$.

\vspace{-2mm}
\section{Spectral Feature Diversity}
\label{apdx:spectral_diversity}
\begin{figure}[H]
\vspace{-7mm}
    \centering
    \includegraphics[width=0.98\textwidth, height=!]{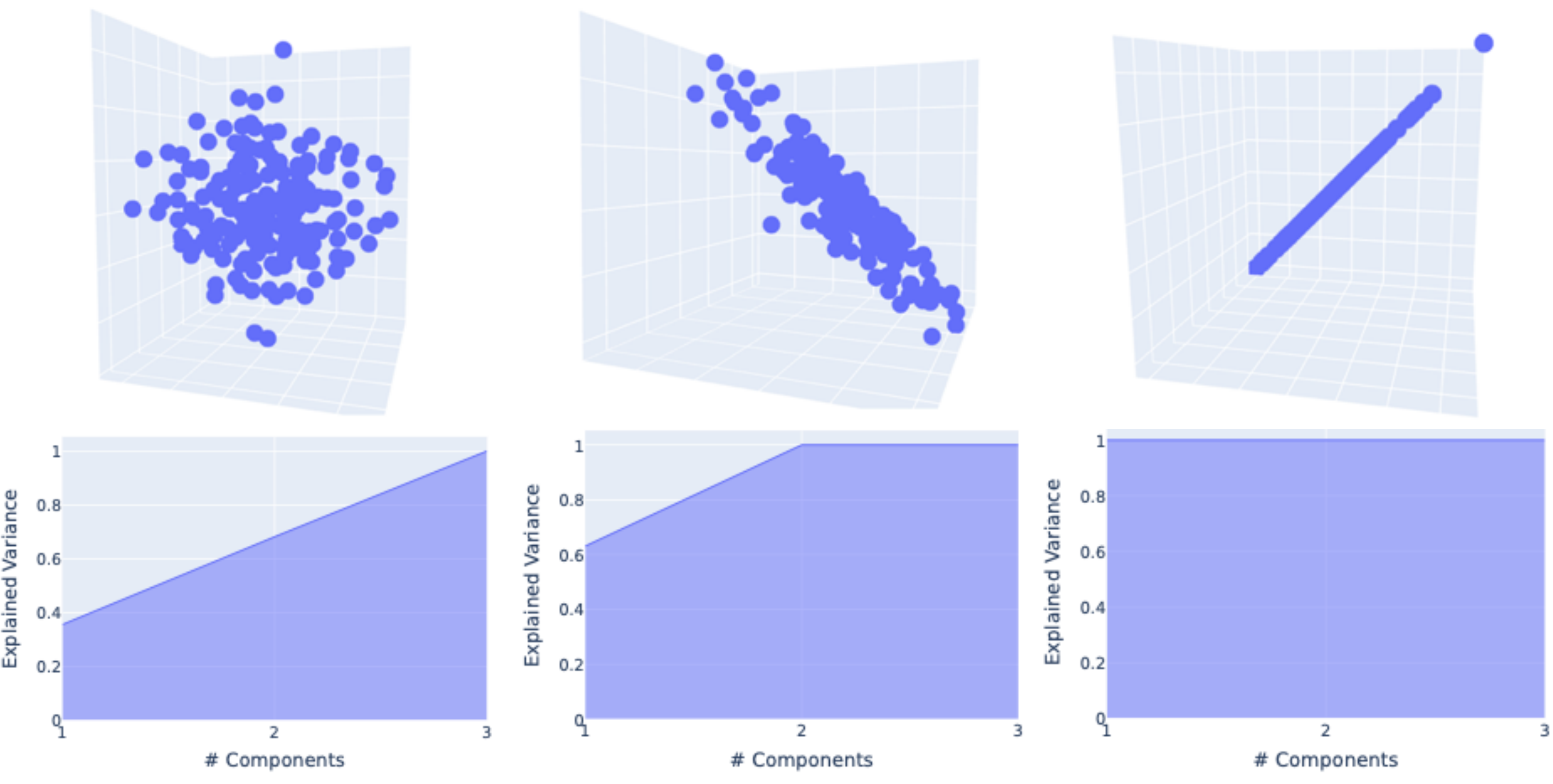}
\vspace{-2mm}
  \caption{(Top) Samples drawn from a normal Gaussian distribution of 3D (left) 2D (middle) and 1D (right) support. (Buttom) The corresponding variance explained by the first principal components and colored AUC. The more diverse the data - the smaller the AUC.}
  \label{fig:spectral_div}
  \vspace{-6mm}
\end{figure}

\vspace{-2mm}
Here we provide the technical details of the computation of the proposed Spectral Feature Diversity in section~\ref{sec:feature_diversity}. 
We compute the singular value decomposition (SVD)\cite{svd} of the feature covaraince matrix $W^T\!\cdot\!W$ to get the orthogonal singular vectors $U_W$ and the rectangular diagonal matrix of positive singular values $\Sigma^2_W$, such that, $\sigma_{W1}^2>\sigma_{W2}^2>\dots>\sigma_{Wd}^2>0$ respectively, and $W^T\!\cdot\!W=U_W\Sigma_W^2 U_W^T$ with $\{\sigma_{Wi}\}_{i=1}^d$ the non-zero eigenvalues of $W$. The variance explained by the $K$ first principal components can be computed as: $\bar{\Sigma}_W^K=\frac{\sum_{k=1}^K\sigma_{Wk}^2}{\sum_{k=1}^d\sigma_{Wk}^2}$. The faster $\bar{\Sigma}_W^K$ increases, the more variance is explained by fewer principal components implying lower diversity. 
This can be quantified by the area under the $\bar{\Sigma}_W^K$ curve, i.e. $\sum_{K=1}^d\bar{\Sigma}_W^K$, as intuitively illustrated in Figure~\ref{fig:spectral_div}. 
Finally, we define the spectral feature diversity of layer $l\!=\!1,\dots,L$ as $\mathcal{D}_\Sigma(W^{(l)})=1-\sum_{K=1}^d\bar{\Sigma}_{W^{(l)}}^K$ and the overall feature diversity of a model is taken as the average over all layers: $\bar{\mathcal{D}}_\Sigma=\frac{1}{L}\sum_{l=1}^L\mathcal{D}_\Sigma(W^{(l)})$.

\section{Measuring the Abstraction of Representations by Centered Kernel Alignment (CKA)} 
Here we explain the way CKA is utilized for measuring the level of abstraction of the representations learnt by the pre-trained model. This measure is considered as a feature implying on the transferablity of the model by \cite{kornblith2021better,islam2021broad} and in Section~\ref{sec:fesature_importance} and Appendix~\ref{apdx:xgboost}. 

We follow the exact calculation of \cite{kornblith2021better} and provide here the details for completeness. Linear CKA provides a way to measure similarity of neural network representations that is invariant to rotation and isotropic scaling in representation space~\cite{kornblith2019similarity,cortes2012algorithms,shawe2002kernel} . Unlike other ways of measuring representational similarity between neural networks, linear CKA can identify architectural correspondences between layers of networks trained from different initializations~\cite{kornblith2019similarity}. Given two matrices $X\in\mathbb{R}^{n\times p_1}$ and $Y\in\mathbb{R}^{n\times p_2}$ containing activations to the same n examples, linear CKA computes the cosine similarity between the reshaped $n\times n$ covariance matrices between examples: 
\begin{align}
    CKA_{linear}(X,Y) = \frac{vec(X^TX)\cdot vec(Y^TY)}{||X^TX||_F||Y^TY||_F}
\end{align}
We measured CKA between all possible pairings of ResNet resolution stages 
of all the models in Table~\ref{tab:feature_importance_wide}. 
To reduce memory requirements, we used minibatch CKA~\cite{nguyen2021do} with minibatches of size $600$ and processed the ImageNet validation set for $3$ epochs.

\begin{figure}[htb]
\begin{subfigure}{1\textwidth}
  \centering
    \includegraphics[width=0.2
    \textwidth, height=!]{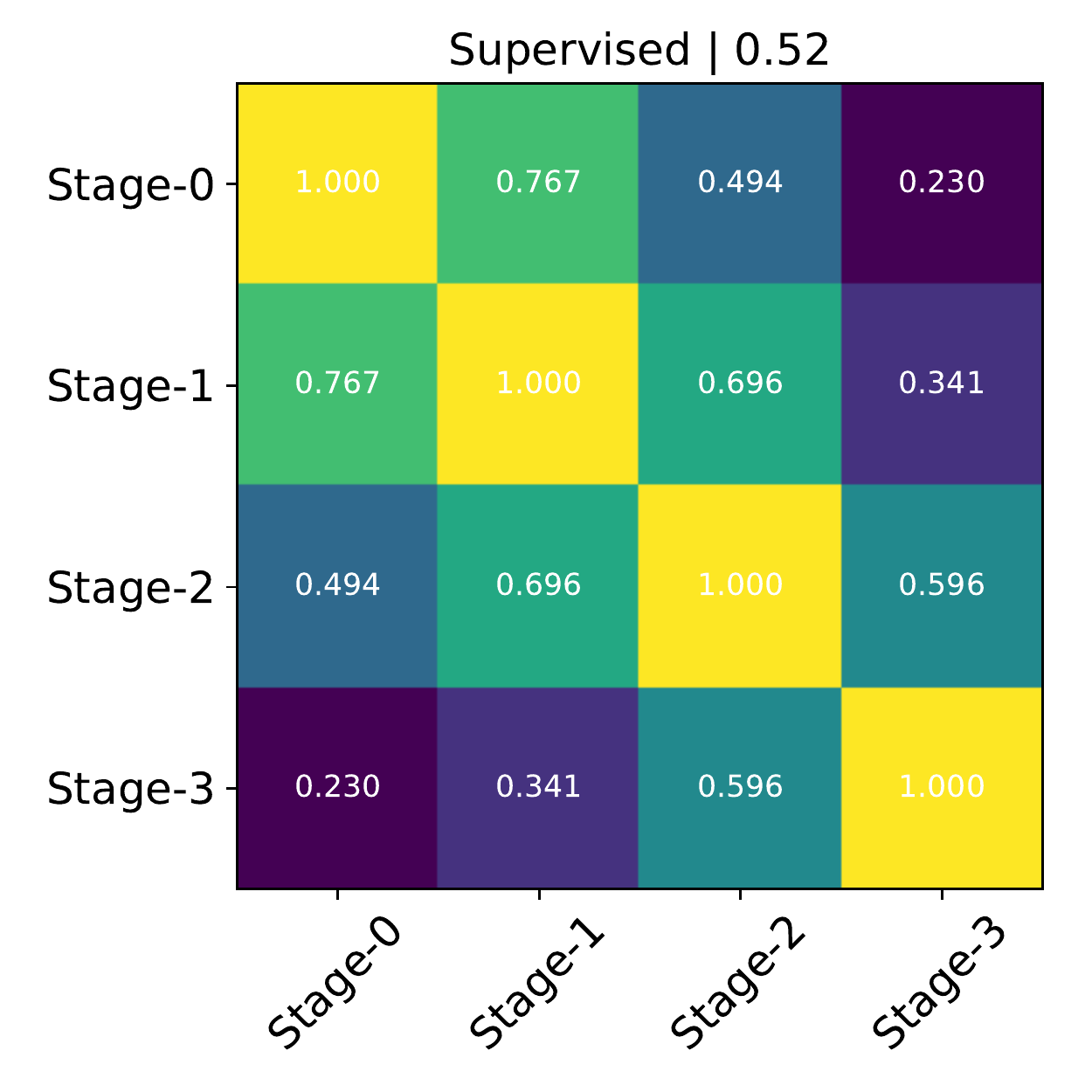}
\end{subfigure}
\begin{subfigure}{1\textwidth}
  \centering
    \includegraphics[width=1\textwidth, height=!]{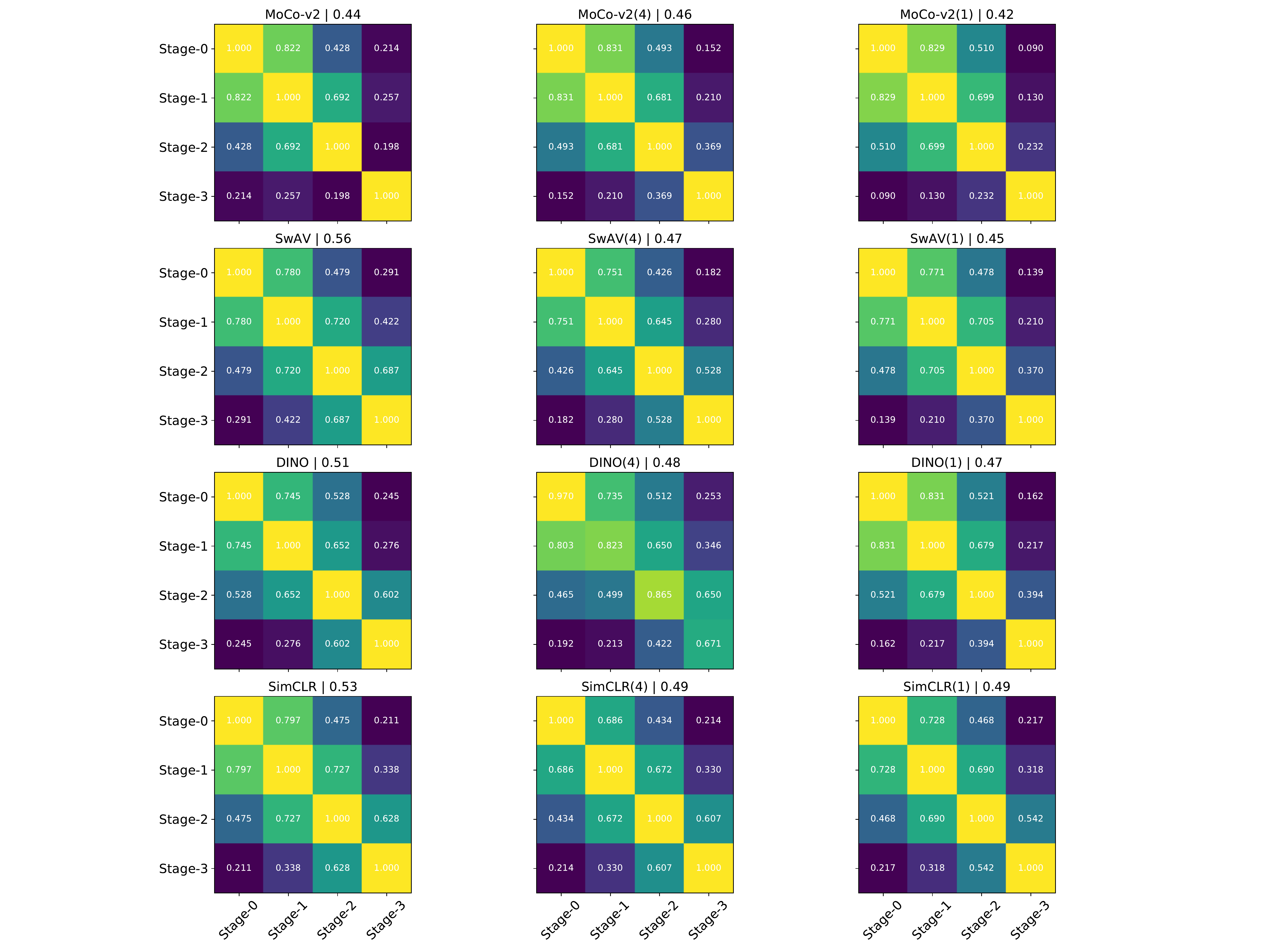}
\end{subfigure}
  \caption{Similarity between different stages of the same model in terms of the CKA for several different pretraining procedures. The label injection control cycle appears in the parentheses. The corresponding CKA score appears in the titles.}
  \label{fig:cka_matrices}
\end{figure}

Figure~\ref{fig:cka_matrices} shows the similarity between different stages of the same model in terms of the CKA for several different pretraining procedures. The numbers in the titles are the overall CKA score reported in Table~\ref{tab:feature_importance_wide}, calculated as the average of the off-diagonal values. 

\section{Intra-class Variance and Class Separation}
Supervised learning models learn feature representations by objectives that also increase the inter-class separation. Related
However, although might be harmful for in-domain performance, \cite{islam2021broad} argued that increasing the intra-class variation was beneficial for learning rich feature representations in transfer learning. 
Thos measure is considered as a feature implying on the transferablity of the model by \cite{kornblith2021better,islam2021broad} and in Section~\ref{sec:fesature_importance} and Appendix~\ref{apdx:xgboost}.

The intra-class variation and inter-class separation are computed as follows \cite{kornblith2021better}:
\begin{align}
    V_{intra} 
    &
    = \sum_{k=1}^K\sum_{m=1}^{N_k}\sum_{n=1}^{N_k}\frac{1-cosine(x_{k,m}, x_{k,n})}{KN_k^2}
    \\
    S_{inter} 
    &
    = \sum_{j=1}^K\sum_{k=1}^K\sum_{m=1}^{N_j}\sum_{n=1}^{N_k}\frac{1-cosine(x_{k,m}, x_{k,n})}{K^2 N_k N_j}
\end{align}
where $x_{k,m}$ is the embedding of example $m$ in class $k\in\{1, \dots , K\}$ and $N_k$ is the number of examples in class $k$. 
Those metrics for all of the pretrained CNN models are listed in Table~\ref{tab:feature_importance_wide}.

\subsection{The Mean Silhouette Coefficient} (MSC) \cite{rousseeuw1987silhouettes} is an appropriate metric for quantifying class separation in the embedding space.  
This measure is considered as a feature implying on the transferablity of the model in Section~\ref{sec:fesature_importance} and Appendix~\ref{apdx:xgboost}.

For each embedding vector $x_{k,m}$ of example $m$ in class $k\in\{1, \dots , K\}$ and $N_k$, the number of examples in class $k$, the Silhouette coefficient $SC_m$ is the relationship between the intra-class distances $v_m$ and the nearest class distances $s_m$ as follows:
\begin{align}
    SC_m &=\frac{s_m-v_m}{\max(s_m-v_m)}
    \\
    v_m &= \sum_{n=1}^{N_k}\frac{1-cosine(x_{k,m}, x_{k,n})}{N_k-1}
    \\
    s_m &= \min_{j\in\{1,\dots,K\}\setminus \{k\}}\sum_{n=1}^{N_j}\frac{1-cosine(x_{k,m}, x_{k,n})}{N_j}
\end{align}
Since $|s-v|\leq \max(s,v)$, the mean Silhouette coefficient $MSC = \frac{\sum_{k=1}^K\sum_{m=1}^{N_k}SC_m}{\sum_{k=1}^K N_k}$ is bounded between $-1$ to $1$. 
Those values for all of the pretrained CNN models are listed in Table~\ref{tab:feature_importance_wide}. 
There is a positive correlation between MSC and Imagenet accuracy, indeed validating that representations with higher class separation obtain higher accuracy on the upstream task, and thus class seperation by its own does not add much information over Imagenet accuracy. 

\section{Quantifying Feature Importance by Gradient Boosting Decision Trees (XGBoost)}  
\label{apdx:xgboost}
In section~\ref{sec:fesature_importance} we use gradient boosting decision trees for quantifying the importance of different factors on the transferability. A benefit of using gradient boosting \cite{chen2016xgboost} for solving a regression problem is that after the boosted trees are constructed, it is relatively straightforward to retrieve importance scores for each input feature. 
Generally, importance provides a score that indicates how useful or valuable each feature was in the construction of the boosted decision trees within the model. The more a feature is used to make key decisions with decision trees, the higher its relative importance. 
This importance is calculated explicitly for each feature, allowing features to be ranked and compared to each other. 
Importance is calculated for a single decision tree by the amount that each feature split point improves the performance measure, weighted by the number of observations the node is responsible for, see \cite{elith2008working} for more details. 
The feature importance scores are then averaged across all of the the decision trees within the model.
To this end, XGBoost~\cite{chen2016xgboost}, with the default \textit{Scikit-Learn} parameters of 100 trees of maximal depth of 6, is fed with the set of inspected features \{Imagenet accuracy, feature diversity, CKA, MSC, intra-class variation, inter-class separation\} together with the corresponding transfer learning scores for each of the 40 pretrained models forming our dataset in Table~\ref{tab:feature_importance_wide}. 
The derived feature importance is presented in Figure~\ref{fig:xgboost}. It again clearly shows that most of the importance is attributed to Imagenet accuracy together with feature diversity, as the rest of the features, previously considered important \cite{kornblith2021better,islam2021broad}, are overshadowed by the former.

\input{experiments/full_feature_importance}

\section{Fully Connected Initialization}
\label{apdx:fc_init}
A standard practice for fully-connected initializion when fine tuning a pretrained backbone on a new task is either linear probe or random initializaiton. Linear probe uses a fixed backbone as feature extractor and the classification head is trained with gradient descent. We find that it is undesirable to start training with a randomly initialized fully connected layer on top of a pretrained backbone when fine-tuning on the downstream tasks, as it requires the linear probing to have multiple epochs and many hyper parameters. Instead we look at the fully connected as class centroids that approximate to nearest neighbor classification. Hence, we initialize the $i^{th}$ column of the fully connected to be the mean class representation for the $i^{th}$ class and such that $W^i=\frac{1}{|C_i|}\sum_{x_i\in C_i}f(x_i)$.

\section{Linear Probing}
\label{apdx:linear_prob}
Despite the focus of this work is on the fine-tuning of the pretrained models, in this section we show that the conclusions presented throughout the paper are also valid for linear probing over the CNN models, when the pretrained backbones are fixed and a logistic regression classifier is trained on the downstream datasets, using features extracted from the penultimate layer their as inputs.

\subsection{Linear Probing by Logistic Regression}
\label{apdx:lin_prob_logistic}
In this section we present the technical details involved in performing the linear probing by solving a fitting a logistic regression. Although we follow the protocol proposed by \cite{kornblith2019better}, we give here the details for completeness. For each dataset, we extracted features from the penultimate layer of the network. We trained a multinomial logistic
regression classifier implemented by \textit{Scikit-Learn} with the default settings, using L-BFGS, with an L2 regularization parameter applied to the sum of the per-example losses, selected from a range of 45 logarithmically spaced values from $10^{-6}$ to $10^5$ on the validation set. Since the optimization problem is convex, we used the solution at the previous point along the regularization path as a warm start for the next point, which greatly accelerated the search. For these experiments, we did not perform data augmentation or scale aggregation, and we used
the entire image, rather than cropping the central 87.5\% as is common for testing on ImageNet.

\subsection{The Same Conclusions Hold For Linear Probing}\label{apdx:linprob_conc}
\begin{figure}[H]
\vspace{-5mm}
\centering
\begin{subfigure}{.32\textwidth}
  \centering
    \includegraphics[width=0.98\textwidth, height=!]{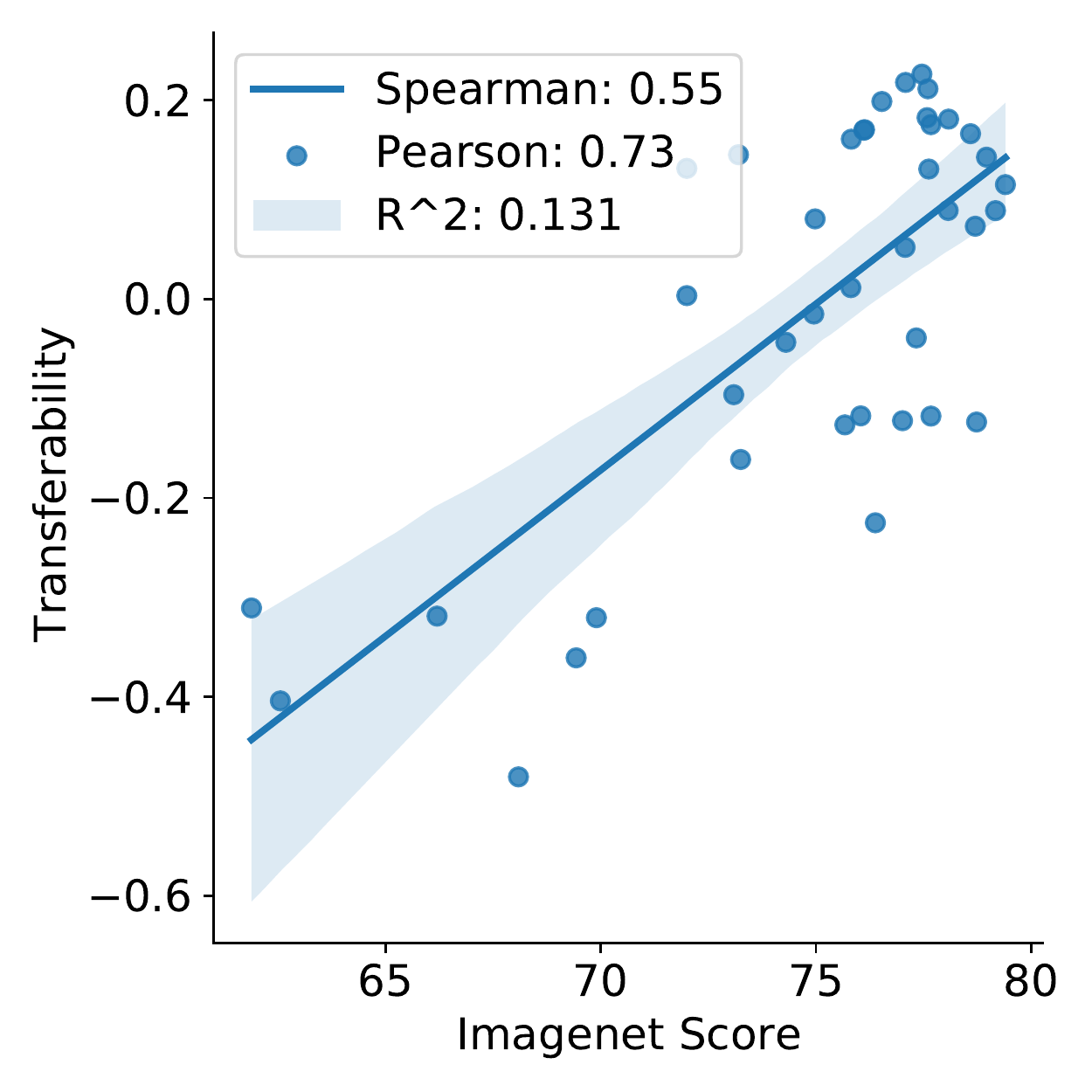}
  \label{fig:r2_imagenet_linprob}
\end{subfigure}
\begin{subfigure}{.32\textwidth}
  \centering
    \includegraphics[width=0.98\textwidth, height=!]{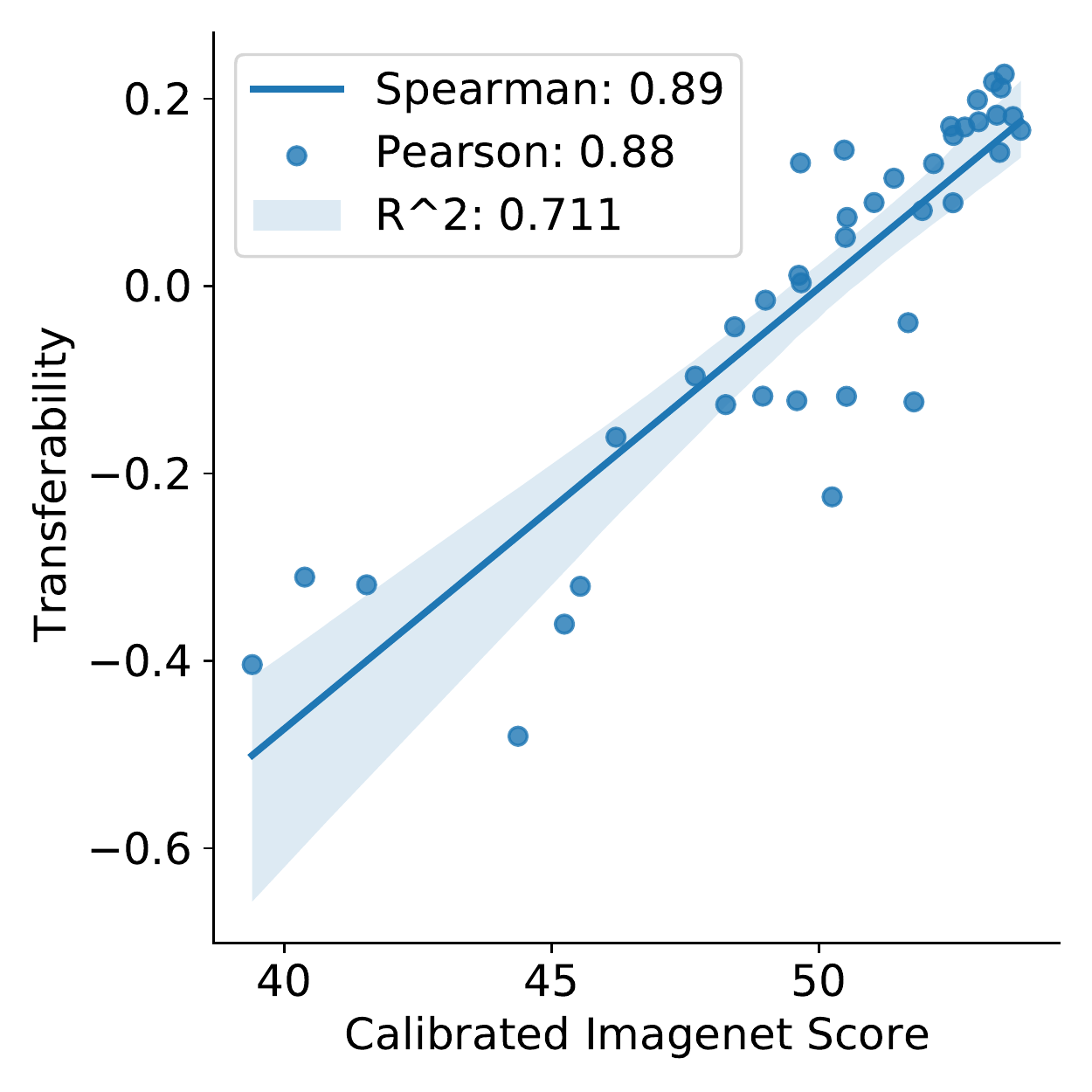}
  \label{fig:r2_calibrated_imagenet_linprob}
\end{subfigure}
\begin{subfigure}{.32\textwidth}
  \centering
    \includegraphics[width=0.98\textwidth, height=!]{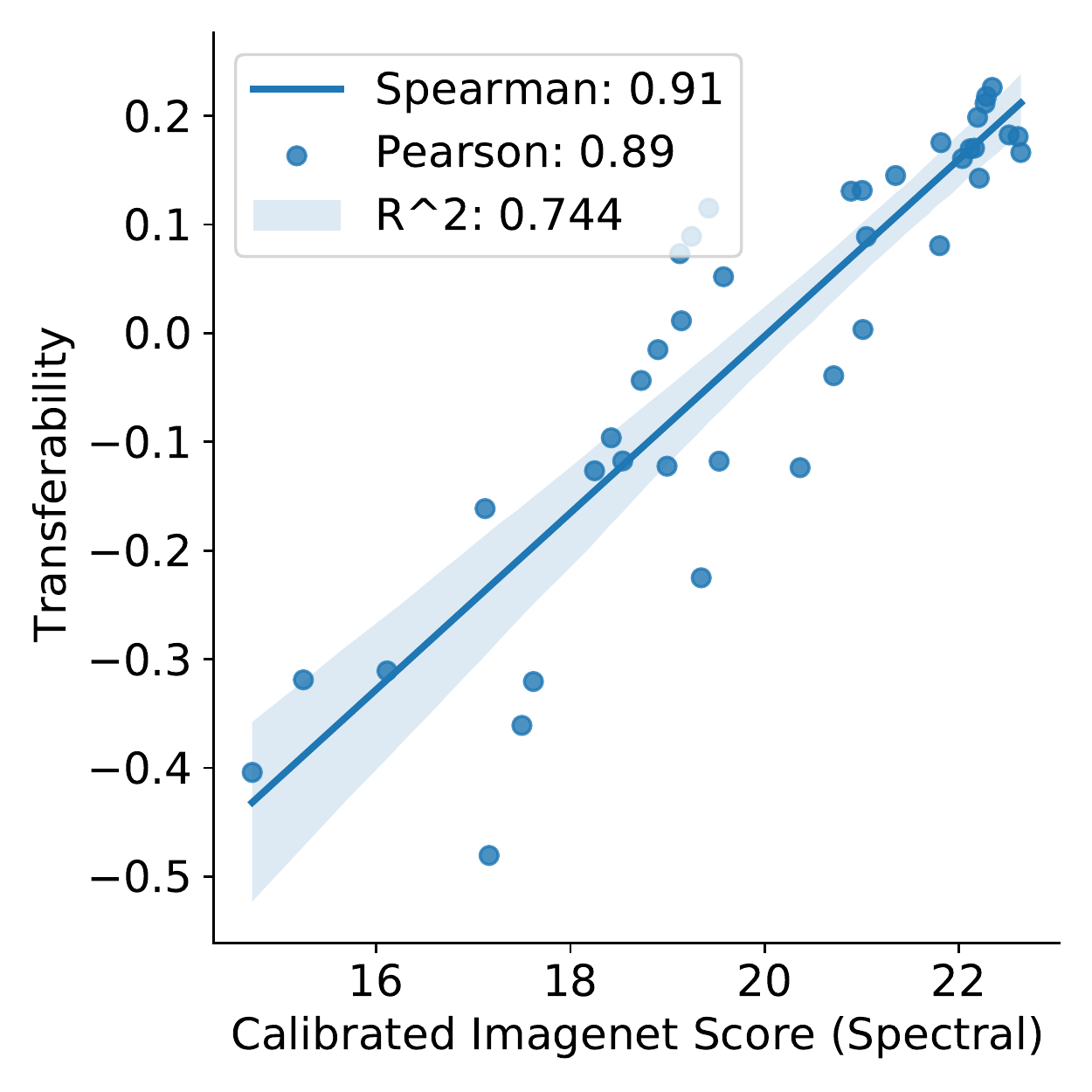}
  \label{fig:r2_spectral_calibrated_imagenet_linprob}
\end{subfigure}

\caption{Linear probing transferability vs Imagent Score (Left), the Clustering  Feature Diversity based CIS (Middle) and the Spectral Feature Diversity based CIS (Right) for 40 models that were pre-trained with supervised learning, self-supervised learning or their combination. The Calibrated Imagenet Score correlates with transferability significantly better. CIS correlates with transferability significantly better than imagenet score in both cases.}
\label{fig:tl_correlation_lin_prob}
\vspace{-5mm}
\end{figure}

\begin{figure}[htb]
\vspace{-5mm}
    \centering
    \includegraphics[width=0.8\textwidth, height=!]{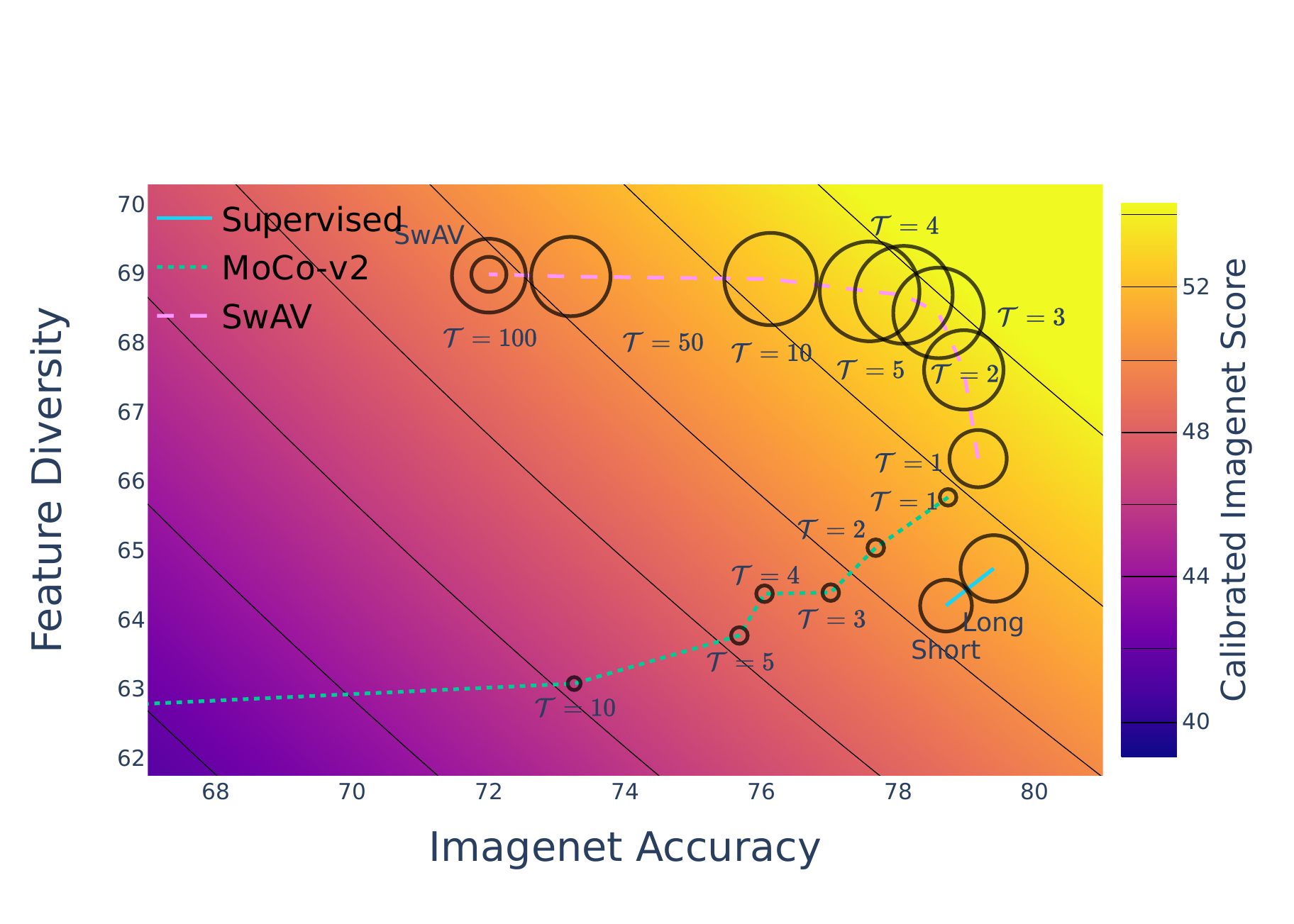}
    \vspace{-3mm}
  \caption{Circle size corresponds to the linear probing transferability averaged over 14 downstream tasks, as a function of Imagenet top-1 accuracy (x axis) and feature diversity (y axis). Results shown for 3 different training methods (see legend). The background colors and curves show the Calibrated Imagenet Score. Evidently, models with both high Imagenet accuracy and high Feature Diversity, that together result in high Calibrated Imagenet Score (in yellow), transfer better (larger circles).}
  \label{fig:imgnt_div_cor_cnn_lin_prob}
  \vspace{-15mm}
  
\end{figure}

\begin{figure}[htb]
    \centering
    \includegraphics[width=0.9\textwidth, height=!]{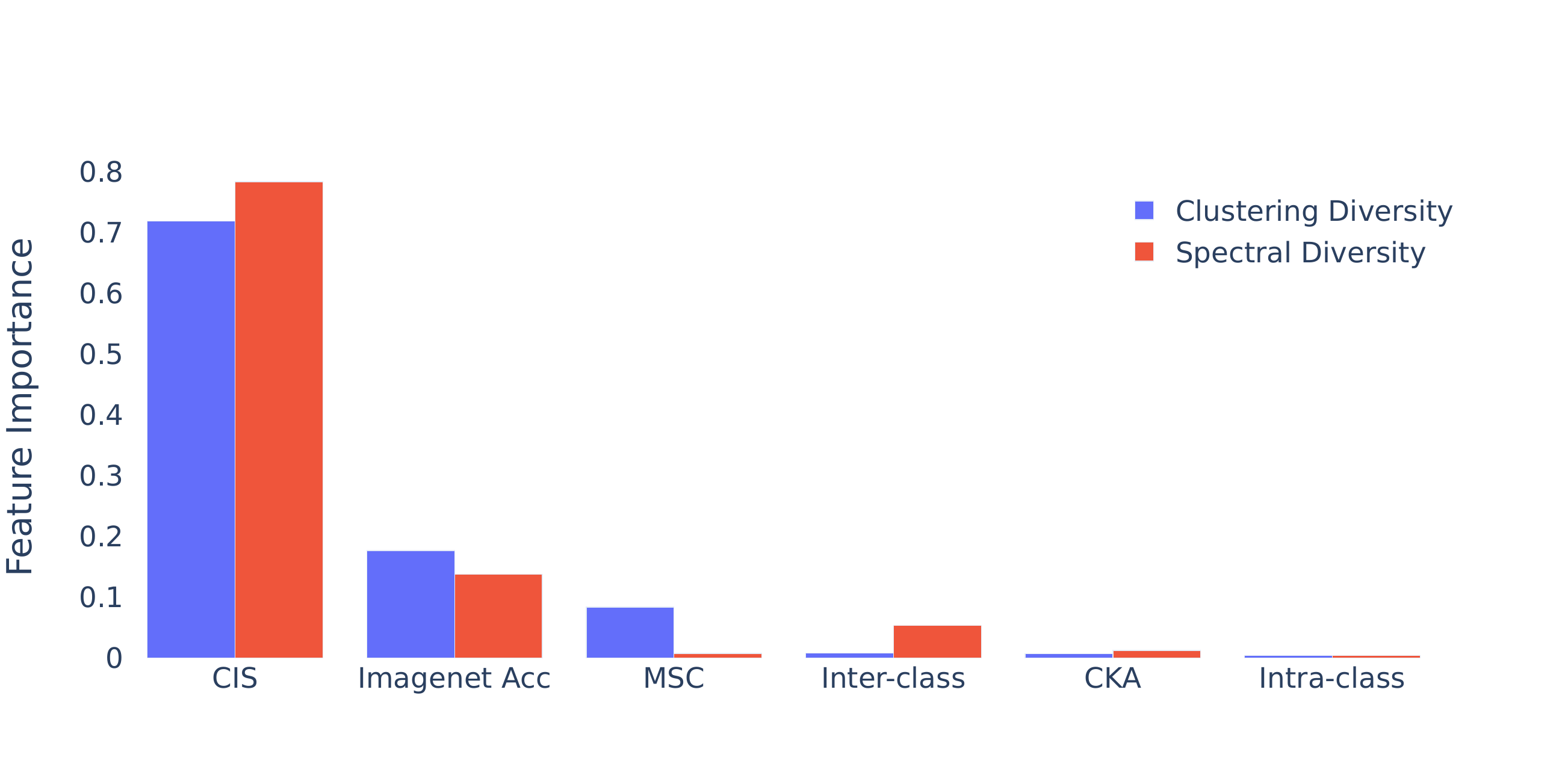}
  \caption{The relative importance for linear probing transferability of different factors are quantified by the popular feature importance derived from XGBoost. Most of the importance is attributed to the Calibrated Imagenet Score (CIS).}
\label{fig:xgboost_lin_prob}
\end{figure}

Here we show the corresponding results of Figure~\ref{fig:tl_correlation},  Figure~\ref{fig:imgnt_div_cor_cnn} and Figure~\ref{fig:xgboost} in Figure~\ref{fig:tl_correlation_lin_prob},  Figure~\ref{fig:imgnt_div_cor_cnn_lin_prob} and Figure~\ref{fig:xgboost_lin_prob} respectively.


\newpage
\section{Transfer Learning Results - Full Tables}
\subsection{Finetune Tables}
In this section we present the full transfer learning results in Tables~\ref{tab:cnn_wide_full} and \ref{tab:vit_wide_full} for CNN and ViT models respectively.

\input{experiments/single_label_res_full}
\input{experiments/single_label_vit_full}

\subsection{Linear Probing CNN Tables}
In this section we present the full transfer learning results for linear probing in Tables~\ref{tab:full_table_linprob} for CNN.
\input{experiments/single_label_res_full_linprob_search}


\section{Feature Diversity for Vision Transformers}\label{apdx:vit_2d}
\begin{figure}[H]
\vspace{-5mm}
    \centering
    \includegraphics[width=0.8\textwidth, height=!]{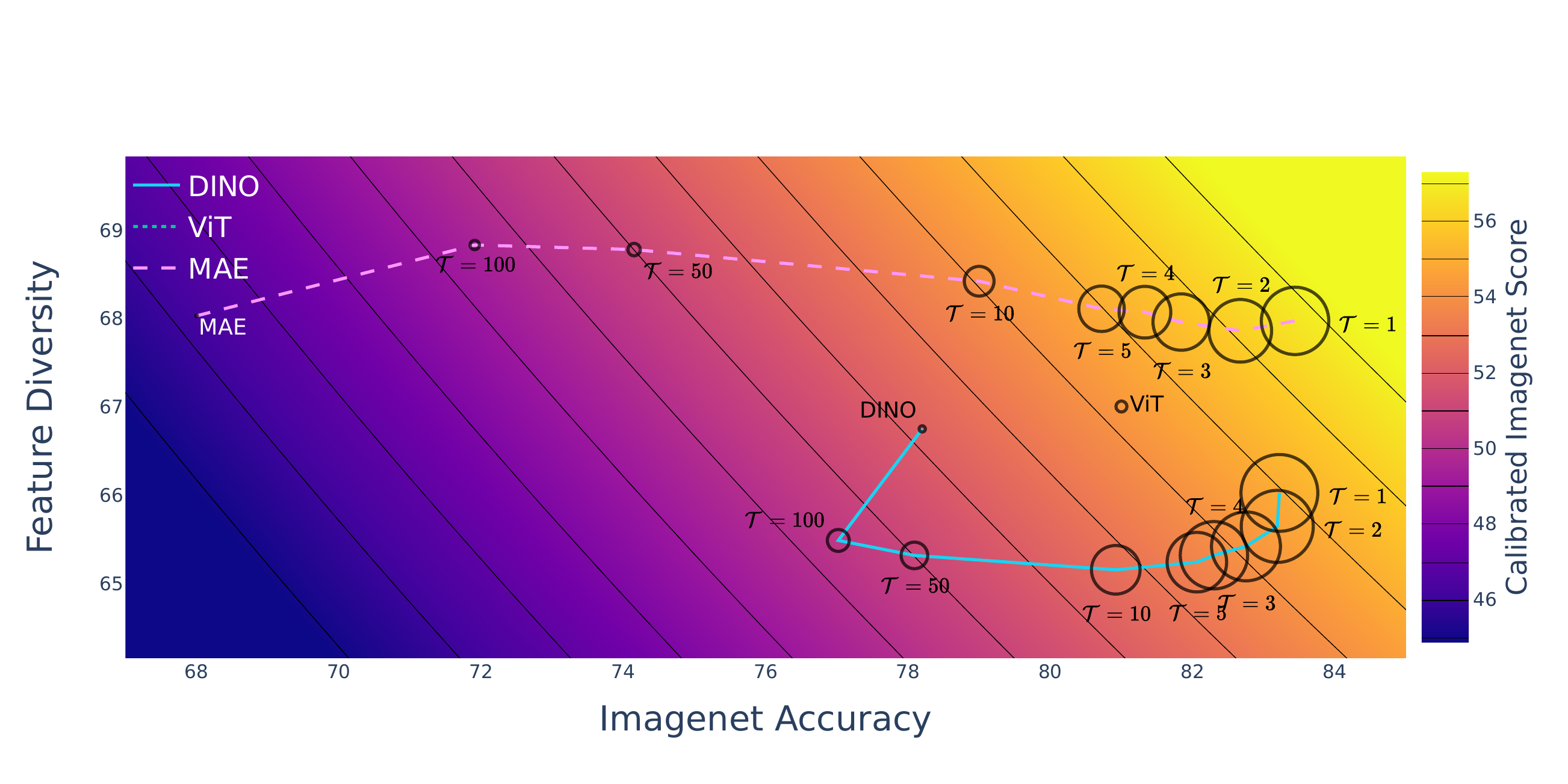}
    \vspace{-3mm}
  \caption{Circle size corresponds to the transferability of \textbf{finetuned ViT} models averaged over 14 downstream tasks, as a function of Imagenet top-1 accuracy (x axis) and feature diversity (y axis). Results shown for 3 different training methods (see legend). The background colors and curves show the Calibrated Imagenet Score. Evidently, models with both high Imagenet accuracy and high Feature Diversity, that together result in high Calibrated Imagenet Score (in yellow), transfer better (larger circles).}
  \label{fig:imgnt_div_cor_vit}
  \vspace{-15mm}
  
\end{figure}

Similarly to Figure~\ref{fig:imgnt_div_cor_cnn} and Figure~\ref{fig:imgnt_div_cor_cnn_lin_prob},  Figure~\ref{fig:imgnt_div_cor_vit} (presenting Table~\ref{tab:vit_wide_full}) shows how the control label injection (CLI) can start off from different SSL pre-trained ViT models of both higher (MAE) and lower (DINO) feature diversity and generate ViT models of different levels of Imagenet accuracy and feature diversity for different control cycle values. Those generated ViT models allow us to make observations about the connection between Imagenet Score and Feature Diversity, associated with Multi-head Self-attention (MSA) layers, to the transferability through the Calibrated Imagenet Score. Similarly to the behaviour of CNNs, the trajectory for every origin SSL model traverses the Calibrated Imagenet Score contour lines towards more transferable regions, as expressed by the size of the circles and the background color. Specifically, the upper trajectory, originated in MAE, represents a case where CLI maintains the feature diversity while improving Imagenet score as the main mean for improving the diversity. Another case is represented by the lower DINO originated trajectory, where at some point both Imagenet score and feature diversity improve together towards better transferability. Especially interesting is the two right most points of this DINO trajectory ($\mathcal{T}=2$ and $\mathcal{T}=1$), that share the very same Imagenet score, yet differ in the feature diversity resulting in the difference in transferability.

%% file: experiments/full_feature_importance.tex
\begin{table}[H]
\vspace{-5mm}
    \centering
    \tiny

\begin{tabular}{l|ccccccc}
\toprule
{}      Pretrain                            &  
\rotatebox[origin=c]{-70}{CIS (Clustering)} &  
\rotatebox[origin=c]{-70}{CIS (Spectral)} &
\rotatebox[origin=c]{-70}{ImNet} &  
\rotatebox[origin=c]{-70}{Inter}&    
\rotatebox[origin=c]{-70}{CKA} &  
\rotatebox[origin=c]{-70}{Intra} &    
\rotatebox[origin=c]{-70}{MSC} \\
\midrule                                                                         
Supervised                          & 51.40 & 19.12 & 79.40 &  0.422 & 63.81 &  0.510 &  0.09 \\
SupCon~\cite{khosla2020supervised}  & 51.67 & 20.71 & 77.33 &  0.276 & 57.84 &  0.275 &  0.14 \\
\midrule                                                                                          
CE+SelfSupCon~\cite{islam2021broad} & 50.25 & 19.34 & 76.380 &  0.296 & 59.625 &  0.322 &  0.137 \\
Moco-V2 ($\mathcal{T}= 1$)          & 51.78 & 20.36 & 78.73 &  0.276 & 56.13 &  0.278 &  0.18 \\
Moco-V2 ($\mathcal{T}= 2$)          & 50.52 & 19.53 & 77.67 &  0.273 & 57.27 &  0.270 &  0.16 \\
Moco-V2 ($\mathcal{T}= 3$)          & 49.59 & 18.99 & 77.01 &  0.269 & 58.25 &  0.266 &  0.14 \\
Moco-V2 ($\mathcal{T}= 4$)          & 48.95 & 18.54 & 76.04 &  0.266 & 59.19 &  0.264 &  0.13 \\
Moco-V2 ($\mathcal{T}= 5$)          & 48.26 & 18.24 & 75.67 &  0.263 & 59.96 &  0.262 &  0.12 \\
Moco-V2 ($\mathcal{T}= 10$)         & 46.49 & 17.22 & 73.70 &  0.257 & 61.85 &  0.258 &  0.08 \\
Moco-V2 ($\mathcal{T}= 50$)         & 43.62 & 16.01 & 69.52 &  0.271 & 63.40 &  0.268 &  0.03 \\
Moco-V2 ($\mathcal{T}= 100$)        & 42.72 & 15.96 & 67.83 &  0.290 & 63.91 &  0.279 &  0.00 \\
Moco-V2                             & 44.17 & 17.62 & 67.70 &  0.377 & 57.63 &  0.419 &  0.04 \\
\midrule                                                                            
SwAV ($\mathcal{T}= 1$)             & 52.51 & 21.05 & 79.17 &  0.321 & 58.42 &  0.356 &  0.12 \\
SwAV ($\mathcal{T}= 2$)             & 53.38 & 22.21 & 78.96 &  0.294 & 59.20 &  0.335 &  0.10 \\
SwAV ($\mathcal{T}= 3$)             & 53.78 & 22.64 & 78.59 &  0.277 & 59.80 &  0.317 &  0.08 \\
SwAV ($\mathcal{T}= 4$)             & 53.63 & 22.61 & 78.08 &  0.269 & 60.15 &  0.307 &  0.06 \\
SwAV ($\mathcal{T}= 5$)             & 53.33 & 22.52 & 77.58 &  0.265 & 60.34 &  0.302 &  0.06 \\
SwAV ($\mathcal{T}= 10$)            & 52.46 & 22.16 & 76.12 &  0.260 & 60.71 &  0.296 &  0.04 \\
SwAV ($\mathcal{T}= 50$)            & 50.47 & 21.35 & 73.20 &  0.265 & 61.45 &  0.308 &  0.01 \\
SwAV ($\mathcal{T}= 100$)           & 49.65 & 21.00 & 72.00 &  0.267 & 61.52 &  0.629 &  0.00 \\
SwAV                                & 49.67 & 21.01 & 72.00 &  0.259 & 67.23 &  0.367 & -0.00 \\
\midrule                                                                            
DINO ($\mathcal{T}= 1$)             & 52.14 & 20.89 & 77.62 &  0.301 & 60.05 &  0.692 &  0.11 \\
DINO ($\mathcal{T}= 2$)             & 52.99 & 21.81 & 77.67 &  0.280 & 60.92 &  0.656 &  0.08 \\
DINO ($\mathcal{T}= 3$)             & 53.40 & 22.27 & 77.60 &  0.266 & 61.14 &  0.625 &  0.07 \\
DINO ($\mathcal{T}= 4$)             & 53.47 & 22.34 & 77.46 &  0.260 & 61.34 &  0.608 &  0.06 \\
DINO ($\mathcal{T}= 5$)             & 53.27 & 22.28 & 77.08 &  0.257 & 61.32 &  0.600 &  0.05 \\
DINO ($\mathcal{T}= 10$)            & 52.96 & 22.19 & 76.53 &  0.254 & 61.66 &  0.594 &  0.03 \\
DINO ($\mathcal{T}= 50$)            & 52.73 & 22.11 & 76.11 &  0.260 & 62.69 &  0.625 &  0.01 \\
DINO ($\mathcal{T}= 100$)           & 52.52 & 22.03 & 75.82 &  0.263 & 63.22 &  0.643 &  0.00 \\
DINO                                & 51.93 & 21.80 & 74.98 &  0.274 & 63.10 &  0.781 &  0.00 \\
\midrule

SimCLR ($\mathcal{T}= 1$)           & 51.03 & 19.24 & 78.07 &  0.404 & 62.04 &  0.958 &  0.08 \\
SimCLR ($\mathcal{T}= 2$)           & 50.50 & 19.57 & 77.07 &  0.399 & 61.89 &  0.944 &  0.06 \\
SimCLR ($\mathcal{T}= 3$)           & 49.62 & 19.14 & 75.81 &  0.394 & 61.69 &  0.925 &  0.05 \\
SimCLR ($\mathcal{T}= 4$)           & 49.00 & 18.90 & 74.95 &  0.391 & 61.79 &  0.914 &  0.04 \\
SimCLR ($\mathcal{T}= 5$)           & 48.42 & 18.73 & 74.30 &  0.390 & 61.87 &  0.909 &  0.03 \\
SimCLR ($\mathcal{T}= 10$)          & 47.69 & 18.42 & 73.09 &  0.389 & 61.88 &  0.906 &  0.02 \\
SimCLR ($\mathcal{T}= 50$)          & 45.54 & 17.62 & 69.90 &  0.375 & 62.16 &  0.979 &  0.00 \\
SimCLR ($\mathcal{T}= 100$)         & 45.24 & 17.50 & 69.43 &  0.371 & 62.30 &  0.977 &  0.00 \\
SimCLR                              & 44.37 & 17.16 & 68.09 &  0.365 & 64.69 &  0.973 & -0.00 \\
\bottomrule
\end{tabular}
\small
\caption{Full metrics for all models used in computing feature importance for CNN models, presented in Figure~\ref{fig:xgboost}.} 
\label{tab:feature_importance_wide}
\vspace{-10mm}
\end{table}

%% file: experiments/single_label_res_full.tex
\begin{table}[htb]
\vspace{-5mm}
\centering
\tiny
\begin{tabular}{l|c|cccccccccccccc|c}
\toprule
Pretrain & \rotatebox[origin=c]{-70}{ImNet} 
& \rotatebox[origin=c]{-70}{Caltech} 
& \rotatebox[origin=c]{-70}{CIFAR10} 
& \rotatebox[origin=c]{-70}{CIFAR100} 
& \rotatebox[origin=c]{-70}{CUB} 
& \rotatebox[origin=c]{-70}{DTD} 
& \rotatebox[origin=c]{-70}{Aircraft} 
& \rotatebox[origin=c]{-70}{Food} 
& \rotatebox[origin=c]{-70}{Indoor} 
& \rotatebox[origin=c]{-70}{Birds} 
& \rotatebox[origin=c]{-70}{Flowers} 
& \rotatebox[origin=c]{-70}{Pets} 
& \rotatebox[origin=c]{-70}{Cars} 
& \rotatebox[origin=c]{-70}{Dogs} 
& \rotatebox[origin=c]{-70}{SUN} 
& \rotatebox[origin=c]{-70}{Transfer} \\
\midrule
Supervised &           78.7 &         86.9 &     97.6 &      86.0 &     85.9 & 69.3 &           82.0 &     85.3 &        81.3 &     74.9 &                99.1 &         92.4 &           94.4 &           82.9 &    65.3 &       0.012 \\
SupCon~\cite{khosla2020supervised} &           77.3 &         86.3 &     97.6 &      85.7 &     85.0 & 69.8 &           84.1 &     86.1 &        81.4 &     73.1 &                99.0 &         91.9 &           94.7 &           83.0 &    65.2 &       0.007 \\
\midrule
CE+SelfSupCon~\cite{islam2021broad} & 76.4 & 86.3 & 97.6 & 85.8 & 85.9 & 69.5 & 83.3 & 86.3 & 80.9 & 74.1 & 98.7 & 92.4 & 94.8 & 85.3 & 64.4 &  0.003 \\
MoCo-v2 ($\mathcal{T}= 1$)        &           78.7 &         87.3 &     97.6 &      86.3 &     85.9 & 70.0 &           83.0 &     86.1 &        82.2 &     73.9 &                99.2 &         92.8 &           94.4 &           84.4 &    65.1 &       0.054 \\
MoCo-v2 ($\mathcal{T}= 2$)        &           77.7 &         87.0 &     97.6 &      86.4 &     85.7 & 70.7 &           82.9 &     86.1 &        81.9 &     73.8 &                99.1 &         92.4 &           94.4 &           82.8 &    65.3 &       0.026 \\
MoCo-v2 ($\mathcal{T}= 3$)        &           77.0 &         86.8 &     97.6 &      86.1 &     85.5 & 70.4 &           83.2 &     86.1 &        81.6 &     73.4 &                98.9 &         92.0 &           94.3 &           82.0 &    64.8 &      -0.001 \\
MoCo-v2 ($\mathcal{T}= 4$)        &           76.0 &         86.4 &     97.6 &      86.1 &     85.7 & 69.7 &           82.8 &     86.1 &        82.3 &     73.3 &                98.9 &         92.0 &           94.3 &           81.5 &    64.8 &      -0.010 \\
MoCo-v2 ($\mathcal{T}= 5$)        &           75.7 &         86.2 &     97.7 &      85.9 &     85.2 & 70.7 &           83.6 &     86.3 &        82.5 &     73.3 &                98.9 &         91.7 &           94.5 &           80.7 &    64.6 &      -0.006 \\
MoCo-v2 ($\mathcal{T}= 10$)       &           73.2 &         85.3 &     97.6 &      85.7 &     84.9 & 69.5 &           82.7 &     86.1 &        81.0 &     72.3 &                98.9 &         91.2 &           94.3 &           78.9 &    63.9 &      -0.056 \\
MoCo-v2 ($\mathcal{T}= 50$)       &           66.2 &         82.8 &     97.3 &      84.4 &     83.7 & 67.1 &           82.9 &     85.7 &        79.1 &     69.6 &                98.9 &         89.2 &           94.3 &           75.7 &    61.3 &     -0.152 \\
MoCo-v2 ($\mathcal{T}= 100$)      &           62.6 &         81.6 &     97.1 &      83.8 &     82.6 & 66.9 &           83.4 &     85.8 &        77.6 &     68.2 &                98.9 &         88.2 &           94.2 &           74.9 &    60.0 &     -0.202 \\
MoCo-v2                         &           61.9 &         78.1 &     96.7 &      82.0 &     79.4 & 66.3 &           80.1 &     85.4 &        75.6 &     61.1 &                98.5 &         86.8 &           93.5 &           73.4 &    56.3 &     -0.352 \\
\midrule
SwAV ($\mathcal{T}= 1$)         &           79.2 &         87.5 &     97.7 &      86.5 &     86.5 & 71.3 &           83.4 &     86.8 &        82.3 &     75.8 &                99.0 &         92.5 &           94.6 &           83.6 &    66.4 &       0.062 \\
SwAV ($\mathcal{T}= 2$)         &           79.0 &         87.8 &     97.7 &      86.9 &     86.9 & 72.3 &           83.7 &     87.2 &        83.4 &     76.0 &                99.3 &         92.4 &           94.7 &           82.7 &    67.1 &      0.104 \\
SwAV ($\mathcal{T}= 3$)         &           78.6 &         88.2 &     97.8 &      87.0 &     86.8 & 72.4 &           83.3 &     87.4 &        84.1 &     76.4 &                99.4 &         92.5 &           94.7 &           82.4 &    67.6 &      0.125 \\
SwAV ($\mathcal{T}= 4$)         &           78.1 &         88.4 &     97.8 &      87.1 &     86.6 & 72.9 &           82.8 &     87.6 &        84.4 &     76.1 &                99.3 &         91.9 &           94.6 &           82.1 &    67.8 &      0.118 \\
SwAV ($\mathcal{T}= 5$)         &           77.6 &         88.6 &     97.9 &      87.2 &     86.4 & 73.0 &           83.2 &     87.6 &        84.1 &     76.1 &                99.3 &         91.6 &           94.6 &           81.8 &    67.9 &      0.118 \\
SwAV ($\mathcal{T}= 10$)        &           76.1 &         88.4 &     97.9 &      87.2 &     85.8 & 72.5 &           82.4 &     87.6 &        84.8 &     75.6 &                99.4 &         91.3 &           94.3 &           81.4 &    68.0 &      0.103 \\
SwAV ($\mathcal{T}= 50$)        &           73.2 &         88.0 &     97.8 &      86.8 &     84.9 & 71.7 &           82.7 &     87.6 &        83.2 &     75.4 &                99.0 &         90.7 &           93.9 &           80.6 &    67.7 &       0.027 \\
SwAV ($\mathcal{T}= 100$)       &           72.0 &         87.8 &     97.9 &      86.6 &     85.0 & 71.7 &           82.9 &     87.5 &        83.6 &     75.2 &                99.0 &         90.4 &           93.8 &           80.5 &    67.6 &       0.028 \\
SwAV                          &           72.0 &         87.0 &     97.8 &      86.6 &     84.3 & 72.1 &           82.3 &     87.4 &        83.1 &     75.1 &                98.9 &         90.3 &           93.6 &           80.6 &    67.8 &       0.005 \\ \hline
DINO ($\mathcal{T}= 1$)         &           77.6 &         87.4 &     97.7 &      86.7 &     86.3 & 71.1 &           83.0 &     87.1 &        82.7 &     76.2 &                99.4 &         92.5 &           94.7 &           82.9 &    66.2 &       0.095 \\
DINO ($\mathcal{T}= 2$)         &           77.7 &         87.4 &     97.7 &      86.9 &     86.7 & 72.1 &           82.8 &     87.3 &        83.1 &     76.3 &                99.4 &         92.1 &           94.7 &           82.4 &    67.0 &      0.106 \\
DINO ($\mathcal{T}= 3$)         &           77.6 &         88.1 &     97.8 &      87.1 &     86.8 & 71.5 &           82.4 &     87.6 &        83.9 &     76.7 &                99.3 &         92.3 &           94.5 &           82.3 &    67.5 &      0.103 \\
DINO ($\mathcal{T}= 4$)         &           77.5 &         88.2 &     97.8 &      87.5 &     86.5 & 72.1 &           82.8 &     87.6 &        84.1 &     76.5 &                99.4 &         92.1 &           94.7 &           82.1 &    67.6 &      0.126 \\
DINO ($\mathcal{T}= 5$)         &           77.1 &         88.3 &     97.9 &      87.3 &     86.0 & 71.9 &           82.6 &     87.7 &        84.7 &     76.4 &                99.3 &         92.1 &           94.7 &           81.7 &    67.7 &      0.116 \\
DINO ($\mathcal{T}= 10$)        &           76.5 &         88.2 &     97.9 &      87.4 &     85.6 & 71.3 &           82.3 &     87.7 &        84.0 &     75.9 &                99.3 &         91.5 &           94.4 &           80.9 &    67.7 &       0.088 \\
DINO ($\mathcal{T}= 50$)        &           76.1 &         87.8 &     98.0 &      87.2 &     84.8 & 72.1 &           82.5 &     87.5 &        82.7 &     75.4 &                99.1 &         90.4 &           94.0 &           80.2 &    67.6 &       0.036 \\
DINO ($\mathcal{T}= 100$)       &           75.8 &         87.7 &     97.8 &      86.8 &     84.7 & 71.7 &           82.1 &     87.6 &        82.4 &     75.4 &                99.0 &         90.1 &           93.9 &           80.0 &    67.4 &       0.010 \\
DINO                          &           75.0 &         87.2 &     97.8 &      86.9 &     83.7 & 72.1 &           80.6 &     87.5 &        83.2 &     74.5 &                98.7 &         89.6 &           93.8 &           80.1 &    67.6 &      -0.024 \\ \hline
SimCLR ($\mathcal{T}= 1$)       &           78.1 &         86.3 &     97.6 &      86.3 &     85.6 & 70.0 &           82.8 &     85.9 &        80.7 &     72.8 &                99.1 &         91.4 &           94.5 &           80.8 &    65.5 &      -0.013 \\
SimCLR ($\mathcal{T}= 2$)       &           77.1 &         86.3 &     97.8 &      86.4 &     84.6 & 69.3 &           82.1 &     85.6 &        80.4 &     70.7 &                98.9 &         90.5 &           94.2 &           79.7 &    64.9 &      -0.058 \\
SimCLR ($\mathcal{T}= 3$)       &           75.8 &         86.6 &     97.9 &      86.5 &     83.9 & 69.8 &           81.7 &     85.1 &        81.0 &     69.2 &                98.7 &         90.1 &           93.7 &           78.5 &    65.1 &      -0.087 \\
SimCLR ($\mathcal{T}= 4$)       &           75.0 &         86.6 &     97.7 &      86.7 &     82.7 & 69.5 &           81.1 &     84.9 &        79.3 &     68.0 &                98.7 &         89.5 &           93.5 &           77.9 &    64.9 &     -0.120 \\
SimCLR ($\mathcal{T}= 5$)       &           74.3 &         86.6 &     97.9 &      86.7 &     82.4 & 69.3 &           80.8 &     84.8 &        79.3 &     67.3 &                98.5 &         89.5 &           93.4 &           77.3 &    64.6 &     -0.138 \\
SimCLR ($\mathcal{T}= 10$)      &           73.1 &         86.3 &     98.0 &      86.3 &     81.1 & 68.7 &           79.2 &     84.5 &        78.4 &     65.5 &                97.8 &         88.6 &           92.7 &           75.9 &    64.1 &     -0.214 \\
SimCLR ($\mathcal{T}= 50$)      &           69.9 &         85.3 &     97.9 &      86.0 &     78.8 & 67.0 &           77.4 &     84.0 &        75.0 &     63.1 &                97.4 &         86.3 &           92.0 &           73.9 &    63.0 &     -0.321 \\
SimCLR ($\mathcal{T}= 100$)     &           69.4 &         85.3 &     97.8 &      86.0 &     78.7 & 66.6 &           77.3 &     84.0 &        74.9 &     62.9 &                97.4 &         86.2 &           91.9 &           73.7 &    63.0 &     -0.327 \\
SimCLR                        &           68.1 &         85.4 &     97.9 &      86.3 &     78.3 & 65.9 &           77.2 &     84.0 &        75.4 &     62.6 &                97.6 &         86.6 &           91.6 &           73.8 &    62.9 &     -0.31.8 \\
\bottomrule
\end{tabular}
\small
\caption{Performance of different \textbf{CNN} models (an extension of Table~\ref{tab:cnn_wide}), including different levels of label injected models) fine-tuned on the downstream datasets in terms of top-1 accuracy (\%) (averaged over 3 runs) and the overall transferability score. The models are grouped by the underlying base SSL method. 
} 
\vspace{-10mm}
\label{tab:cnn_wide_full}
\end{table}

%% file: experiments/single_label_vit_full.tex
\begin{table}[htb]
\centering
\tiny
\begin{tabular}{l|c|cccccccccccccc|c}
\toprule
Pretrain & \rotatebox[origin=c]{-70}{ImNet} 
& \rotatebox[origin=c]{-70}{Caltech} 
& \rotatebox[origin=c]{-70}{CIFAR10} 
& \rotatebox[origin=c]{-70}{CIFAR100} 
& \rotatebox[origin=c]{-70}{CUB} 
& \rotatebox[origin=c]{-70}{DTD} 
& \rotatebox[origin=c]{-70}{Aircraft} 
& \rotatebox[origin=c]{-70}{Food} 
& \rotatebox[origin=c]{-70}{Indoor} 
& \rotatebox[origin=c]{-70}{Birds} 
& \rotatebox[origin=c]{-70}{Flowers} 
& \rotatebox[origin=c]{-70}{Pets} 
& \rotatebox[origin=c]{-70}{Cars} 
& \rotatebox[origin=c]{-70}{Dogs} 
& \rotatebox[origin=c]{-70}{SUN} 
& \rotatebox[origin=c]{-70}{Transfer} \\
\midrule
Supervised              &           81.0 &         90.9 &     98.6 &      89.6 &     82.3 & 70.8 &           60.8 &     87.9 &        83.2 &     79.6 &                91.3 &         93.8 &           85.4 &           91.5 &    68.1 &     -0.101 \\
\midrule
DINO ($\mathcal{T} = 1$)   &           83.2 &         93.1 &     99.0 &      91.2 &     84.7 & 74.6 &           72.1 &     89.9 &        86.3 &     84.9 &                94.7 &         94.3 &           89.4 &           90.5 &    71.5 &      0.148 \\
DINO ($\mathcal{T} = 2$)   &           83.2 &         93.3 &     98.7 &      91.1 &     84.7 & 75.7 &           71.3 &     89.9 &        86.7 &     85.1 &                95.2 &         94.6 &           89.1 &           89.4 &    71.3 &      0.140 \\
DINO ($\mathcal{T} = 3$)   &           82.8 &         93.1 &     98.9 &      90.7 &     84.7 & 75.0 &           71.7 &     89.8 &        86.1 &     84.9 &                95.1 &         94.7 &           89.1 &           88.6 &    71.3 &      0.133 \\
DINO ($\mathcal{T} = 4$)   &           82.3 &         92.8 &     98.8 &      91.0 &     84.7 & 75.2 &           71.2 &     90.0 &        85.8 &     85.2 &                95.4 &         94.7 &           89.1 &           88.3 &    71.5 &      0.131 \\
DINO ($\mathcal{T} = 5$)   &           82.1 &         92.5 &     98.9 &      90.9 &     84.5 & 74.2 &           72.0 &     89.6 &        85.2 &     85.0 &                95.3 &         94.6 &           89.3 &           88.0 &    70.9 &      0.113 \\
DINO ($\mathcal{T} = 10$)  &           80.9 &         92.4 &     98.8 &      90.8 &     84.5 & 74.4 &           71.7 &     89.7 &        85.3 &     84.6 &                95.1 &         94.3 &           88.9 &           87.0 &    70.8 &       0.089 \\
DINO ($\mathcal{T} = 50$)  &           78.1 &         92.0 &     98.6 &      90.4 &     83.3 & 73.5 &           68.6 &     89.5 &        84.9 &     84.2 &                94.9 &         93.7 &           87.4 &           85.3 &    69.7 &       0.012 \\
DINO ($\mathcal{T} = 100$) &           77.0 &         91.7 &     98.6 &      90.0 &     83.3 & 72.9 &           68.2 &     89.5 &        85.5 &     84.2 &                94.2 &         93.8 &           86.8 &           84.6 &    70.0 &      -0.015 \\
DINO                    &            78.2 &         91.1 &     98.5 &      90.0 &     79.5 & 72.1 &           53.0 &     89.3 &        84.5 &     83.9 &                90.1 &         92.2 &           80.9 &           85.2 &    69.3 &     -0.187 \\
\midrule
MAE ($\mathcal{T} = 1$)    &           83.4 &         93.0 &     98.8 &      90.6 &     84.3 & 73.7 &           73.1 &     90.6 &        85.4 &     86.2 &                94.4 &         94.8 &           89.8 &           89.5 &    71.1 &      0.131 \\
MAE ($\mathcal{T} = 2$)    &           82.7 &         93.1 &     98.8 &      90.3 &     84.5 & 74.2 &           72.3 &     90.6 &        85.3 &     86.1 &                94.4 &         94.8 &           89.4 &           88.9 &    71.2 &      0.122 \\
MAE ($\mathcal{T} = 3$)    &           81.8 &         92.7 &     98.8 &      90.4 &     84.0 & 74.1 &           71.6 &     90.2 &        85.0 &     85.5 &                94.7 &         94.8 &           89.7 &           88.1 &    71.1 &      0.107 \\
MAE ($\mathcal{T} = 4$)    &           81.3 &         92.2 &     98.8 &      90.1 &     84.5 & 73.2 &           72.4 &     90.2 &        85.0 &     85.6 &                94.6 &         94.6 &           89.2 &           88.2 &    71.0 &       0.095 \\
MAE ($\mathcal{T} = 5$)    &           80.7 &         92.5 &     98.8 &      89.9 &     83.6 & 73.2 &           71.8 &     90.1 &        84.2 &     85.5 &                94.7 &         94.4 &           89.2 &           87.6 &    70.8 &       0.078 \\
MAE ($\mathcal{T} = 10$)   &           79.0 &         91.8 &     98.7 &      89.5 &     83.6 & 72.7 &           71.1 &     89.7 &        84.4 &     85.0 &                93.8 &         94.4 &           88.5 &           86.9 &    70.5 &       0.029 \\
MAE ($\mathcal{T} = 50$)   &           74.2 &         90.4 &     98.6 &      88.6 &     81.8 & 71.7 &           67.8 &     89.4 &        83.7 &     83.5 &                92.6 &         93.6 &           87.3 &           85.2 &    69.5 &      -0.080 \\
MAE ($\mathcal{T} = 100$)  &           71.9 &         90.2 &     98.5 &      88.0 &     80.9 & 70.7 &           66.3 &     89.2 &        83.1 &     83.5 &                92.3 &         92.8 &           87.1 &           84.7 &    69.1 &     -0.122 \\
MAE                     &           68.0 &         89.2 &     98.1 &      86.9 &     75.4 & 68.5 &           53.8 &     88.8 &        82.2 &     81.2 &                70.6 &         91.2 &           79.6 &           83.2 &    67.6 &     -0.425 \\
\bottomrule
\end{tabular}
\caption{Performance of different \textbf{ViT} models (an extension of Table~\ref{tab:vit_wide}) fine-tuned on the downstream datasets in terms of top-1 accuracy (\%) and the overall transferability score. The models are grouped by the underlying base SSL method. 
} 
\label{tab:vit_wide_full}
\end{table}

%% file: experiments/single_label_res_full_linprob_search.tex
\begin{table}[H]
\centering
\tiny
\begin{tabular}{l|c|cccccccccccccc|c}
\toprule
Pretrain &  
\rotatebox[origin=c]{-70}{ImNet} &  
\rotatebox[origin=c]{-70}{Aircraft}&  
\rotatebox[origin=c]{-70}{Birds}   &
\rotatebox[origin=c]{-70}{CIFAR10}    & 
\rotatebox[origin=c]{-70}{CIFAR100}&  
\rotatebox[origin=c]{-70}{CUB}   &
\rotatebox[origin=c]{-70}{Caltech}     &
\rotatebox[origin=c]{-70}{Cars}   &
\rotatebox[origin=c]{-70}{DTD}   &
\rotatebox[origin=c]{-70}{Dogs}   &
\rotatebox[origin=c]{-70}{Flowers} &
\rotatebox[origin=c]{-70}{Food} &
\rotatebox[origin=c]{-70}{Indoor}    &
\rotatebox[origin=c]{-70}{Pets}   &
\rotatebox[origin=c]{-70}{SUN}   &
\rotatebox[origin=c]{-70}{Transfer}\\
\midrule
Supervised           &           78.7 &      46.4 &   60.9 &     93.0 &      77.1 & 71.5 &     89.1 &  67.4 & 69.5 &  90.4 &     86.5 &  70.0 &    78.7 &  93.0 & 63.1 &       7.3 \\
SupCon~\cite{khosla2020supervised} & 77.3 &   50.9 &   56.6 &     94.9 &      79.2 & 69.0 &     88.5 &  72.6 & 70.2 &  90.5 &     89.1 &  68.6 &    79.3 &  92.5 & 63.6 &   12.6 
\\ \hline
CE + SelfSupCon~\cite{islam2021broad} & 77.3 &      40.2 &   52.8 &     93.3 &      76.5 & 63.0 &     87.3 &  58.1 & 67.6 &  94.0 &     85.6 &  67.4 &    76.6 &  92.6 & 61.3 &  -3.9 \\
MoCo-v2 ($\mathcal{T} = 1$)    &           78.7 &      35.8 &   55.9 &     92.4 &      74.6 & 65.6 &     87.3 &  52.7 & 67.8 &  91.6 &     80.6 &  65.4 &    76.3 &  93.1 & 60.5 &     -12.4 \\
MoCo-v2 ($\mathcal{T} = 2$)    &           77.7 &      38.6 &   56.4 &     92.5 &      74.6 & 65.6 &     87.7 &  55.0 & 67.4 &  89.7 &     81.5 &  66.2 &    76.7 &  92.6 & 60.4 &     -11.8 \\
MoCo-v2 ($\mathcal{T} = 3$)    &           77.0 &      39.9 &   57.4 &     92.4 &      75.1 & 66.3 &     87.2 &  55.7 & 68.0 &  88.2 &     82.3 &  66.3 &    76.0 &  92.1 & 60.3 &     -12.2 \\
MoCo-v2 ($\mathcal{T} = 4$)    &           76.0 &      41.3 &   57.3 &     92.2 &      74.4 & 66.2 &     87.0 &  57.0 & 66.9 &  87.5 &     83.5 &  67.2 &    76.9 &  91.9 & 60.8 &     -11.8 \\
MoCo-v2 ($\mathcal{T} = 5$)    &           75.7 &      41.8 &   56.9 &     92.0 &      75.0 & 66.0 &     86.9 &  56.9 & 66.9 &  86.3 &     83.9 &  66.8 &    75.8 &  92.5 & 59.9 &     -12.7 \\
MoCo-v2 ($\mathcal{T} = 10$)   &           73.2 &      42.0 &   55.6 &     92.4 &      74.1 & 65.8 &     85.9 &  58.0 & 66.3 &  83.8 &     84.0 &  66.4 &    75.4 &  91.4 & 59.8 &     -16.1 \\
MoCo-v2 ($\mathcal{T} = 50$)   &           66.2 &      41.5 &   49.6 &     91.5 &      72.5 & 61.5 &     82.7 &  56.3 & 65.1 &  76.6 &     82.5 &  65.0 &    72.5 &  88.9 & 56.4 &     -31.9 \\
MoCo-v2 ($\mathcal{T} = 100$)  &           62.6 &      39.8 &   46.0 &     91.0 &      71.8 & 59.1 &     81.7 &  55.4 & 64.5 &  73.5 &     81.7 &  64.1 &    69.8 &  87.1 & 54.5 &     -40.4 \\
MoCo-v2                        &           61.9 &      43.9 &   38.9 &     93.4 &      76.4 & 53.8 &     83.5 &  59.3 & 69.7 &  68.0 &     85.3 &  68.5 &    76.0 &  84.6 & 60.6 &     -31.1  \\ \hline
SwAV ($\mathcal{T} = 1$)     &           79.2 &      44.0 &   62.8 &     93.3 &      77.4 & 71.2 &     89.2 &  64.2 & 70.7 &  91.1 &     86.6 &  70.2 &    80.1 &  93.3 & 63.5 &       8.9 \\
SwAV ($\mathcal{T} = 2$)     &           79.0 &      47.7 &   64.0 &     93.6 &      78.4 & 72.8 &     89.3 &  66.7 & 71.4 &  90.0 &     88.6 &  71.7 &    80.1 &  93.6 & 64.6 &      14.3 \\
SwAV ($\mathcal{T} = 3$)     &           78.6 &      50.9 &   65.0 &     93.0 &      78.3 & 73.4 &     89.8 &  67.8 & 72.8 &  88.4 &     89.6 &  72.9 &    81.3 &  93.2 & 65.8 &      16.6 \\
SwAV ($\mathcal{T} = 4$)     &           78.1 &      53.4 &   65.7 &     93.1 &      78.8 & 73.2 &     89.8 &  69.8 & 72.2 &  87.0 &     90.4 &  73.2 &    81.6 &  93.1 & 65.8 &      18.1 \\
SwAV ($\mathcal{T} = 5$)     &           77.6 &      54.2 &   65.4 &     93.6 &      79.1 & 73.0 &     89.4 &  70.3 & 72.7 &  85.9 &     90.3 &  73.8 &    81.9 &  92.9 & 65.6 &      18.2 \\
SwAV ($\mathcal{T} = 10$)    &           76.1 &      55.2 &   65.0 &     93.6 &      78.8 & 72.2 &     88.9 &  71.9 & 72.6 &  83.9 &     90.7 &  74.1 &    82.0 &  92.3 & 65.8 &      17.0 \\
SwAV ($\mathcal{T} = 50$)    &           73.2 &      56.0 &   63.5 &     93.6 &      79.2 & 71.3 &     88.3 &  73.3 & 72.7 &  80.7 &     91.2 &  74.5 &    82.0 &  91.1 & 65.6 &      14.5 \\
SwAV ($\mathcal{T} = 100$)   &           72.0 &      56.3 &   62.1 &     93.5 &      79.1 & 70.9 &     88.2 &  73.6 & 72.8 &  79.9 &     91.1 &  74.3 &    82.0 &  90.7 & 65.7 &      13.1 \\
SwAV                         &           72.0 &      52.0 &   53.3 &     93.2 &      77.8 & 66.7 &     86.5 &  71.0 & 71.4 &  76.4 &     90.6 &  73.2 &    81.6 &  88.9 & 65.1 &       0.3  \\ \hline
DINO ($\mathcal{T} = 1$)     &           77.6 &      46.0 &   63.4 &     93.5 &      78.3 & 73.2 &     89.2 &  66.2 & 71.0 &  90.9 &     87.0 &  71.2 &    80.9 &  93.8 & 64.2 &      13.1 \\
DINO ($\mathcal{T} = 2$)     &           77.7 &      49.9 &   64.8 &     93.7 &      78.4 & 73.5 &     89.1 &  68.3 & 72.6 &  89.3 &     89.5 &  72.9 &    80.8 &  93.7 & 65.1 &      17.5 \\
DINO ($\mathcal{T} = 3$)     &           77.6 &      52.0 &   65.9 &     94.0 &      79.1 & 73.6 &     89.7 &  69.8 & 72.8 &  88.3 &     91.0 &  74.2 &    81.6 &  93.2 & 65.9 &      21.1 \\
DINO ($\mathcal{T} = 4$)     &           77.5 &      53.8 &   66.0 &     93.8 &      79.5 & 74.3 &     89.7 &  70.5 & 73.0 &  86.9 &     91.8 &  74.8 &    82.2 &  93.3 & 66.0 &      22.6 \\
DINO ($\mathcal{T} = 5$)     &           77.1 &      54.5 &   66.0 &     93.8 &      79.4 & 74.1 &     89.5 &  71.3 & 72.5 &  85.8 &     91.9 &  75.2 &    81.8 &  93.0 & 65.9 &      21.8 \\
DINO ($\mathcal{T} = 10$)    &           76.5 &      55.6 &   65.6 &     93.8 &      79.5 & 73.5 &     88.8 &  72.2 & 72.8 &  83.1 &     92.3 &  75.4 &    81.8 &  92.3 & 65.8 &      19.9 \\
DINO ($\mathcal{T} = 50$)    &           76.1 &      56.9 &   63.8 &     93.7 &      79.3 & 72.1 &     88.2 &  74.2 & 73.1 &  79.8 &     92.6 &  75.6 &    82.2 &  90.9 & 65.8 &      17.0 \\
DINO ($\mathcal{T} = 100$)   &           75.8 &      56.6 &   62.4 &     93.9 &      79.6 & 72.0 &     88.1 &  74.5 & 72.9 &  78.7 &     92.6 &  75.3 &    82.2 &  90.6 & 65.9 &      16.1 \\
DINO                         &           75.0 &      54.8 &   54.8 &     93.7 &      78.6 & 68.9 &     87.1 &  74.5 & 72.7 &  75.9 &     92.5 &  74.7 &    81.7 &  89.3 & 66.1 &       8.1 \\ \hline
SimCLR ($\mathcal{T} = 1$)   &           78.1 &      47.8 &   61.1 &     93.8 &      77.7 & 71.3 &     88.7 &  65.8 & 70.9 &  88.9 &     87.5 &  70.5 &    80.2 &  92.9 & 64.0 &       8.9 \\
SimCLR ($\mathcal{T} = 2$)   &           77.1 &      49.9 &   58.6 &     93.8 &      77.6 & 70.1 &     88.3 &  66.5 & 69.8 &  86.6 &     87.6 &  70.3 &    78.9 &  92.5 & 63.7 &       5.2 \\
SimCLR ($\mathcal{T} = 3$)   &           75.8 &      50.0 &   56.9 &     93.7 &      77.9 & 68.9 &     88.1 &  65.9 & 69.8 &  84.2 &     87.7 &  69.8 &    77.8 &  91.7 & 63.2 &       1.1 \\
SimCLR ($\mathcal{T} = 4$)   &           75.0 &      50.4 &   55.3 &     93.7 &      77.6 & 67.5 &     87.5 &  66.8 & 69.8 &  82.4 &     88.0 &  69.2 &    77.0 &  91.3 & 63.0 &      -1.5 \\
SimCLR ($\mathcal{T} = 5$)   &           74.3 &      50.1 &   54.3 &     93.2 &      77.1 & 66.8 &     87.4 &  65.8 & 69.8 &  81.2 &     88.2 &  69.2 &    75.9 &  91.2 & 62.8 &      -4.3 \\
SimCLR ($\mathcal{T} = 10$)  &           73.1 &      49.9 &   51.3 &     93.2 &      77.3 & 64.4 &     86.8 &  65.7 & 69.4 &  77.7 &     87.8 &  68.4 &    76.4 &  89.9 & 62.1 &      -9.6 \\
SimCLR ($\mathcal{T} = 50$)  &           69.9 &      44.3 &   39.4 &     92.9 &      75.0 & 54.5 &     84.3 &  60.2 & 66.1 &  69.4 &     87.0 &  64.5 &    73.1 &  86.6 & 60.2 &     -32.0 \\
SimCLR ($\mathcal{T} = 100$) &           69.4 &      43.8 &   38.0 &     92.5 &      74.2 & 52.5 &     83.7 &  59.1 & 66.2 &  68.3 &     86.6 &  64.0 &    72.5 &  85.8 & 59.8 &     -36.1 \\
SimCLR                       &           68.1 &      43.4 &   35.3 &     89.1 &      69.0 & 50.5 &     82.4 &  56.2 & 65.4 &  65.4 &     85.2 &  62.2 &    72.4 &  83.8 & 58.2 &     -48.0  \\ 
\bottomrule
\end{tabular}
\caption{Linear probing performance of different \textbf{CNN} models (an extension of Table~\ref{tab:cnn_wide_linprob}), including different levels of label injected models) fit on the downstream datasets in terms of top-1 accuracy (\%) and the overall transferability score. The models are grouped by the underlying base SSL method.} 
\label{tab:full_table_linprob}
\end{table}